\documentclass[journal,comsoc]{IEEEtran}

\usepackage[T1]{fontenc}
\ifCLASSINFOpdf
\usepackage{graphicx} 
\usepackage{subfigure}
\usepackage{makecell}
\usepackage{threeparttable}
\usepackage{multirow}
\usepackage{amsthm,amsmath,amssymb}
\usepackage{mathrsfs}
 \usepackage{cite} 
\usepackage{mathptmx}
\usepackage{amssymb}
\usepackage{bbding}

\DeclareMathAlphabet{\mathcal}{OMS}{cmsy}{m}{n}

\usepackage{pifont}
\newcommand{\cmark}{\ding{51}}%
\newcommand{\xmark}{\ding{55}}%
\usepackage{float}
\usepackage{bm}
\usepackage[cmintegrals]{newtxmath}
\usepackage[colorlinks,linkcolor=blue]{hyperref}
\usepackage{array}
\usepackage{booktabs}
\usepackage{xcolor}
\usepackage{soul}

\usepackage{color}
\newcommand{\tabincell}[2]{\begin{tabular}{@{}#1@{}}#2\end{tabular}}
\newcommand{\etal}{\textit{et al}. }
\newcommand{\etall}{\textit{et al}.}
\newcommand{\ie}{\textit{i}.\textit{e}.}
\newcommand{\eg}{\textit{e}.\textit{g}.}

\soulregister{\cite}7 
\soulregister{\citep}7 
\soulregister{\citet}7 
\soulregister{\ref}7 
\soulregister{\pageref}7 

\begin{document}
\title{Bias-Eliminated Semantic Refinement for \\ Any-Shot Learning}
\author{Liangjun~Feng,
        Chunhui~Zhao,~\IEEEmembership{Senior Member,~IEEE},
       and Xi Li,~\IEEEmembership{Senior Member,~IEEE} 

\thanks{This work is supported by the National Science Fund for Distinguished Young Scholars (No. 62125306), the NSFC-Zhejiang Joint Fund for the Integration
of Industrialization and Informatization (No. U1709211) and the Research Project of the State Key Laboratory of Industrial Control Technology (ICT2021A15). (The corresponding author is Chunhui Zhao)}
\thanks{Liangjun Feng and Chunhui Zhao are with the State Key Laboratory of Industrial Control Technology, College of Control Science and Engineering, Zhejiang University, Hangzhou 310027, China. (E-mail:  chhzhao@zju.edu.cn).}
\thanks{Xi Li is with the College of Computer Science and Technology, Zhejiang University, Hangzhou, China, 310027. }}

\markboth{}%
{Shell \MakeLowercase{\textit{et al.}}: Bare Demo of IEEEtran.cls for IEEE Communications Society Journals}
\maketitle

\vspace{-1.5em}
\begin{abstract}
When training samples are scarce, the semantic embedding technique, \ie, describing class labels with attributes, provides a condition to generate visual features for unseen objects by transferring the knowledge from seen objects. However, semantic descriptions are usually obtained in an external paradigm, such as manual annotation, resulting in weak consistency between descriptions and visual features. In this paper, we refine the coarse-grained semantic description for any-shot learning tasks, \ie, zero-shot learning (ZSL), generalized zero-shot learning (GZSL), and few-shot learning (FSL). A new model, namely, the semantic refinement Wasserstein generative adversarial network (SRWGAN) model, is designed with the proposed multihead representation and hierarchical alignment techniques. Unlike conventional methods, semantic refinement is performed with the aim of identifying a bias-eliminated condition for disjoint-class feature generation and is applicable in both inductive and transductive settings. We extensively evaluate model performance on six benchmark datasets and observe state-of-the-art results for any-shot learning; \eg, we obtain 70.2\% harmonic accuracy for the Caltech UCSD Birds (CUB) dataset and 82.2\% harmonic accuracy for the Oxford Flowers (FLO) dataset in the standard GZSL setting. Various visualizations are also provided to show the bias-eliminated generation of SRWGAN. Our code is available.\footnote{Source code: https://github.com/LiangjunFeng/SRWGAN}
\end{abstract}


\IEEEpeerreviewmaketitle
\section{Introduction}
\IEEEPARstart{D}{riven} by advances in deep learning, there has been remarkable progress in visual recognition over recent decades \cite{1,2,3}. Intelligent models, \eg, convolutional neural networks (CNNs), optimize millions of parameters to improve recognition ability based on large-scale annotated datasets and abundant computational resources \cite{4,5}. The classic supervised learning paradigm is effective when sufficient training samples are provided for each object category \cite{6}. However, due to the long-tailed distribution of object categories, the collection of numerous samples for some rarely seen objects is generally challenging in practice.\par
To recognize these rarely seen objects, zero-shot learning (ZSL) \cite{7, 8, 9, 10}, generalized zero-shot learning (GZSL) \cite{11, 12, 13}, and few-shot learning (FSL) \cite{14, 15, 16} have been explored in recent years. ZSL seeks to perform knowledge transfer from seen categories to unseen categories, in which some auxiliary information, \eg, the attribute and textual description of the class label, is used to bridge the class gap. This technique, known as semantic embedding, allows for the identification of a new object by having only a description of it \cite{9,17,18,19}. Empirical studies have also shown that incorporating the features from a pretrained CNN, known as feature embedding, makes ZSL more effective than using the original images \cite{12,20}. However, since the seminal work of Lampert \etall \cite{7}, the studies on ZSL have been limited by a strong assumption that the test data come from unseen classes by default. Since recognition for both seen and unseen object categories is usually required in practice, the unreasonable restriction on ZSL is lifted in GZSL, so that all classes are allowed at the test phase. Hence, GZSL significantly improves the task's practicability in comparison with ZSL \cite{11,12}. Additionally, it may be possible for us to collect a few, \eg, one or two, training images for these rarely seen categories, and then, the FSL setting can be adjusted. Similarly, the general paradigm of FSL is to train a model on classes with sufficient training samples and generalize it to classes with few samples without learning new parameters. This strategy avoids the overfitting risk of directly training the model in the case of a few samples \cite{14,21}. For a unified presentation, we also name the classes with sufficient training samples as seen classes and the classes with a few samples as unseen classes in this paper.\par
Due to the importance and practicability of ZSL, GZSL, and FSL, many solutions have been proposed, such as classic probability-based methods \cite{7,22}, compatibility-based methods \cite{23,24,25,26,27}, and generative models \cite{28,29,30,31}. Among these paradigms, generative models, \eg, generative adversarial networks (GANs) \cite{32,33} and variational autoencoders (VAEs) \cite{34,35}, are promising solutions that synthesize the exemplars of unseen classes and naturally address the insufficient data problem. A basic semantic embedding-based feature generation paradigm is described in Figure 1. As shown, the semantic descriptions bridge the gap between the seen and unseen categories and work as essential conditions for feature generation.
\begin{figure}[htb]
\centering
\includegraphics[width=.45\textwidth]{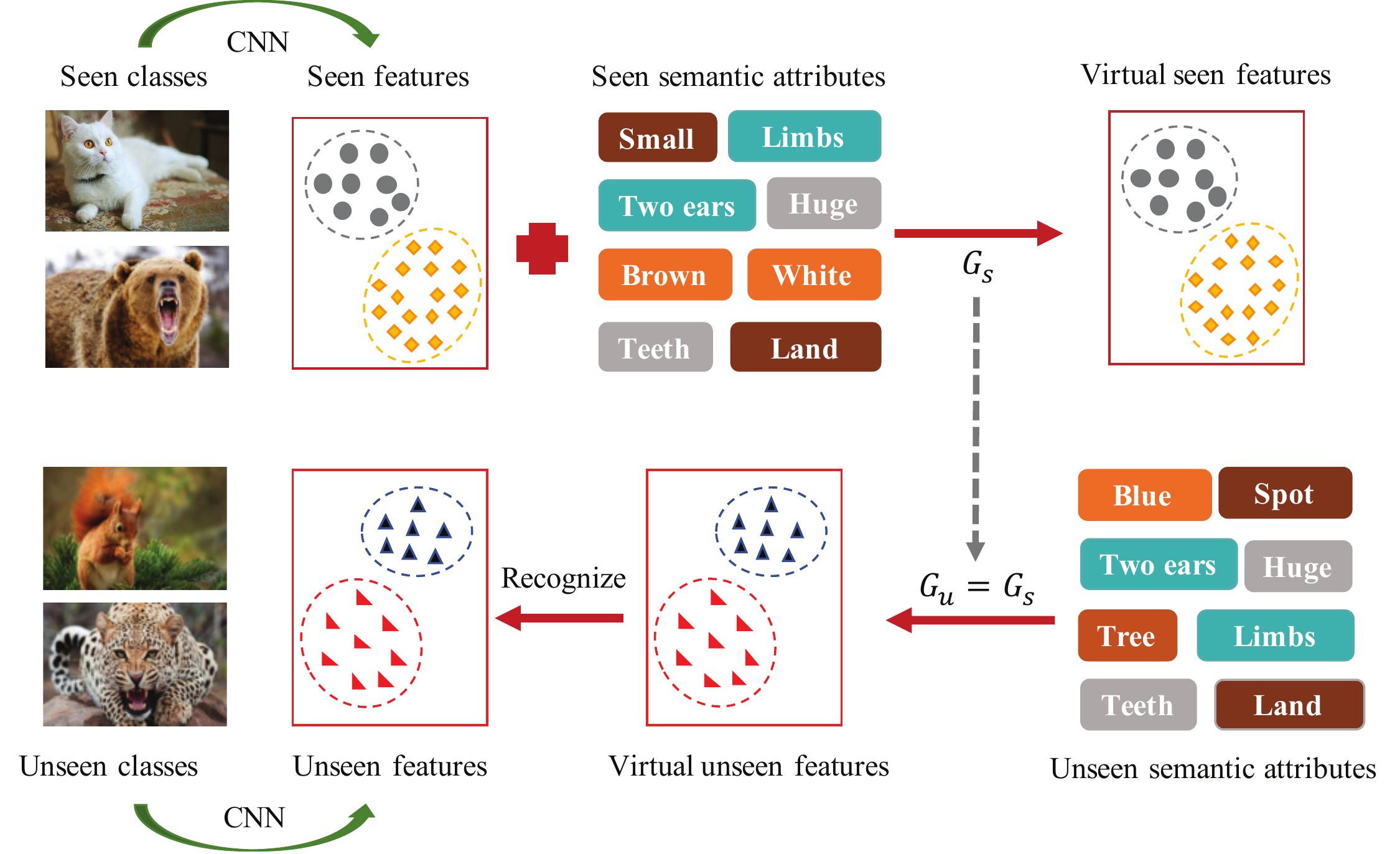} 
\caption{A basic semantic embedding-based feature generation paradigm. The generator of seen classes $G_{s}$ is conditionally trained on the seen semantic attributes and then used as $G_{u}$ to synthesize the features of unseen categories.}
\end{figure}
In practice, semantic descriptions are usually obtained in an external paradigm, such as an individual representation model \cite{36} and manual annotation \cite{7, 37}. For example, eighty-five manually annotated attributes are used in the benchmark Animals with Attributes (AWA) dataset \cite{7} to describe fifty kinds of animals, and 1,024-dimensional features extracted by the character-based CNN-RNN model \cite{38} are used in the benchmark Oxford Flowers (FLO) \cite{36} dataset to describe 102 kinds of flowers. As a secondary representation of objects, a semantic description is supposed to be fine-grained and aligned with visual features for reliable generation and knowledge transfer \cite{39,40}. However, the external paradigm usually extracts coarse-grained semantic descriptions that rarely consider the relationship between the seen and unseen classes. Specifically, Figure 1 reveals that the generator of seen categories, \ie, $G_{s}$, is usually used as that of unseen categories, \ie, $G_{u}$, for unseen feature generation \cite{41,42}. The generator transfer from seen to unseen categories tackles the insufficient data problem for learning tasks. However, the generator $G_{s}$ is not guaranteed to be fully applicable to the unseen categories, \ie, $G_{u} = G_{s}$ \cite{11,12}. Intuitively, neither the adversarial training of GAN nor the variational Bayes algorithm of VAE allows the generator to generate samples for new categories. The mismatch between the semantic and visual spaces makes the bias problem even more challenging. It is desirable to refine the coarse-grained semantic descriptions, and hence, the trained generator can then be more effective for disjoint-class feature generation.\par
To address the above problems, a new model, namely, the semantic refinement Wasserstein generative adversarial network (SRWGAN) model, is developed, in which the semantic description is refined for any-shot learning tasks, \ie, ZSL, GZSL, and FSL. Two proposed techniques, \ie, the multihead semantic representation and the hierarchical semantic alignment, allow SRWGAN to perform bias-eliminated feature generation with abundant and hierarchical semantic information. In SRWGAN, the semantic description of an objective category is no longer a vector but multigroup conditional Gaussian distributions and is no longer independent but highly aligned with visual features. In this way, reliable and effective features can be synthesized by the transferred generator to tackle data insufficiency. In addition, unsupervised regularization in the hierarchical semantic alignment provides effective transductive learning ability for SRWGAN, and then state-of-the-art results can be readily obtained for any-shot learning tasks. Our contributions are threefold:\par
(1) A new model, namely, the semantic refinement Wasserstein generative adversarial network model, is proposed for any-shot tasks and is applicable in both inductive and transductive settings.\par
(2) Two techniques, \ie, the multihead semantic representation and hierarchical semantic alignment, are designed to offer bias-eliminated feature generation for unseen classes.\par
(3) Six benchmark datasets are used in experiments to show state-of-the-art results with detailed analysis for the proposed model and techniques.\par
\section{Related Work}
\subsection{Notations and Formulations for Any-Shot Learning}
We have a set of seen classes, $\mathcal{Y}_{s} = \{1,...,p\}$, and a set of unseen classes, $\mathcal{Y}_{u} = \{p+1,...,p+q\}$. The two sets are disjoint, \ie, $\mathcal{Y}_{s} \cap \mathcal{Y}_{u} = \varnothing$, $p$ is the number of seen classes, and $q$ is the number of unseen classes. Suppose that a dataset of seen classes is $\mathcal{D}_{s} =  \{(x_{s}^{(i)},y_{s}^{(i)},a_{s}^{(i)}),i=1,...,N_{s}\}$, where $x_{s}^{(i)} \in \mathcal{X}_{s}$ is a visual feature with a corresponding seen class label $y_{s}^{(i)} \in \mathcal{Y}_{s}$ and a semantic description $a_{s}^{(i)} \in \mathcal{A}_{s}$. Similarly, a dataset of unseen classes is $\mathcal{D}_{u} =  \{(x_{u}^{(i)},y_{u}^{(i)},a_{u}^{(i)}),i=1,...,N_{u}\}$, where $x_{u}^{(i)} \in \mathcal{X}_{u}$, $y_{u}^{(i)} \in \mathcal{Y}_{u}$, and $a_{u}^{(i)} \in \mathcal{A}_{u}$.\par
In this paper, any-shot learning contains three tasks, \ie, ZSL, GZSL, and FSL. The transductive learning scenarios \cite{43} of ZSL and GZSL, \ie, transductive zero-shot learning (TZSL) and transductive generalized zero-shot learning (TGZSL), are also considered. The five tasks are formulated as follows:\par
\subsubsection{Zero-shot learning} ZSL \cite{7,8} trains a discriminant function based on seen classes and all semantic descriptions. The evaluation is performed on the unseen dataset. The  learning paradigm of ZSL can be summarized as $\mathcal{X}_{s} \cup \mathcal{Y}_{s} \cup \mathcal{A} \rightarrow \mathcal{X}_{u}$.

\subsubsection{Generalized zero-shot learning} GZSL \cite{11,12} is an extension of ZSL. Both the seen and unseen classes are allowed for test. The paradigm of GZSL is $\mathcal{X}_{s}^{tr} \cup \mathcal{Y}_{s}^{tr} \cup \mathcal{A} \rightarrow \mathcal{X}_{u} \cup \mathcal{X}_{s}^{te}$, where $\mathcal{X}_{s}^{tr} \cup \mathcal{Y}_{s}^{tr}$ denotes the seen training sample pairs, and $\mathcal{X}_{s}^{te}$ denotes the seen test samples.

\subsubsection{Few-shot learning} FSL \cite{14,16} has a few unseen sample pairs for training. We perform the FSL task in a general setting, \ie, $\mathcal{X}_{s}^{tr} \cup \mathcal{Y}_{s}^{tr} \cup \mathcal{X}_{u}^{tr} \cup \mathcal{Y}_{u}^{tr} \cup \mathcal{A} \rightarrow \mathcal{X}_{s}^{te} \cup \mathcal{X}_{u}^{te}$, where $\mathcal{X}_{u}^{tr} \cup \mathcal{Y}_{u}^{tr}$ and $\mathcal{X}_{u}^{te}$ denote a few unseen training sample pairs and the unseen test samples, respectively.\par

\subsubsection{Transductive zero-shot learning} TZSL \cite{44,45} uses unlabeled unseen samples with labeled seen samples together for model training. The evaluation is performed on the same unlabeled unseen samples. The paradigm of TZSL can be summarized as $\mathcal{X}_{s} \cup \mathcal{Y}_{s} \cup \mathcal{X}_{u}^{te} \cup \mathcal{A} \rightarrow \mathcal{X}_{u}^{te}$.

\subsubsection{Transductive generalized zero-shot learning} Similarly, TGZSL \cite{46, 47} uses unlabeled test samples with labeled seen samples together for model training. Both the seen and unseen classes are allowed for test. The learning paradigm of TGZSL can be summarized as $\mathcal{X}_{s}^{tr} \cup \mathcal{Y}_{s}^{tr} \cup \mathcal{X}_{u}^{te} \cup \mathcal{X}_{s}^{te} \cup \mathcal{A} \rightarrow \mathcal{X}_{u}^{te} \cup \mathcal{X}_{s}^{te}$.\par

\subsection{Method Review for Any-Shot Learning}
For the ZSL and GZSL tasks in any-shot learning, the solutions can be categorized into classic probability-based methods, compatibility-based methods, and generative models. The probability-based
methods are represented by direct attribute-based prediction and indirect attribute-based prediction \cite{7}, as these methods learn a number of probabilistic classifiers for attributes and combine the scores to make predictions. Feng \etal \cite{37} used incremental classifiers to balance the learning between the seen and unseen classes. The compatibility-based methods learn a compatibility function between the semantic embedding space and the visual embedding space. For example, Romera-Paredes \etal \cite{23} applied squared loss and implicit regularization between embedding spaces to rank the class possibilities. Ji \etal \cite{26} trained a hashing model using the semantic embedding space to transfer knowledge and implement cross-model learning. For generative models, we introduce them in detail in the third subsection. Current FSL research usually adopts meta-learning, distance-based methods, generative methods, etc. Finn \etal \cite{48} designed a model-agnostic meta-learning strategy to learn good weight initialization that can be efficiently adapted to small datasets. Zhang \etal \cite{49} developed a transferable graph model based on relation and distance to mitigate the domain shift problem of knowledge transfer. Xian \etal \cite{21} trained a strong feature generation model based on GAN and VAE to tackle the insufficient data problem of FSL. Vinyals \etal \cite{50} and Jake \etal \cite{15} designed a matching network and prototypical network to predict an image label based on support sets and applied the episode training strategy mimicking few-shot testing. For TZSL and TGZSL, Wu \etal \cite{44} proposed an end-to-end domain-aware generative network by integrating self-supervised learning into a feature generating model. Li \etal \cite{45} addressed TZSL by generating classifiers from class embeddings and used target data to calibrate the classifier generator. Liu \etal \cite{46} proposed a practical latent feature-guided attribute attention framework to perform object-based attribute attention for semantic disambiguation. Fu \etal \cite{47} combined multiple semantic embeddings and performed hypergraph label propagation. Sariyildiz \etal \cite{51} performed TZSL and TGZSL by designing a gradient signal matching-based unsupervised training loss to improve performance. In this paper, ZSL, GZSL, FSL, TZSL, and TGZSL problems are addressed and compared in a unified generative paradigm. 
\subsection{Generative Models}
Generative models synthesize virtual exemplars to address the insufficient data problem and are popular in ZSL, GZSL, and FSL. The most famous generative models are VAEs and GANs. VAE uses an encoder that represents the input as a latent variable with a Gaussian distribution assumption and a decoder that reconstructs the input from the latent variable. Shao \etal \cite{52} proposed a multichannel Gaussian mixture VAE, which introduced a Gaussian mixture model into the multimodal VAE with multiple channels. In contrast, a GAN consists of a generator that synthesizes fake samples and a discriminator that distinguishes fake samples from real samples. Since the standard GAN has a difficult training process, Arjovsky \etal \cite{32} used the Wasserstein distance and constructed the Wasserstein GAN (WGAN) to improve learning stability. Xian \etal \cite{53} proposed the feature-generating GAN to enhance the separability of generated features with an external softmax classifier. Gao \etal \cite{54} designed a joint generative model that coupled VAE and GAN to generate high-quality unseen features. The redundancy-free model proposed by Han \etal \cite{55} took feature generation one step further and extracted the redundancy-free features from virtual samples. An interesting application of the WGAN was the LisGAN by Li \etall \cite{56}, which introduced soul samples as the invariant side to regularize the generating process. In addition, meta-learning is also integrated with WGAN for ZSL and GZSL. Verma \etal \cite{57} proposed a novel episodic training strategy to generate unseen class examples in the training stage, which contributed to overcoming the missing data problem. Our SRWGAN is also based on WGAN and improved by the proposed semantic refinement techniques.\par
\subsection{Semantic Representation and Alignment}
Representation and alignment techniques have been proposed to improve coarse-grained semantic descriptions in recent years. Schonfeld \etal \cite{58} learned latent spaces for semantic descriptions by VAE and then used the designed cross-alignment and distribution-alignment training loss to align the embedding space. Sariyildiz \etal \cite{51} and Wang \etal \cite{59} transferred the semantic description to a conditional multivariate Gaussian distribution to enhance its representation ability. Yu \etal \cite{60} designed a cycle model with two generators in a meta-learning framework to improve the consistency between the visual features and semantic descriptions. Shen \etal \cite{61} proposed an invertible zero-shot flow model to learn factorized data embeddings with a forward pass of an invertible flow network. Similarly, Vyas \etal \cite{62} designed LsrGAN to generate visual features that mirror the semantic relationships. Peng \etal \cite{63} encoded the semantic descriptions into tractable latent distributions, conditioned on the fact that the generative flow optimizes the exact log-likelihood of the observed visual features. However, the bias problems mentioned are usually between visual and semantic spaces. The above methods ignore the relationship between the seen and unseen scopes. From a new view, the proposed SRWGAN provides bias-eliminated generator transfer from seen to unseen classes by augmenting and refining the semantic information.
\begin{figure*}[t]
\begin{center}
\includegraphics[width=1\linewidth]{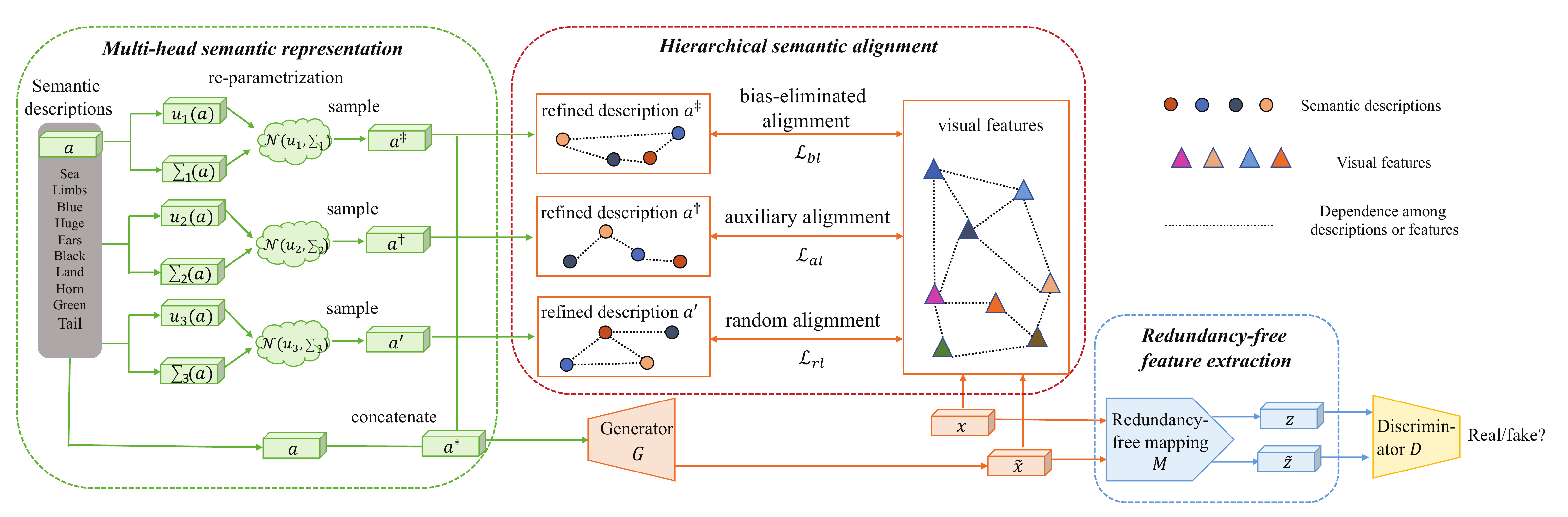}
\end{center}
   \caption{The structure of the SRWGAN. The SRWGAN learns a feature generator, $G$, under the generative adversarial paradigm. The input of $G$ is the concatenated semantic description $a^{*}=[a^{\ddagger},a^{\dagger},{a}',a]$, and the output is the fake feature $\tilde{x}$. The multihead semantic representation technique provides three refined descriptions $a^{\ddagger},a^{\dagger},{a}'$ for $G$, which are regularized by the hierarchical semantic alignment technique from bias-eliminated $\mathcal{L}_{bl}$, auxiliary $\mathcal{L}_{al}$, and random $\mathcal{L}_{rl}$ views. Redundancy-free mapping, $M$, in \cite{55} is also used to extract the clean feature $z$ for better classification and discrimination.}
\end{figure*}
\section{Proposed Approach}
\noindent\textbf{Overall Idea} The proposed SRWGAN is illustrated in Figure 2. To achieve bias-free generator transfer from seen classes to unseen classes, we perform bias-eliminated alignment $\mathcal{L}_{bl}$ between the semantic space $\mathcal{A}$ and visual feature space $\mathcal{X}$, which refines the semantic description from $a$ to $a^{\ddagger}$. With similar paradigms, auxiliary alignment $\mathcal{L}_{al}$ and random alignment $\mathcal{L}_{rl}$ are designed to enhance the model's feature generation ability and transductive learning ability. The three kinds of alignment loss form the proposed hierarchical semantic alignment technique for the designed multihead semantic representation $a^{\ddagger},a^{\dagger},{a}'$. Moreover, redundancy-free mapping $M$ is applied in model implementation to make the latent feature $z$ intraclass contraction and interclass separation for classification.\par

\subsection{Motivation Analysis: Bias-Eliminated Condition for Generator Transfer}
Here, the bias-eliminated condition for generator transfer from seen classes to unseen classes is analyzed to explain our motivation for semantic refinement.\par
Let $U_{xs} = [\mu_{xs}^{(1)},...,\mu_{xs}^{(p)}] \in \mathbb{R}^{p \times n}$ and $U_{xu} = [\mu_{xu}^{(1)},...,\mu_{xu}^{(q)}] \in \mathbb{R}^{q \times n}$, denoting the category-prototype matrices in $\mathcal{X}$ for the seen and unseen classes, respectively. Similarly, we have $U_{as} = [\mu_{as}^{(1)},...,\mu_{as}^{(p)}] \in \mathbb{R}^{p \times d}$ and $U_{au} = [\mu_{au}^{(1)},...,\mu_{au}^{(q)}] \in \mathbb{R}^{q \times d}$, denoting the prototype matrices in $\mathcal{A}$. The prototype $\mu$ is the mean of a category, and $n$ and $d$ denote the dimensions of visual features and semantic descriptions, respectively.\par
\noindent\textbf{Definition 1} A \textit{matching function}, $f$, for the visual feature, $U_{x}$, and semantic description, $U_{a}$, can be given with a trainable parameter, $W \in \mathbb{R}^{n \times d}$, as:
\begin{equation}
f(U_{x},U_{a};W) = U_{x}WU_{a}^{T}.
\end{equation}\par
The function $f(U_{x},U_{a};W)$ outputs a matrix $U_{y}$, which scores the matching degree between each visual feature and semantic description prototype. Considering the seen scope, $\mathcal{Y}_{s}$, and unseen scope, $\mathcal{Y}_{u}$, we have:\par
\noindent\textbf{Definition 2} An optimal $W^{*}$ can be obtained when the \textit{seen bridging condition} (SBC) $f(U_{xs},U_{as};W^{*}) = U_{ys}$ and the \textit{unseen bridging condition} (UBC) $f(U_{xu},U_{au};W^{*}) = U_{yu}$ are satisfied.\par 

Here, $U_{ys} = I_{s} \in \mathbb{R}^{p \times p}$ and $U_{yu} = I_{u} \in \mathbb{R}^{q \times q}$ are one-hot labels, which denote that the visual feature and semantic description of each category are correctly matched. Additionally, we have two generators, $G_{s}(U_{as}):U_{as} \rightarrow U_{xs}$ and $G_{u}(U_{au}):U_{au} \rightarrow U_{xu}$, for the seen and unseen classes, respectively. Without loss of generality, we assume that the generators are linear, \ie, $G_{s}  = Z_{s} \in \mathbb{R}^{d \times n}$ and $G_{u}  = Z_{u} \in \mathbb{R}^{d \times n}$. The feature generation process can then be denoted by $G_{s}(U_{as}) = U_{as}Z_{s} = U_{xs}$ and $G_{u}(U_{au}) = U_{au}Z_{u} = U_{xu}$ for the seen and unseen classes, respectively. Based on the above notations and definitions, the bias-eliminated generator transfer can be given by the following lemma:\par
\noindent\textbf{Lemma} Given $W^{*}$, if $U_{as}^{T}U_{as} = U_{au}^{T}U_{au}$, namely, \textit{cross bridging condition} (CBC), even a linear generator $G_{s}(U_{as}):U_{as} \rightarrow U_{xs}$ learned from the seen set $\mathcal{Y}_{s}$ can be used as $G_{u}(U_{au}):U_{au} \rightarrow U_{xu}$ to synthesize exemplars for $\mathcal{Y}_{u}$ without bias.\par
\noindent\textbf{Proof} The use of $G_{s}(U_{as})$ and $G_{u}(U_{au})$ replaces $U_{xs}$ and $U_{xu}$ in SBC and UBC, respectively, and we have:
\begin{equation}
\begin{aligned}
G_{s}(U_{as})W^{*}U_{as}^{T}  = U_{as}Z_{s}W^{*}U_{as}^{T} = U_{ys},
\end{aligned}
\end{equation}
\begin{equation}
\begin{aligned}
G_{u}(U_{au})W^{*}U_{au}^{T}  = U_{au}Z_{u}W^{*}U_{au}^{T} = U_{yu}.
\end{aligned}
\end{equation}
Let $Q_{s} = Z_{s}W^{*}$ and $Q_{u} = Z_{u}W^{*}$,
\begin{equation}
\begin{aligned}
Q_{u} &= U_{au}^{-1}U_{yu}(U_{au}^{T})^{-1} = U_{au}^{-1}I_{u}(U_{au}^{T})^{-1} = U_{au}^{-1}(U_{au}^{T})^{-1} \\
      &= (U_{au}^{T}U_{au})^{-1} = (U_{as}^{T}U_{as})^{-1} = U_{as}^{-1}I_{s}(U_{as}^{T})^{-1} \\
      &= U_{as}^{-1}U_{ys}(U_{as}^{T})^{-1} = Q_{s}.
\end{aligned}
\end{equation}\par
Then, identical generator $Z_{u} =  Z_{s}$, \ie, $G_{u} = G_{s}$, can be obtained. The bias-eliminated condition for the disjoint-class generator transfer is summarized as follows:\par
\noindent\textbf{Bias-Eliminated Condition} (1) SBC, (2) UBC, and (3) CBC.\par
For better understanding, we visualize the bias-eliminated condition in Figure 3. SBC and UBC align the semantic space $\mathcal{A}$ and feature space $\mathcal{X}$ in the seen and unseen scopes, respectively. The third condition provides a cross-scope connection for generator transfer. In implementation, it is challenging for coarse-grained semantic descriptions to satisfy SBC, UBC, and CBC concurrently. Hence, we refine the semantic description for the bias-eliminated condition. Additionally, in UBC, unseen exemplars, \ie, $U_{xu}$, are needed, while none or a few unseen samples are available in ZSL, GZSL, and FSL for model training. We address this problem and describe the proposed semantic refinement techniques in detail in the next subsection.\par
\begin{figure}[htb]
\centering
\includegraphics[width=.45\textwidth]{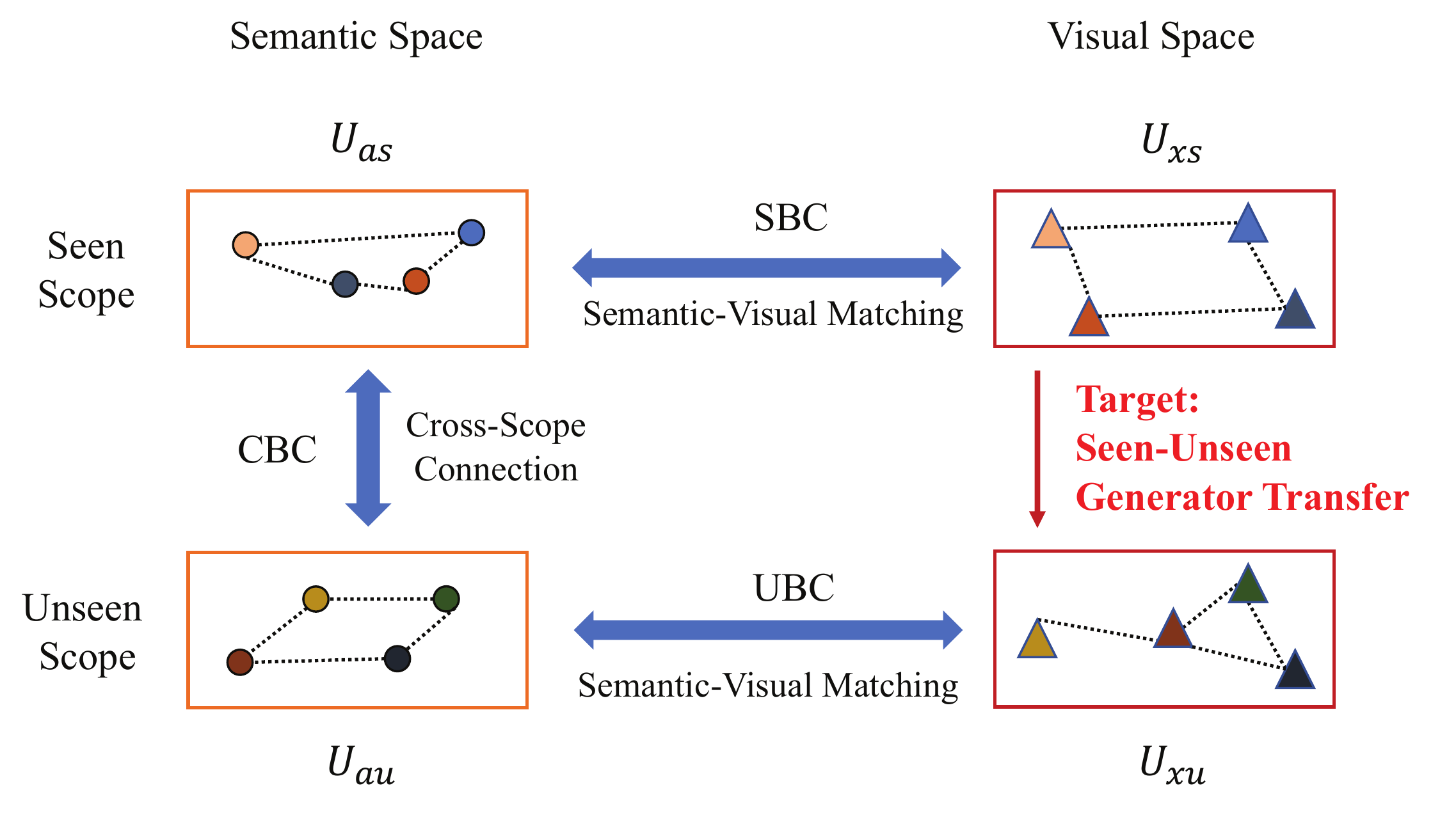} 
\caption{Visualization for the three bias-eliminated conditions.}
\end{figure}
\subsection{Semantic Refinement Techniques}
In SRWGAN, two semantic refinement techniques are proposed, \ie, the multihead semantic representation and hierarchical semantic alignment. For clarification, we introduce the multihead semantic representation starting with the Gaussian semantic representation and the hierarchical semantic alignment starting with the bias-eliminated semantic alignment.\par
\subsubsection{Gaussian semantic representation} {In any-shot learning, a semantic vector $a$ is provided as the learning condition for each class. To model the latent distribution corresponding to its class, the standard Gaussian distribution $\mathcal{N}(0,I)$ is usually imposed on $a$ as prior knowledge, thus defining a conditional multivariate Gaussian distribution $\mathcal{N}(\mu(a),\sum(a))$. A sample can be taken from this distribution as the input of the feature generator. The $\mathbb{R}^{d_{a}}$-dimensional mean $\mu(a)$ and covariance $\sum(a)$ are estimated as follows:
\begin{equation}
\begin{aligned}
\mu(a) = LeakyReLU(W_{u}a+b_{u}), \\
\sum(a) = LeakyReLU(W_{c}a+b_{c}),
\end{aligned}
\end{equation}
where $W \in \mathbb{R}^{d_{a} \times d}$ and $b \in \mathbb{R}^{d_{a}}$ are the trainable weights and biases, respectively, and $LeakyReLU$ is the activation function. For the $i$-th class, the Gaussian semantic representation requests $\mu(a_{i})$ and $\sum(a_{i})$ and then takes a sample from $\mathcal{N}(\mu(a_{i}),\sum(a_{i}))$. The reparameterization strategy in \cite{35} is used to make the sampling differentiable. The Gaussian semantic representation is used by Verm \etal \cite{64} and Sariyildiz \etal \cite{51} to enhance the representation of the semantic description from a single vector to a multivariate Gaussian distribution.\par

\subsubsection{Multihead semantic representation}
In this paper, we extend the Gaussian semantic representation to a multihead setting. Three different multivariate Gaussian distributions, \ie, $\mathcal{N}(\mu_{i}(a), \sum\nolimits_{i}(a))$ and $i=1,2,3$, are estimated conditioned on $a$ with three groups of weights and biases. As shown in Figure 2, three semantic descriptions, \ie, $a^{\ddagger}$, $a^{\dagger}$, and ${a}'$, are sampled from $\mathcal{N}(\mu_{1}(a), \sum\nolimits_{1}(a_{i}))$, $\mathcal{N}(\mu_{2}(a), \sum\nolimits_{2}(a))$, and $\mathcal{N}(\mu_{3}(a), \sum\nolimits_{3}(a))$, constructing the refined description $a^{*}=[a^{\ddagger},a^{\dagger},{a}',a]$. In comparison with the common Gaussian semantic representation, two additional descriptions are provided for feature generation. However, it is noted that the three multivariate Gaussian distributions may have similar information, since all of them are estimated in the same way and conditioned on the common $a$. In the next two subsections, we design three types of semantic alignment losses to regularize the three semantic descriptions for disjoint-class feature generation in a hierarchical way.\par


\subsubsection{Bias-eliminated semantic alignment}
Based on the \emph{bias-eliminated condition}, the bias-eliminated alignment loss $\mathcal{L}_{bl}$ for $a^{\ddagger}$ is designed as:\par
\begin{equation}
\begin{aligned}
\mathcal{L}_{bl} = \alpha \mathcal{L}_{SBC} + \beta \mathcal{L}_{UBC} + \delta \mathcal{L}_{CBC},
\end{aligned}
\end{equation}
where $\mathcal{L}_{SBC}$, $\mathcal{L}_{UBC}$, and $\mathcal{L}_{CBC}$ are the implementations of SBC, UBC, and CBC, respectively, and $\alpha$, $\beta$, and $\delta$ are given weights. \par

First, $\mathcal{L}_{SBC}$ is given as:\par
\begin{equation}
\begin{aligned}
\mathcal{L}_{SBC} = -\mathbb{E}_{p(x_{s})}[logq(x_{s})], 
\end{aligned}
\end{equation}
and
\begin{equation}
\begin{aligned}
q(x_{s}) = softmax(f(x_{s},U_{a^{\ddagger}s};W)),
\end{aligned}
\end{equation}
where $U_{a^{\ddagger}s} \in \mathbb{R}^{p \times d_{a^{\ddagger}}}$ is the refined semantic prototype matrix of $a^{\ddagger}$ for seen classes and $d_{a^{\ddagger}}$ is the dimension of $a^{\ddagger}$. In a supervised paradigm, $\mathcal{L}_{SBC}$ applies cross-entropy loss to achieve SBC at the sample level.\par
Second, $\mathcal{L}_{UBC}$ is given as :
\begin{equation}
\begin{aligned}
\mathcal{L}_{UBC} = -\mathbb{E}_{p(\tilde{x}_{u})}[q(\tilde{x}_{u})logq(\tilde{x}_{u})],
\end{aligned}
\end{equation}
and 
\begin{equation}
\begin{aligned}
q(\tilde{x}_{u}) = softmax(f(\tilde{x}_{u},U_{a^{\ddagger}u};W)),
\end{aligned}
\end{equation}
where $U_{a^{\ddagger}u} \in \mathbb{R}^{q \times d_{a^{\ddagger}}}$ is the refined semantic prototype matrix of $a^{\ddagger}$ for unseen classes. Because few real samples are available for unseen classes, $\mathcal{L}_{UBC}$ applies the entropy loss to learn UBC from fake unseen samples in an unsupervised manner. Note that the fake data $\tilde{x}_{u}$ should not be regularized in a supervised paradigm, which may result in the mode collapse problem and has been discussed in \cite{51}.\par
For $\mathcal{L}_{CBC}$, the implementation is
\begin{equation}
\begin{aligned}
\mathcal{L}_{CBC} = \frac{1}{d_{a^{\ddagger}}}||U^{T}_{a^{\ddagger}s}U_{a^{\ddagger}s} - U^{T}_{a^{\ddagger}u}U_{a^{\ddagger}u}||^{2}_{2}.
\end{aligned}
\end{equation}\par
Specifically, the losses $\mathcal{L}_{SBC}$ and $\mathcal{L}_{UBC}$ require that the refined description $a^{\ddagger}$ be aligned with the visual features in the seen scope and in the unseen scope, respectively. The loss $\mathcal{L}_{CBC}$ builds connections between the seen and unseen scopes in the semantic embedding space. The bias-eliminated condition for generator transfer is implemented with the three losses together. \par

\subsubsection{Hierarchical semantic alignment}
With the above discussed bias-eliminated alignment loss $\mathcal{L}_{bl}$, the inductive hierarchical semantic alignment loss is given as:
\begin{equation}
\begin{aligned}
\mathcal{L}_{in-sr} = \mathcal{L}_{bl} + \mathcal{L}_{in-al} + \mathcal{L}_{rl},
\end{aligned}
\end{equation}
and the transductive hierarchical semantic alignment loss is given as:
\begin{equation}
\begin{aligned}
\mathcal{L}_{tr-sr} = \mathcal{L}_{bl} + \mathcal{L}_{tr-al} + \mathcal{L}_{rl},
\end{aligned}
\end{equation}
where $\mathcal{L}_{in-al}$ and $\mathcal{L}_{tr-al}$ are the inductive and transductive auxiliary alignment loss for $a^{\dagger}$, respectively, and $\mathcal{L}_{rl}$ is the random alignment loss for ${a}'$.\par
The auxiliary alignment loss learns semantic-visual matching from unlabeled data and is applicable in both transductive and inductive settings. Similar to $\mathcal{L}_{bl}$, the inductive auxiliary alignment loss $\mathcal{L}_{in-al}$ is defined as:
\begin{equation}
\begin{aligned}
\mathcal{L}_{in-al} & = \alpha \mathcal{L}_{UN-SBC} + \delta \mathcal{L}_{CBC} \\
       & = -\alpha \mathbb{E}_{p(x_{s})}[q(x_{s})logq(x_{s})] \\
       & +  \frac{\delta}{d_{a^{\dagger}}}||U^{T}_{a^{\dagger}s}U_{a^{\dagger}s} - U^{T}_{a^{\dagger}u}U_{a^{\dagger}u}||^{2}_{2},
\end{aligned}
\end{equation}
and the transductive $\mathcal{L}_{tr-al}$ is defined as:
\begin{equation}
\begin{aligned}
&\mathcal{L}_{tr-al} = \alpha \mathcal{L}_{UN-SBC} + \beta \mathcal{L}_{UN-UBC} + \delta \mathcal{L}_{CBC} \\
       & = -\alpha \mathbb{E}_{p(x_{s})}[q(x_{s})logq(x_{s})] -\beta \mathbb{E}_{p(x_{u})}[q(x_{u})logq(x_{u})] \\
       & +  \frac{\delta}{d_{a^{\dagger}}}||U^{T}_{a^{\dagger}s}U_{a^{\dagger}s} - U^{T}_{a^{\dagger}u}U_{a^{\dagger}u}||^{2}_{2},
\end{aligned}
\end{equation}
where $U_{a^{\dagger}s} \in \mathbb{R}^{p \times d_{a^{\dagger}}}$ and $U_{a^{\dagger}u} \in \mathbb{R}^{q \times d_{a^{\dagger}}}$ are the refined semantic prototype matrices of $a^{\dagger}$ for the seen and unseen classes, respectively, and $d_{a^{\dagger}}$ is the dimension of $a^{\dagger}$.\par
In comparison with $\mathcal{L}_{bl}$, $\mathcal{L}_{al}$ is completely based on unlabeled samples and releases the one-hot restriction in SBC and UBC, which enhances the semantic-visual matching from the entropy view and makes the model effective for transductive learning. Moreover, a random alignment loss $\mathcal{L}_{rl}$ is applied on ${a}'$ to enrich the seen-unseen compactness randomly for disjoint-class feature generation in the semantic embedding space, which is given as:
\begin{equation}
\begin{aligned}
\mathcal{L}_{rl} &=  \delta \mathcal{L}_{CBC} =   \frac{\delta}{d_{{a}'}}||U^{T}_{{a}'s}U_{{a}'s} - U^{T}_{{a}'u}U_{{a}'u}||^{2}_{2},
\end{aligned}
\end{equation}
where $U_{{a}'s} \in \mathbb{R}^{p \times d_{{a}'}}$ and $U_{{a}'u} \in \mathbb{R}^{q \times d_{{a}'}}$ are the refined semantic prototype matrices of ${a}'$ for the seen and unseen classes, respectively, and $d_{{a}'}$ is the dimension of ${a}'$.\par
To summarize, the designed hierarchical semantic alignment technique takes the bias-eliminated condition into consideration and learns from both labeled and unlabeled data with triple alignments. In experiments, the hierarchical loss provides more abundant semantic information than the single loss does and shows higher accuracy improvement.

\subsection{Regularized Feature Generation for Any-Shot Learning}
For feature generation, we begin from the WGAN \cite{32}, which consists of a generator $G$ and a discriminator $D$. The training loss of WGAN is
\begin{equation}
\begin{aligned}
\mathcal{L}_{wgan} &= \mathbb{E}_{p(x_{s})}[D(x_{s})] - \mathbb{E}_{p_{G}(\tilde{x}_{s})}[D(\tilde{x}_{s})] \\
         & + \mathbb{E}_{p(\hat{x}_{s})}[(||\bigtriangledown_{\hat{x}_{s}}D(\hat{x}_{s})||_{2}-1)^{2}],
\end{aligned}
\end{equation}
where $\tilde{x}_{s} = G(a^{*}_{s})$ is the fake seen feature based on the refined description $a^{*}_{s}$, and $\hat{x}_{s}$ is the interpolation between $x_{s}$ and $\tilde{x}_{s}$.\par
Since ZSL, GZSL, and FSL are evaluated as a classification task, a popular strategy is to apply a pretrained classifier on the generated features \cite{53}, and hence, the fake features can be more separable. Instead, we use the similarity between the fake and real features as a classification regularization item, which is defined as:
\begin{equation}
\begin{aligned}
\mathcal{L}_{cls} = -\mathbb{E}_{p(\tilde{x}_{s})}[log(c(\tilde{x}_{s}))],
\end{aligned}
\end{equation}
where $c(\tilde{x}_{s}) = softmax(\tilde{x}_{s}U^{T}_{xs})$, and $U_{xs}$ is the real category-prototype matrix in $\mathcal{X}_{s}$. In comparison with the conventional pretrained classifier, Eq. (18) optimizes the similarity between the real and fake seen features by the cross-entropy loss at each training batch, which enhances the compactness between the real and fake features.\par
Additionally, the redundancy-free mapping approach in \cite{55} is used to extract clean features from the original visual features, \ie, $z_{s}=M(x_{s})$, for better discrimination. The objective of $M$ is 
 \begin{equation}
\begin{aligned}
\mathcal{L}_{m} &= \mathbb{E}_{p_{M}(z_{s}|x_{s})}[max(0,\Delta+||z_{s}-c_{y}||^{2}_{2}-||z_{s}-c_{{y}'}||^{2}_{2})] \\
& s.t. \; \mathbb{E}_{p_{M}(z_{s}|x_{s})}[D_{KL}[p_{M}(z_{s}|x_{s})||r(z_{s})]] \leq b,
\end{aligned}
\end{equation}
where $p_{M}(z_{s}|x_{s})$ denotes the conditional distribution of $z_{s}$ on $x_{s}$, $\Delta$ is a margin to make $M$ more robust, $c$ is the randomly initialized class center and is optimized with $M$, $y$ is the class label of $x$, ${y}'$ is the class label other than $y$, $D_{KL}$ denotes the Kullback-Leibler (KL) divergence, and $b$ is the imposed upper bound.\par
\begin{table}[!htb]
\centering
    \setlength{\tabcolsep}{2 mm}
    \centering
    \begin{tabular}{ll}
        \Xhline{1pt}
        \multicolumn{2}{l}{\textbf{Algorithm 1} Learning Procedure of the Proposed SRWGAN}\\
        \hline
        \multicolumn{2}{l}{\textbf{Inductive Input -> Output:}}\\
        \multicolumn{2}{l}{\tabincell{l}{ \;\;\;\;\; $\mathcal{X}_{s} \cup \mathcal{Y}_{s} \cup \mathcal{A} \rightarrow \mathcal{X}_{u}$ for ZSL; $\mathcal{X}_{s}^{tr} \cup \mathcal{Y}_{s}^{tr} \cup \mathcal{A} \rightarrow \mathcal{X}_{u} \cup \mathcal{X}_{s}^{te}$ for GZSL; \\ \;\;\;\;\; $\mathcal{X}_{s}^{tr} \cup \mathcal{Y}_{s}^{tr} \cup \mathcal{X}_{u}^{tr} \cup \mathcal{Y}_{u}^{tr} \cup \mathcal{A} \rightarrow \mathcal{X}_{s}^{te} \cup \mathcal{X}_{u}^{te}$ for (G)FSL.}} \\
        \multicolumn{2}{l}{\textbf{Transductive Input -> Output:}}\\
        \multicolumn{2}{l}{\tabincell{l}{ \;\;\;\;\; $\mathcal{X}_{s} \cup \mathcal{Y}_{s} \cup \mathcal{X}_{u}^{te} \cup \mathcal{A} \rightarrow \mathcal{X}_{u}^{te}$ for TZSL; \\ \;\;\;\;\; $\mathcal{X}_{s}^{tr} \cup \mathcal{Y}_{s}^{tr} \cup \mathcal{X}_{u}^{te} \cup \mathcal{X}_{s}^{te} \cup \mathcal{A} \rightarrow \mathcal{X}_{u}^{te} \cup \mathcal{X}_{s}^{te}$ for TGZSL.}} \\
        \hline
        \multicolumn{2}{l}{\textbf{Training Phase:}}\\
        1    & \tabincell{l}{Construct the computational graph as in Figure 2 and initialize the \\model weights with the normal distribution ($\mathcal{N}(0,0.02)$).}  \\
        2    & \tabincell{l}{Four Adam optimizers (0.5, 0.999) are used for optimization. \\ Optimizer C $\rightarrow M$ in Eq. (19); Optimizer D $\rightarrow D$ in Eq. (20); \\ Optimizer G $\rightarrow G$ in Eq. (20) (including the multihead modules); \\ Optimizer B $\rightarrow W$ in Eq. (8).\\ }  \\
        \multicolumn{2}{l}{\;\;\;\;\;\;\; \textbf{for $i=1:n$ (number of epochs) do}}\\
        \multicolumn{2}{l}{\;\;\;\;\;\;\;\;\;\; \textbf{for $j=1:m$ (rounds in an epoch) do}}\\
        \multicolumn{2}{l}{\;\;\;\;\;\;\;\;\;\;\;\;\; \textbf{for $m=1:5$ (discriminator training) do}}\\     
        3   & \tabincell{l}{\;\;\;\;\;\;\;\; Sample a batch of data.}\\   
        4   & \tabincell{l}{\;\;\;\;\;\;\;\; Obtain $\mathcal{L}_{m}$ in Eq. (19) and $\mathcal{L}^{z}_{wgan}$ of $D$ in Eq. (20)}.\\
        5   & \tabincell{l}{\;\;\;\;\;\;\;\; Perform backpropagation for $\lambda_{m}\mathcal{L}_{m}$ and $\mathcal{L}^{z}_{wgan}$ of $D$}.\\
        6   & \tabincell{l}{\;\;\;\;\;\;\;\; Apply optimizer C for $M$ and apply optimizer D for $D$.}\\
            & \tabincell{l}{\;\;\;\;\; \textbf{end for}}\\
        7   & \tabincell{l}{\;\;\;\;\; Sample a batch of data.}\\
        8   & \tabincell{l}{\;\;\;\;\; Obtain $\mathcal{L}_{cls}$ in Eq. (18), and $\mathcal{L}^{z}_{wgan}$ of $G$ in Eq. (20).}\\
        9   & \tabincell{l}{\;\;\;\;\; Perform backpropagation for $\lambda_{c}\mathcal{L}_{cls}$ and $\mathcal{L}^{z}_{wgan}$ of G.}\\
        10  & \tabincell{l}{\;\;\;\;\; Apply optimizer G for $G$. }\\    
        11   & \tabincell{l}{\;\;\;\;\; \textbf{If} $i > k$ ($k$ defines the number of pretraining epochs).}\\
        12   & \tabincell{l}{\;\;\;\;\;\;\;\; Obtain $\mathcal{L}_{sr}$ in Eq. (12) or (13).}\\
        13   & \tabincell{l}{\;\;\;\;\;\;\;\; Perform backpropagation for $\lambda_{a}\mathcal{L}_{sr}$.}\\
        14  & \tabincell{l}{\;\;\;\;\;\;\;\; Apply optimizer B for $W$. }\\
        \multicolumn{2}{l}{\;\;\;\;\;\;\;\;\; \textbf{end for}}\\
        \multicolumn{2}{l}{\;\;\;\;\;\;\; \textbf{end for}}\\
        \hline
        \multicolumn{2}{l}{\textbf{Testing Phase:}}\\
        15   & \tabincell{l}{Generate virtual unseen samples using $\tilde{x}_{u} = G(a^{*}_{u})$}.\\
        16   & \tabincell{l}{Training a softmax classifier with $[x_{s}, \tilde{x}_{u}]$ for any-shot learning.}\\
        \Xhline{1pt}  
    \end{tabular}
\end{table}

The constraint in Eq. (19) determines how much information in $x_{s}$ can be conveyed to $z_{s}$ evaluated by the KL divergence, and the main objective decides whether the remaining information is discriminative or not. The strategy in \cite{63} is applied to optimize an unconstrained form of Eq. (19) derived by the Lagrange multiplier method. With redundancy-free mapping $M$, the generative adversarial loss in Eq. (17) can be rewritten as follows: 
\begin{equation}
\begin{aligned}
\mathcal{L}^{z}_{wgan} &= \mathbb{E}_{p_{M}(z_{s}|x_{s})}[D(z_{s})] - \mathbb{E}_{p_{G,M}(\tilde{z}_{s}|\tilde{x}_{s})}[D(\tilde{z}_{s})] \\
         & + \mathbb{E}_{p_{M}(\hat{z}_{s}|\hat{x}_{s})}[(||\bigtriangledown_{\hat{z}_{s}}D(\hat{z}_{s})||_{2}-1)^{2}],
\end{aligned}
\end{equation}
where $p_{M}(z_{s}|x_{s})$, $p_{G,M}(\tilde{z}_{s}|\tilde{x}_{s})$, and $p_{M}(\hat{z}_{s}|\hat{x}_{s})$ are conditional distributions on $x_{s}$, $\tilde{x}_{s}$, and $\hat{x}_{s}$, respectively.\par
\noindent\textbf{Inductive Objective} Based on the above discussion, the final objective of the SRWGAN in the inductive setting is as follows:
\begin{equation}
\begin{aligned}
\mathop{min}_{G,W,M}&\mathop{max}_{D} \;\;  \mathcal{L}^{z}_{wgan} + \lambda_{a}\mathcal{L}_{in-sr} + \lambda_{c}\mathcal{L}_{cls} + \lambda_{m}\mathcal{L}_{m}  \\
s.t. \; &\mathbb{E}_{p_{M}(z^{*}_{s}|x^{*}_{s})}[D_{KL}[p_{M}(z^{*}_{s}|x^{*}_{s})||r(z^{*}_{s})]] \leq b,
\end{aligned}
\end{equation}\par
\noindent\textbf{Transductive Objective} The final transductive objective of the proposed SRWGAN is as follows:
\begin{equation}
\begin{aligned}
\mathop{min}_{G,W,M}&\mathop{max}_{D} \;\;  \mathcal{L}^{z}_{wgan} + \lambda_{a}\mathcal{L}_{tr-sr} + \lambda_{c}\mathcal{L}_{cls} + \lambda_{m}\mathcal{L}_{m}  \\
s.t. \; &\mathbb{E}_{p_{M}(z^{*}_{s}|x^{*}_{s})}[D_{KL}[p_{M}(z^{*}_{s}|x^{*}_{s})||r(z^{*}_{s})]] \leq b,
\end{aligned}
\end{equation}
where $x^{*}_{s}$ denotes either the real or fake seen samples, $G$ is the generator, $W$ represents the weights of the matching function in the hierarchical loss, $M$ is redundancy-free mapping, and $D$ is the discriminator. $\lambda_{a}$, $\lambda_{c}$, and $\lambda_{m}$ are given parameters. Note that the multihead semantic representation is implemented as a submodule of $G$ and that their parameters are optimized together.\par
In the final objective, the first item, \ie, $\mathcal{L}^{z}_{wgan}$, trains a generator for seen classes in an adversarial paradigm, which makes the synthesized features and real features as similar as possible. The second item, \ie, $\mathcal{L}_{sr}$, allows for the generator of seen classes to be applicable for the unseen classes by refining the semantic descriptions towards a bias-eliminated condition. An additional constraint is used to extract clean features and remove redundant information from the visual features. The third item, \ie, $\mathcal{L}_{cls}$, and the last item, \ie, $\mathcal{L}_{m}$, make the generated visual features and clean features more separable and effective for classification. For the transductive setting, $\mathcal{L}_{tr-sr}$ uses the unsupervised entropy loss to learn additional knowledge from unlabeled samples.
\begin{table}[!htb]
\centering
    \textbf{\caption{Statistics of the Six Benchmark Datasets}}
    \setlength{\tabcolsep}{2 mm}
        \vspace{-1em}
    \centering
    \begin{tabular}{cccccccc}
        \Xhline{1pt}
        \multirow{2}*{Dataset} & \multirow{2}*{\#$\mathcal{A}$} & \multicolumn{2}{c}{Class number} & &  \multicolumn{3}{c}{Image number}  \\
        \cline{3-4}         \cline{6-8}
            &     &  \#$\mathcal{Y}_{s}$  &   \#$\mathcal{Y}_{u}$ && Total &  \#$\mathcal{X}^{tr}_{s}$   &   \#$\mathcal{X}^{te}_{s}$/\#$\mathcal{X}^{te}_{u}$ \\
        \hline
        \textbf{CUB}  & 312 & 150 & 50 && 11788 & 7057  & 1764/2967 \\
        \textbf{aPY}  & 64  & 20  & 12 && 15339 & 5932  & 7924/1483 \\ 
        \textbf{AWA}  & 85  & 40  & 10 && 25517 & 19832 & 4958/5685 \\
        \textbf{AWA2} & 85  & 40  & 10 && 37322 & 23527 & 5882/7913 \\
        \textbf{SUN}  & 102 & 645 & 72 && 14340 & 10320 & 2580/1440 \\
        \textbf{FLO}  & 1024& 82  & 20 && 8189  & 5631  & 1403/1155 \\
        \Xhline{1pt}
        \end{tabular}
        \vspace{-2em}
\end{table}

\begin{table}[b]
\centering
    \textbf{\caption{The Grid Search Setting for Hyperparameters}}
    \setlength{\tabcolsep}{4.1 mm}
    \vspace{-1em}
    \centering
    \begin{tabular}{ccccccc}
        \Xhline{1pt}
       Parameters       &     Search Range   \\
        \hline
       $d_{a^{\ddagger}}$, $d_{a^{\dagger}}$,  $d_{{a}'}$   &     \{10, 25, 50, 100, 150, 200, 400, 500\}    \\
       $  \alpha                $    &     \{1e2, 1e1, 1e0, 1e-1, 1e-2, 1e-3, 1e-4, 1e-5\}     \\ 
       $  \beta                 $     &     \{1e2, 1e1, 1e0, 1e-1, 1e-2, 1e-3, 1e-4, 1e-5\}   \\
       $   \delta               $     &     \{1e2, 1e1, 1e0, 1e-1, 1e-2, 1e-3, 1e-4, 1e-5\}   \\
       $  \lambda_{a}      $     &    \{5e0, 1e0,  5e-1, 1e-1, 5e-2, 1e-2, 5e-3, 1e-3\}  \\
       $  \lambda_{c}      $     &     \{5e0, 1e0,  5e-1, 1e-1, 5e-2, 1e-2, 5e-3, 1e-3\}\\
       $  \lambda_{m}     $     &    \{5e0, 1e0,  5e-1, 1e-1, 5e-2, 1e-2, 5e-3, 1e-3\}\\
        \Xhline{1pt}
        \end{tabular}
\end{table}
\begin{table}[b]
\centering
    \textbf{\caption{The Grid Search Results for the GZSL Setting}}
    \setlength{\tabcolsep}{2.3 mm}
    \vspace{-1em}
    \centering
    \begin{tabular}{ccccccc}
        \Xhline{1pt}
       Dataset                                                                       &  \textbf{CUB}   &  \textbf{aPY}   &   \textbf{AWA}  &   \textbf{AWA2}   &   \textbf{SUN}  &   \textbf{FLO}      \\
        \hline
       $d_{a^{\ddagger}}$, $d_{a^{\dagger}}$,  $d_{{a}'}$   &    150                 & 150                  &       200               &     150         &    200        &     200             \\
       $  \alpha                $                                                   &     1e-3               & 1e-1               &      1e-1                &      1e-1        &      1e-2    &      1e-2        \\ 
       $  \beta                 $                                                   &      1e-3               & 1e-3               &      1e-3                &      1e0        &      1e0    &      1e-3        \\ 
       $   \delta               $                                                   &      1e-3               & 1e-1               &      1e-2               &      1e-1        &      1e-2    &      1e-3        \\ 
       $  \lambda_{a}      $                                                   &       1e-3              & 1e-3              &       1e-3                &       1e-3        &       1e-3       &      1e-1           \\
       $  \lambda_{c}      $                                                   &       5e-2              &  1e-2             &       1e-1               &       1e-1      &       1e-2        &       5e-3            \\
       $  \lambda_{m}     $                                                   &      1e-2               &  1e-1             &       1e-2               &       1e-2      &       1e-1        &       1e-3           \\
        \Xhline{1pt}
        \end{tabular}
\end{table}

\begin{table*}[thb]
\centering
    \textbf{\caption{The State-Of-the-Art Results of GZSL and TGZSL}}
    \setlength{\tabcolsep}{1.4 mm}
    \centering
    \begin{tabular}{c|c|ccc|ccc|ccc|ccc|ccc|ccc}    
        \Xhline{1pt}
        \multirow{2}*{\textbf{Setting}} & \multirow{2}*{\textbf{Method}} & \multicolumn{3}{c}{\textbf{CUB}} & \multicolumn{3}{c}{\textbf{aPY}} & \multicolumn{3}{c}{\textbf{AWA}} & \multicolumn{3}{c}{\textbf{AWA2}} & \multicolumn{3}{c}{\textbf{SUN}} & \multicolumn{3}{c}{\textbf{FLO}}\\
                             &  & $U$ & $S$ & $H$           & $U$ & $S$ & $H$                     & $U$ & $S$ & $H$                  & $U$ & $S$ & $H$               & $U$ & $S$ & $H$  & $U$ & $S$ & $H$       \\
        \hline
\multirow{15}*{\textbf{I}} & GXE(2019) \cite{45}       & 47.4 & 47.6 & 47.5 & 26.5 & 74.0 & 39.0   & 62.7 & 77.0 & 69.1    & 56.4 & 81.4 & 66.7 & 36.3 & 42.8 & 39.3 & -- & -- & --  \\
                           & CRNet(2019) \cite{68}      & 45.5 & 56.8 & 50.5 & 32.4 & 68.4 & 44.0   & 58.1 & 74.7 & 65.4    & 58.1 & 78.8 & 65.4 & 34.1 & 36.5 & 35.3 & -- & -- & --  \\
                           & LisGAN(2019) \cite{56}     & 46.5 & 57.9 & 51.6 & --   & --   & --     & 52.6 & 76.3 & 62.3    & --   & --   & --   & 42.9 & 37.8 & 40.2 & 57.7 & 83.8 & 68.3  \\
                           & CADA-VAE(2019) \cite{58}   & 51.6 & 53.5 & 52.4 & --   & --   & --     & 57.3 & 72.8 & 64.1    & 55.8 & 75.0 & 63.9 & 47.2 & 35.7 & 40.6 & --   & --  & --   \\
                           & f-VAEGAN-D2(2019) \cite{21} & 63.2 & 75.6 & {\color{blue}{68.9}} & --   & --   & --     & 57.1 & 76.1 & 65.2    & --   & --   & --   & 50.1 & 37.8 & 43.1 & 63.3 & {\color{blue}{92.4}} & 75.1 \\
                           & GMGAN(2019) \cite{51} & 57.9  & 71.2 & 63.9  & -- & -- & --      & 63.2 & 78.7   & 70.1 & --   & --   & --   & 55.2 & 40.8   & 46.9 & -- & --   & --    \\
                           & IZF(2020) \cite{61}           & 52.7  & 68.0 & 59.4 & 42.3 & 60.5 & 49.8 & 61.3 & 80.5 & 69.6 & 60.6 & 77.5 & 68.0 & 52.7 & {\color{red}{57.0}} & {\color{blue}{54.8}} & -- & -- & --\\
                           & ZSML(2020)\cite{57}       & 60.0 & 52.1 & 55.7  & 36.3 & 46.6 & 40.9 & 57.4 & 71.1 & 63.5 & 58.9 & 74.6 & 65.8 & -- & -- & -- & -- & -- & -- \\
                           & SGV(2020) \cite{69}        & 53.2 & {\color{red}{80.6}} & 64.1 & --   & --   & --     & --   & --   & --      & 57.1 & {\color{red}{93.1}} & 70.8 & 31.4 & 43.3 & 36.4 & -- & -- & --   \\
                           & DVBE(2020) \cite{70}       & {\color{red}{64.4}} & 73.2 & 68.5 & 37.9 & 55.9 & 45.2   & --   & --   & --      & 62.7 & 77.5 & 69.4 & 44.1 & 41.6 & 42.8 & -- & -- & --  \\
                           & DAZLE(2020) \cite{71}      & 56.7 & 59.6 & 58.1 & --  & --  & --       & --   & --   & --      & 60.3  & 75.7  & 67.1  & 52.3  & 24.3  & 33.2  & --  & --  & --  \\
                           & TIZSL(2020) \cite{37}     & 52.1 & 53.3 & 52.7 & 27.8   & 38.7 & 32.4 & 61.5 & 67.7 & 64.4    & {\color{red}{76.8}}  & 66.9   & {\color{blue}{71.5}}   & 32.3 & 24.6 & 27.9 & 70.4 & 68.7 & 69.5   \\
                           & AFRNet(2020) \cite{72}     & --   & --   & --   & {\color{blue}{48.4}} & {\color{blue}{75.1}} & {\color{blue}{58.9}}   & {\color{blue}{68.2}} & 69.4 & 68.8    & 66.7 & 73.8 & 70.1 & 46.6 & 37.6 & 41.5 & --   & --   & --    \\
                           & E-PGN(2020) \cite{60}      & 52.0 & 61.1 & 56.2 & --   & --   & --     & 62.1 & 83.4 & 71.2    & 52.6 & 83.5 & 64.6 & --   & --   & --   & {\color{blue}{71.5}} & 82.2 & {\color{blue}{76.5}}   \\
                           & RFF-GZSL(2020) \cite{55}    & 59.8 &  {\color{blue}{79.9}} & 68.4 & --   & --   & --     & 67.1 & {\color{blue}{91.9}} & {\color{blue}{77.5}}    & --   & --   & --   & {\color{blue}{58.8}} & 45.3 & 51.2 & 62.0 & 91.9 & 74.0 \\
                           & OCD-GZSL(2020) \cite{73}    & 44.8 & 59.9 & 51.3 & --   & --   & --     & --   & --   & --      & 59.5 & 73.4 & 65.7 & 44.8 & 42.9 & 43.8 & --   & -- & --    \\
                           \cline{2-20} 
                           & SRWGAN(Ours)             & {\color{blue}{63.7}} & 78.9 & {\color{red}{70.2}} & {\color{red}{48.6}} & {\color{red}{91.9}} & {\color{red}{63.5}} & {\color{red}{75.0}} & {\color{red}{91.9}} & {\color{red}{82.6}} & {\color{blue}{72.5}} & {\color{blue}{89.4}} & {\color{red}{80.1}} & {\color{red}{65.2}} & {\color{blue}{51.5}} & {\color{red}{57.5}} & {\color{red}{71.7}} & {\color{red}{96.3}} & {\color{red}{82.2}} \\
        \Xhline{1pt}
\multirow{8}*{\textbf{T}}  & GFZSL(2017) \cite{64} & 24.9 & 45.8 & 32.2 & -- & -- & -- & 31.7 & 67.2 & 43.1 & -- & -- & -- & -- & -- & -- & 21.8 & 75.0 & 33.8 \\
                           & DSRL(2017) \cite{74} & 17.3 & 39.0 & 24.0 & -- & -- & -- & 20.8 & 74.7 & 32.6 & -- & -- & -- & 17.7 & 25.0 & 20.7 & 26.9 & 64.3 & 37.9 \\
                           & UE-finetune(2018) \cite{75} & {\color{red}{74.9}} & {\color{blue}{71.5}} & {\color{red}{73.2}} & -- & -- & -- & {\color{red}{93.1}} & 66.2 & 77.4 & -- & -- & -- & 33.6 & {\color{blue}{54.8}} & 41.7 & --   & --   & -- \\
                           & GMGAN(2019) \cite{51} & 60.2 & 70.6 & 65.0 & -- & -- & -- & 70.8 & 79.2 & 74.8 & -- & -- & -- & 57.1 & 40.7 & 47.5 & --   & --   & --   \\
                           & f-VAEGAN-D2(2019) \cite{21} & 61.4 & 65.1 & 63.2 & -- & -- & -- & 84.8 & 88.6 & 86.7 & -- & -- & -- & 60.6 & 41.9 & 49.6 & {\color{red}{78.7}}  & 87.2  & 82.7   \\
                           & GXE(2019) \cite{45}   & 57.0 & 68.7 & 62.3 & -- & -- & --  & {\color{blue}{87.7}} & {\color{blue}{89.0}} & {\color{blue}{88.4}} & 80.2 & {\color{blue}{90.0}} & 84.8 & 45.4 & {\color{red}{58.1}} & 51.0 & --  & -- & -- \\
                           & SDGN(2020) \cite{44}  & {\color{blue}{69.9}} & 70.2 & 70.1 & -- & -- & --  & 87.3 & 88.1 & 87.7 & {\color{red}{88.8}} & 89.3 & {\color{red}{89.1}} & {\color{blue}{62.0}} & 46.0 & {\color{blue}{52.8}} & 78.3 & {\color{blue}{91.4}} & {\color{blue}{84.4}} \\
                           \cline{2-20} 
                           & SRWGAN(Ours)  & 65.3 & {\color{red}{79.1}} & {\color{blue}{71.6}} & {\color{red}{53.6}} & {\color{red}{91.8}} & {\color{red}{67.7}} & 85.6 & {\color{red}{91.4}} & {\color{red}{88.4}} & {\color{blue}{81.2}} & {\color{red}{92.9}} & {\color{blue}{86.6}} & {\color{red}{68.8}} & 51.8 & {\color{red}{59.1}} & {\color{blue}{76.9}} & {\color{red}{96.1}} & {\color{red}{85.4}} \\
        \Xhline{1pt}       
        \multicolumn{19}{l}{``\textbf{I}'' and ``\textbf{T}'' are the inductive and transductive settings, respectively. {\color{red}{Red font}} and {\color{blue}{blue font}} are the highest and second highest results, respectively.}
        \end{tabular}
\vspace{-1em}
\end{table*}

\subsection{Training Algorithm and Tricks}
The algorithm of the proposed SRWGAN is summarized in Algorithm 1. In addition, some tricks used in the model implementation for accuracy improvement are discussed here.\par
\subsubsection{All categories in a batch} The designed semantic refinement loss considers the relationship between the seen and unseen classes. Hence, all classes should be contained in a training batch. We randomly select $N_{bz}/p$ sample pairs for each seen class and $N_{bz}/q$ (refined) semantic descriptions for each unseen class, where $N_{bz}$ is the batch size. Note that none of the real unseen samples are used for inductive training.\par 
\subsubsection{A large batch size} In the design of SRWGAN, the prototype matrices of semantic features and visual features are usually used, such as $U_{a^{\ddagger}s}$ in Eq. (8), $U_{a^{\ddagger}u}$ in Eq. (10),  $U_{a^{\dagger}s}$ in Eq. (14), and $U_{xs}$ in Eq. (18). It is better to set a large batch size for model training, \eg, 1,024 or 2,048, and hence, the prototype matrices can be estimated more accurately at each training batch.
\subsubsection{A warm-up stage} As shown in Algorithm 1, line 11, the semantic loss $\mathcal{L}_{sr}$ is applied after training $G$ for $k$ epochs. Since UBC is implemented with fake unseen samples in Eq. (9), the designed pretraining stage for $G$ warms up the regularization of $\mathcal{L}_{sr}$. For each dataset, different numbers of pretraining epochs are set: 10 (CUB), 5 (aPY), 5 (AWA), 5 (AWA2), 10 (SUN), and 5 (FLO).
\subsubsection{Naive network structure} Instead of the original images, the deep features of a pretrained convolutional neural network are usually used to perform the ZSL, GZSL, and FSL tasks. Hence, we do not need to design the network in a deep learning paradigm. All the modules of the SRWGAN, \eg, the generator and discriminator, are implemented with only one or two layers, which contributes to an efficient training process and a low risk of overfitting. 
\section{Experiments}
\subsection{Experimental Setting and Model Implementation}
\subsubsection{Benchmark datasets} Six benchmark datasets are used for extensive model evaluation, including CUB \cite{65}, aPascal and aYahoo (aPY) \cite{9}, AWA \cite{7}, Animals with Attributes2 (AWA2) \cite{12}, SUN attributes (SUN) \cite{66}, and FLO \cite{36}. CUB is a dataset with a few variations among different birds. aPY is a combined dataset in which the aYahoo dataset is collected from the Yahoo image search, which is different from the images in aPascal. AWA and AWA2 include fifty kinds of animals and are standard datasets for ZSL, GZSL, and FSL. SUN is a subset of the SUN scene dataset with fine-grained attributes. FLO contains 102 kinds of flowers. We use the proposed training/testing splits by Xian \etal \cite{12} and 2,048-dimensional ResNet101 features for experiments. The statistics of the six datasets are summarized in Table I.
\subsubsection{Evaluation metrics} Model performance is evaluated according to the top-1 per-class accuracy. For GZSL, we report the average top-1 per-class accuracy on the seen classes as $S$ and on the unseen classes as $U$. Their harmonic mean, \ie, $H = (2 \times S \times U)/(S + U)$, is reported as a comprehensive index. For FSL, we also use $S$, $U$, and $H$ to show the accuracy improvement by a few unseen samples in comparison with GZSL. For ZSL, we use the average top-1 per-class accuracy on the unseen classes as a metric.
\subsubsection{Implementation details} The proposed SRWGAN consists of a generator $G$, a discriminator $D$, and redundancy-free mapping $M$. First, the generator $G$ has a 4,096-dimensional hidden layer with the $LeakyReLU(0.2)$ function and a 2,048-dimensional output layer with the $ReLU$ function. The discriminator $D$ is a fully connected layer without activation. The redundancy-free mapping $M$ is a 1,024-dimensional layer with the $LeakyReLU(0.2)$ function, which is implemented with the reparameterization trick. The margin $\Delta$ and bound $b$ of $M$ are set to 190 and 0.1, respectively. Additionally, the Adam optimizer \cite{67} is adopted with $\beta_{1} = 0.5$ and $\beta_{2} = 0.999$ for all the modules. The learning rate is $0.0001$.\par
\subsubsection{Parameter setting} For the proposed multihead semantic representation module and the hierarchical semantic alignment module, the essential hyperparameters include the dimension of the refined semantic descriptions $d_{a^{\ddagger}} = d_{a^{\dagger}} = d_{{a}'}$, the hierarchical weights $\alpha$, $\beta$, $\delta$ in Eq. (6), and the regularization weights $\lambda_{a}$, $\lambda_{c}$, $\lambda_{m}$ in Eq. (21). We perform a detailed grid search in the GZSL setting for these hyperparameters on the six benchmark datasets, which is shown in Table II. The search results for the six benchmark datasets are listed in Table III. In addition, the training batch sizes for the six datasets are: 2048 (CUB), 1024 (aPY), 2048 (AWA), 2048 (AWA2), 800 (SUN), and 2048 (FLO).

\begin{figure}[!htb]
\centering 
\subfigure[GZSL on AWA]{
\includegraphics[width=4cm]{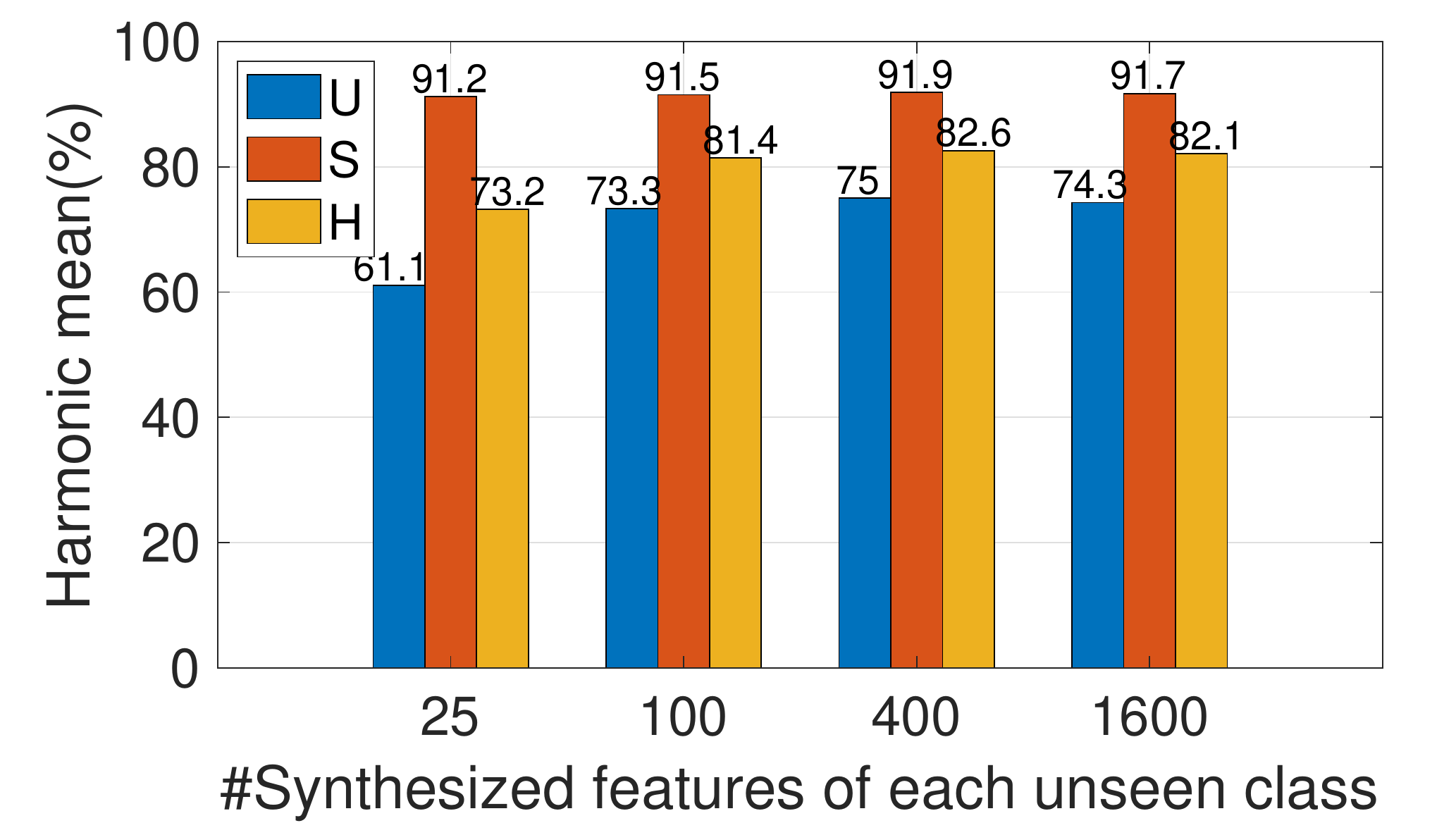}}
\subfigure[GZSL on FLO]{
\includegraphics[width=4cm]{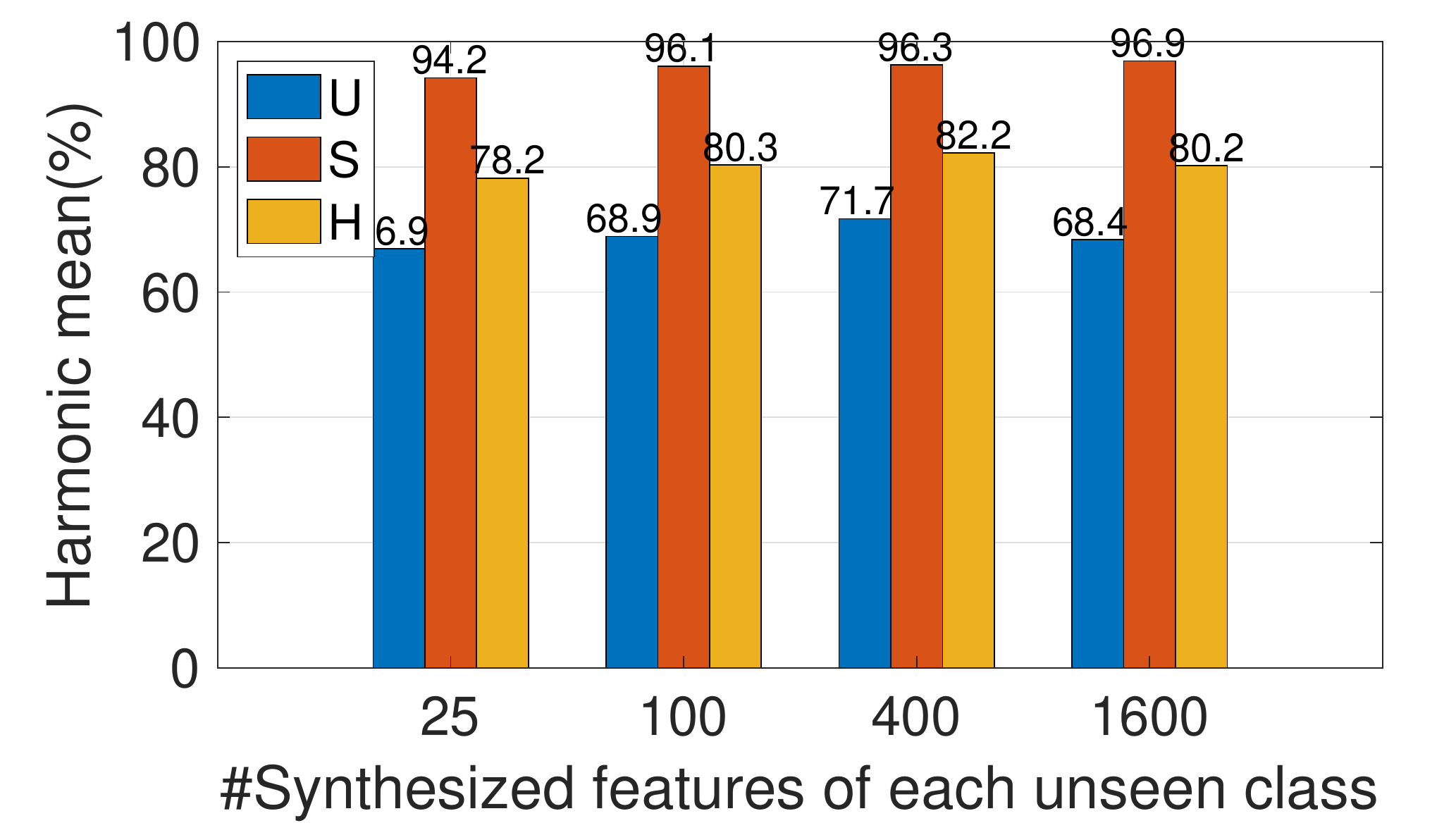}}
\subfigure[TGZSL on AWA]{
\includegraphics[width=4cm]{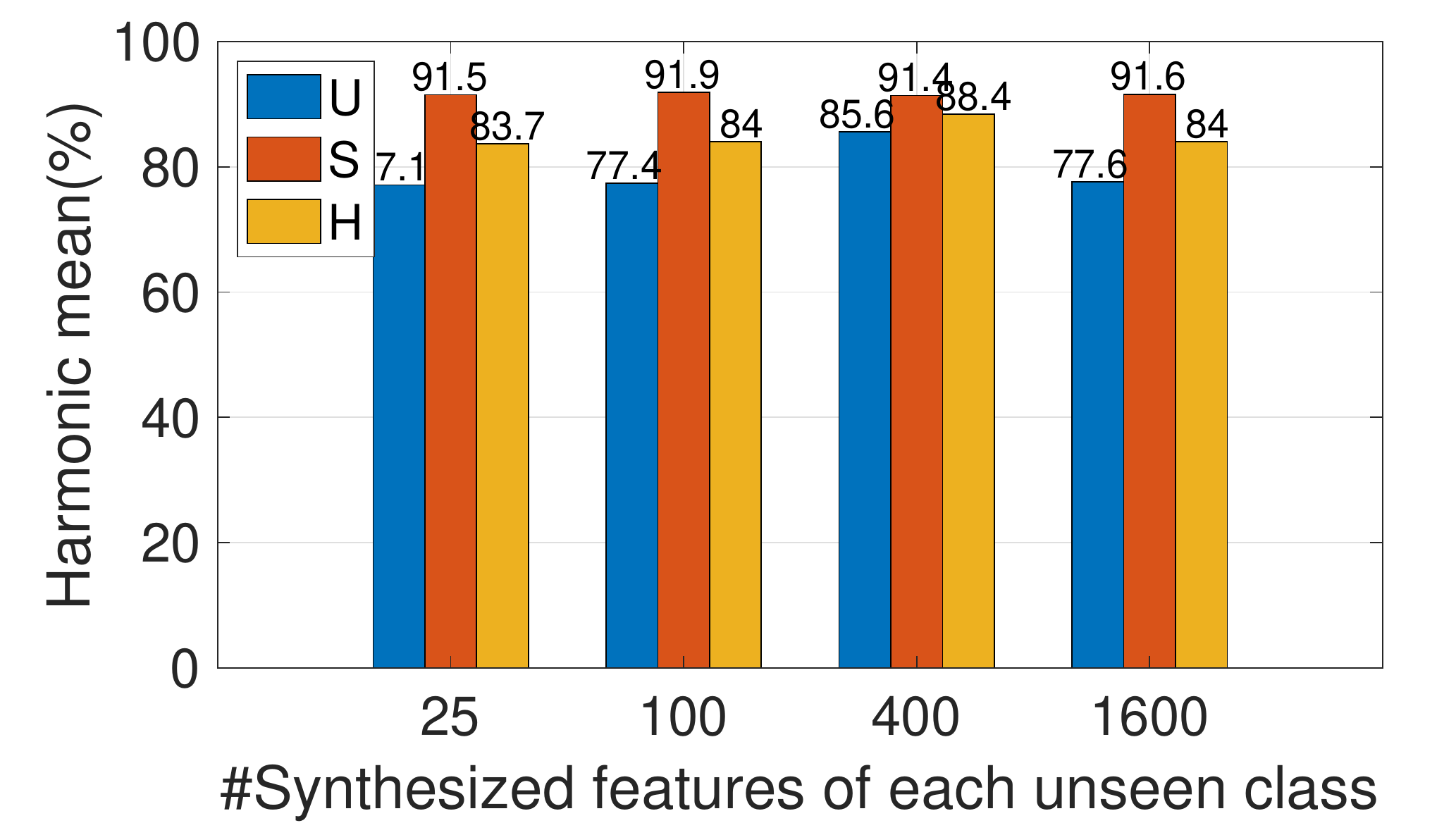}}
\subfigure[TGZSL on FLO]{
\includegraphics[width=4cm]{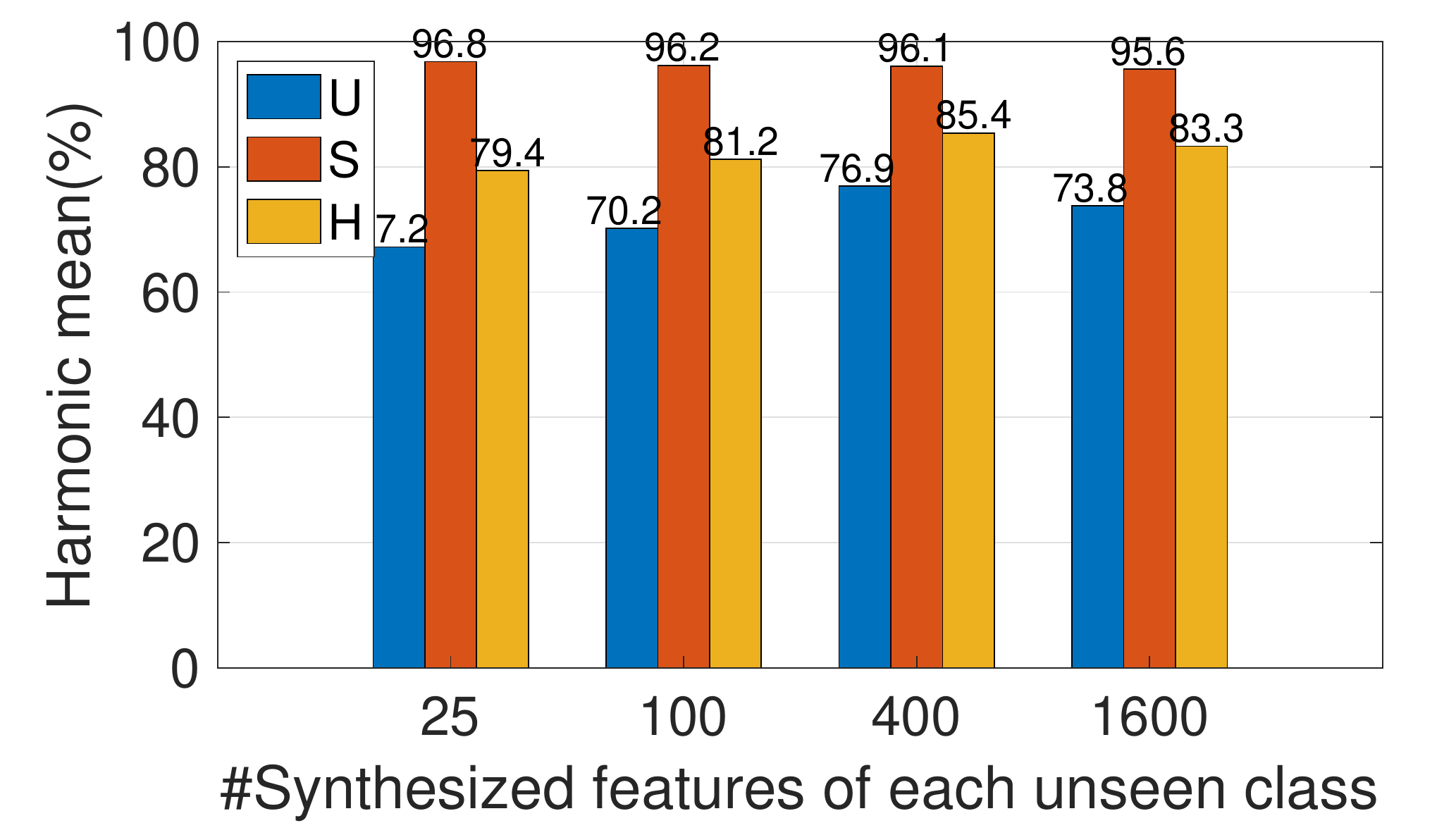}}
\caption{The GZSL and TGZSL results of the proposed SRWGAN with respect to different numbers of synthesized samples per unseen class.}
\end{figure}
\begin{table*}[thb]
\centering
    \textbf{\caption{The State-Of-the-Art Results of FSL}}
    \vspace{-0.5em}
    \setlength{\tabcolsep}{1.5 mm}
    \centering
    \begin{tabular}{c|c|ccc|ccc|ccc|ccc|ccc|ccc}    
        \Xhline{1pt}
        \multirow{2}*{\textbf{Setting}} & \multirow{2}*{\textbf{Method}} & \multicolumn{3}{c}{\textbf{CUB}} & \multicolumn{3}{c}{\textbf{aPY}} & \multicolumn{3}{c}{\textbf{AWA}} & \multicolumn{3}{c}{\textbf{AWA2}} & \multicolumn{3}{c}{\textbf{SUN}} & \multicolumn{3}{c}{\textbf{FLO}}\\
                             &  & $U$ & $S$ & $H$           & $U$ & $S$ & $H$                     & $U$ & $S$ & $H$                  & $U$ & $S$ & $H$               & $U$ & $S$ & $H$  & $U$ & $S$ & $H$       \\
        \hline
\multirow{5}*{\textbf{0-Shot$^{*}$}} & f-CLSWGAN(2018)  & 43.7 & 57.7 & 49.7 & -- & -- & --  & 57.9 & 61.4 & 59.6 & 52.1 & 68.9 & 59.4 & 42.6 & 36.6 & 39.4 & 59.0 & 73.8 & 65.6 \\
                               & GMGAN(2019)      & 57.9  & 71.2 & 63.9  & -- & -- & --      & 63.2 & 78.7   & 70.1 & --   & --   & --   & 55.2 & 40.8   & 46.9 & -- & --   & --   \\
                               & RFF-GZSL(2020)   & 59.8 & 79.9 & 68.4 & -- & --   & --      & 67.1 & 91.9 & 77.5 & --   & --   & --   & 58.8 & 45.3 & 51.2 & 62.0 & 91.9 & 74.0 \\
                               & LsrGAN(2020)     & 48.1 & 59.1 & 53.0 & -- & -- & --  & 52.6 & 76.3 & 62.3 & -- & -- & --   & 44.8 & 37.7 & 40.9 & -- & -- & --   \\
                               \hline  
\multirow{5}*{\textbf{0-Shot}} & f-CLSWGAN(2018)  & 41.4 & 57.3 & 48.1 & 14.9 & 81.4 & 25.2 & 57.3 & 61.5 & 59.3 & 51.3 & 70.4 & 59.3 & 39.3 & 38.6 & 38.9 & 59.1 & 71.5 & 64.7 \\
                               & GMGAN(2019)      & 57.8 & 71.5 & 63.9   & 31.2 & 86.0 & 45.8      & 60.4 & 81.1 & 69.2 & 54.8   & 86.0   & 67.0  & 53.9 & 41.7 & 47.1  & {\color{blue}{59.9}} & 87.0   & 71.0     \\
                               & RFF-GZSL(2020)   & {\color{blue}{58.3}} & {\color{red}{81.9}} & {\color{blue}{68.1}} & 29.6 & {\color{red}{92.9}} & 44.8 & {\color{blue}{69.8}} & {\color{blue}{90.0}} & {\color{blue}{78.6}} & {\color{blue}{67.8}} & {\color{red}{91.7}} & {\color{blue}{78.0}} & {\color{blue}{58.9}} & {\color{blue}{45.2}} & {\color{blue}{51.1}} & 59.5 & {\color{blue}{93.8}} & {\color{blue}{72.8}} \\
                               & LsrGAN(2020)     & 46.2 & 61.3 & 52.7 & {\color{blue}{32.0}} & 85.9 & {\color{blue}{46.6}} & 51.9 & 79.7 & 62.9 & 45.9 & 85.6 & 59.8 & 43.7 & 37.9 & 40.6 & 58.5 & 86.0 & 69.7  \\
                               \cline{2-20}
                               & SRWGAN(Ours)      & {\color{red}{63.7}} & {\color{blue}{78.9}} & {\color{red}{70.2}} & {\color{red}{48.6}} & {\color{blue}{91.9}} & {\color{red}{63.5}} & {\color{red}{75.0}} & {\color{red}{91.9}} & {\color{red}{82.6}} & {\color{red}{72.5}} & {\color{blue}{89.4}} & {\color{red}{80.1}} & {\color{red}{65.2}} & {\color{red}{51.5}} & {\color{red}{57.5}} & {\color{red}{71.7}} & {\color{red}{96.3}} & {\color{red}{82.2}}  \\
                               \hline      
\multirow{5}*{\textbf{1-Shot}} & f-CLSWGAN(2018)  & 43.0 & 55.7 & 48.6 & 17.7 & 75.9 & 28.7 & 59.2 & 61.8 & 60.5 & 55.4 & 72.1 & 62.7 & 41.7 & 40.2 & 40.9 & 59.8 & 72.4 & 66.5  \\
                               & GMGAN(2019)      & {\color{blue}{62.0}}  & 72.8 & 67.0   & {\color{blue}{44.3}} & 86.0 & {\color{blue}{58.5}}      & 65.1 & 81.3 & 72.3 & 59.5   & 85.4   & 70.1 & 55.0 & 43.8 & 48.7   & 63.3 & 86.3   & 73.0    \\
                               & RFF-GZSL(2020)   & 59.5 & {\color{red}{85.0}} & {\color{blue}{70.0}} & 33.5 & {\color{blue}{91.5}} & 49.1 & {\color{blue}{74.3}} & {\color{blue}{90.6}} & {\color{blue}{81.7}} & {\color{blue}{77.4}} & {\color{blue}{92.4}} & {\color{blue}{84.2}} & {\color{blue}{62.3}} & {\color{blue}{47.4}} & {\color{blue}{53.8}} & 63.9 & {\color{red}{96.1}} & {\color{blue}{76.7}}  \\
                               & LsrGAN(2020)      & 46.3 & 62.7 & 53.2 & 35.1 & 86.0 & 49.8 & 58.8 & 81.2 & 68.2 & 53.2 & 85.9 & 65.7 & 46.6 & 41.6 & 44.0 & {\color{blue}{67.0}} & 82.4 & 73.9  \\
                               \cline{2-20}
                               & SRWGAN(Ours)      & {\color{red}{65.5}} & {\color{blue}{80.4}} & {\color{red}{72.2}} & {\color{red}{50.2}} & {\color{red}{91.7}} & {\color{red}{64.9}} & {\color{red}{86.0}} & {\color{red}{91.5}} & {\color{red}{88.6}} & {\color{red}{80.7}} & {\color{red}{92.4}} & {\color{red}{86.1}} & {\color{red}{66.5}} & {\color{red}{51.7}} & {\color{red}{58.2}} & {\color{red}{74.6}} & {\color{blue}{95.9}} & {\color{red}{83.9}} \\
                               \hline    
\multirow{5}*{\textbf{3-Shot}} & f-CLSWGAN(2018)  & 47.6 & 60.3 & 53.2 & 27.1 & 78.4 & 40.2 & 65.9 & 60.3 & 63.0 & 60.4 & 73.3 & 66.2 & 45.8 & 39.1 & 42.2 & 69.5 & 72.4 & 70.9  \\
                               & GMGAN(2019)      & 63.2  & 71.4 & 67.1   & 50.0 & 86.0 & 63.3      & 67.8 & 81.1   & 73.8 & 72.5 & 85.6 & 78.5 & 59.0 & 43.2 & 49.9   & 74.6 & 87.6   & 80.6     \\
                               & RFF-GZSL(2020)   & {\color{blue}{63.6}} & {\color{red}{85.2}} & {\color{blue}{72.8}} & 38.2 & {\color{blue}{91.3}} & 53.9 & {\color{blue}{84.5}} & {\color{blue}{90.4}} & {\color{blue}{87.4}} & {\color{blue}{82.1}} & {\color{blue}{92.4}} & {\color{blue}{87.0}} & {\color{blue}{65.3}} & {\color{blue}{46.8}} & {\color{blue}{54.5}} & 74.4 & {\color{red}{96.6}} & {\color{blue}{84.1}}  \\
                               & LsrGAN(2020)      & 50.2 & 62.3 & 55.7 & {\color{red}{53.5}} & 86.0 & {\color{blue}{66.0}} & 64.3 & 81.9 & 72.1 & 67.2   & 85.9  & 75.4 & 49.3 & 42.8 & 45.8 &  {\color{blue}{79.4}} & 82.7 & 81.0  \\
                               \cline{2-20}
                               & SRWGAN(Ours)      & {\color{red}{66.9}} & {\color{blue}{80.1}} & {\color{red}{72.9}} & {\color{blue}{53.4}} & {\color{red}{92.0}} & {\color{red}{67.5}} & {\color{red}{89.3}} & {\color{red}{91.5}} & {\color{red}{90.4}} & {\color{red}{83.7}} & {\color{red}{92.8}} & {\color{red}{88.1}} & {\color{red}{67.5}} & {\color{red}{51.5}} & {\color{red}{58.4}} & {\color{red}{82.6}} & {\color{blue}{94.6}} & {\color{red}{88.2}} \\
                               \hline   
\multirow{5}*{\textbf{5-Shot}} & f-CLSWGAN(2018)  & 49.0 & 62.7 & 55.0 & 32.7 & 82.0 & 46.8 & 70.3 & 62.1 & 65.9 & 71.3 & 73.9 & 72.6 & 52.7 & 40.0 & 45.5 & 80.6 & 73.2 & 76.7  \\
                               & GMGAN(2019)      & 67.1  & 71.3 & 69.1   & {\color{blue}{57.6}} & 85.9 & 69.0      & 79.0 & 81.2   & 80.1 & 77.2   & 85.9  & 81.4 & 60.0 & 43.6   & 50.5    & 78.8 & 87.0   & 82.7   \\
                               & RFF-GZSL(2020)   & {\color{blue}{68.0}} & {\color{red}{84.3}} & {\color{blue}{75.3}} & 48.5 & {\color{blue}{89.8}} & 63.0 & {\color{red}{96.7}} & {\color{blue}{90.6}} & {\color{red}{93.6}} & {\color{blue}{94.3}} & {\color{blue}{92.2}} & {\color{blue}{93.2}} & {\color{blue}{71.3}} & {\color{blue}{47.9}} & {\color{blue}{57.3}} & 82.4 & {\color{red}{96.0}} & {\color{blue}{88.7}}  \\
                               & LsrGAN(2020)      & 54.0 & 62.3 & 57.8 & {\color{red}{62.8}} & 85.8 & {\color{red}{72.5}} & 80.2 & 81.0 & 80.6 & 82.5 & 85.6 & 84.0 & 52.6 & 42.9 & 47.2 & {\color{blue}{84.7}} & 82.5 & 83.6  \\
                               \cline{2-20}
                               & SRWGAN(Ours)      & {\color{red}{73.9}} & {\color{blue}{81.1}} & {\color{red}{77.3}} & 56.3 & {\color{red}{91.9}} & {\color{blue}{69.8}} & {\color{blue}{95.4}} & {\color{red}{91.5}} & {\color{blue}{93.4}} & {\color{red}{97.1}} & {\color{red}{92.2}} & {\color{red}{94.6}} & {\color{red}{73.8}} & {\color{red}{50.5}} & {\color{red}{60.0}} & {\color{red}{86.3}} & {\color{blue}{95.3}} & {\color{red}{90.6}} \\
                               \Xhline{1pt}    
        \multicolumn{19}{l}{$^{*}$ denotes the results are from the original papers, and others are obtained by our implementation.}
        \end{tabular}
\end{table*}

\begin{figure}[!htb]
\vspace{-0.5em}
\centering 
\subfigure[One-shot on CUB]{
\includegraphics[width=8cm]{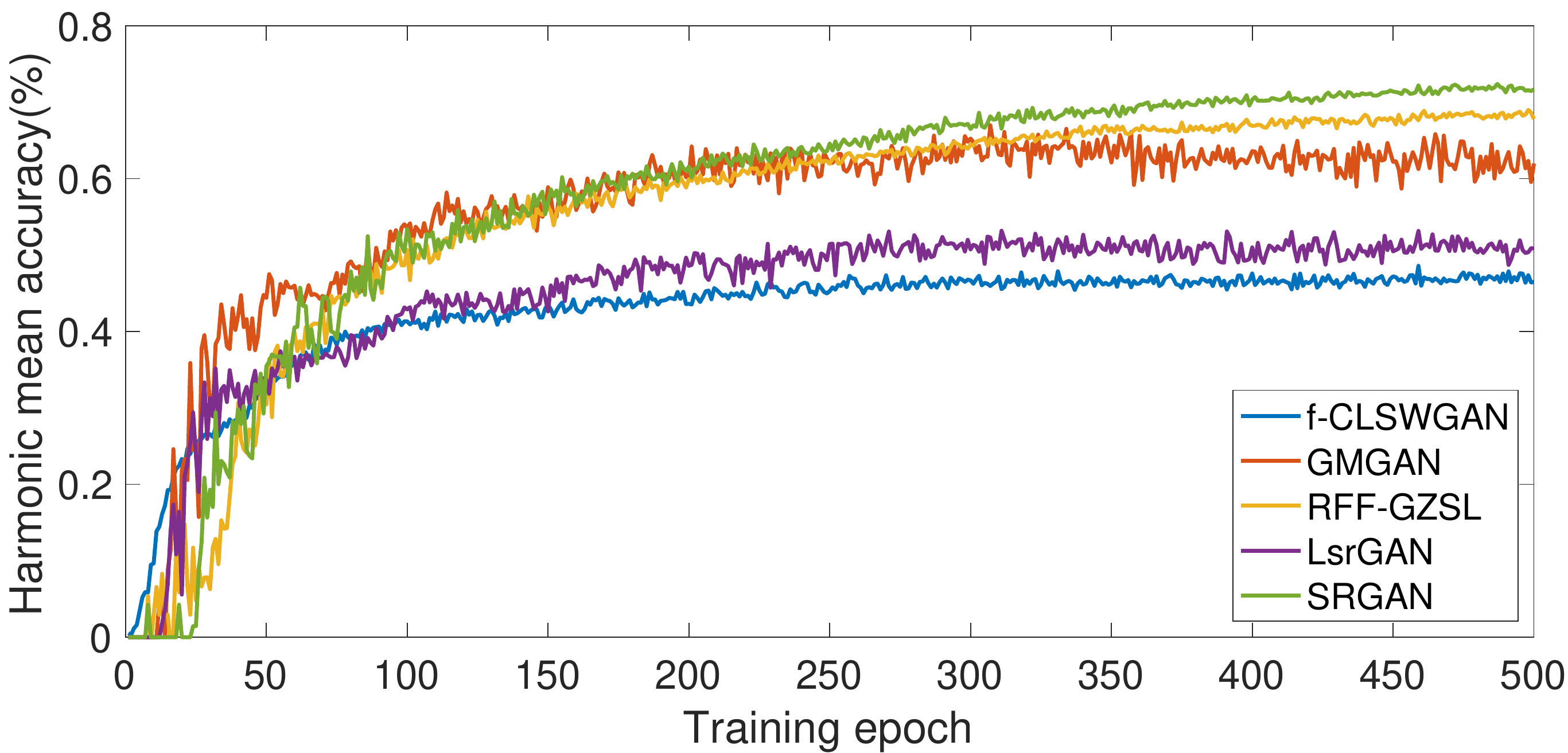}}
\subfigure[Five-shot on CUB]{
\includegraphics[width=8cm]{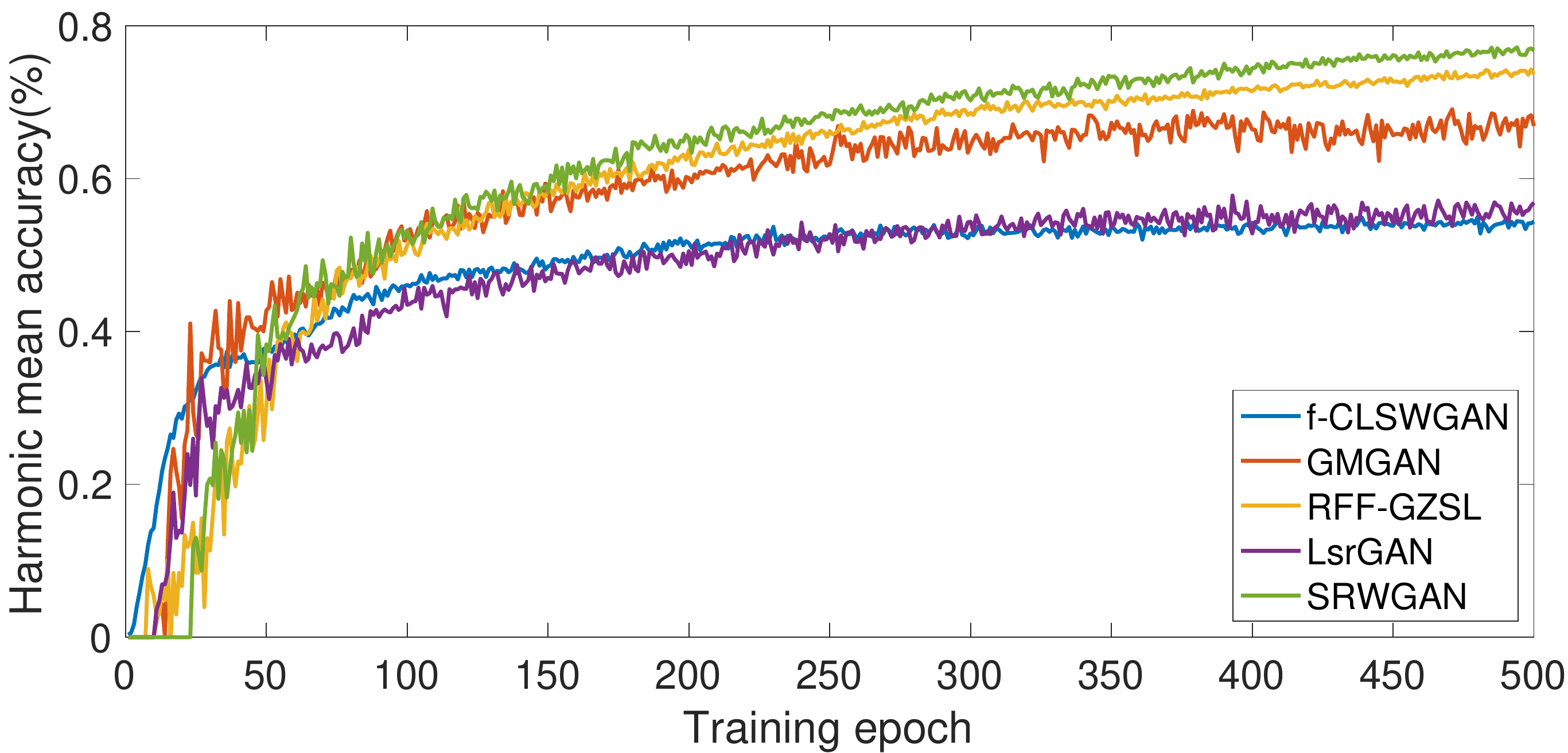}}
\caption{The effects of the number of training epochs on the five compared methods.}
\vspace{-0.5em}
\end{figure}

\subsection{Evaluation in a GZSL setting}
Sixteen and seven strong baselines published in recent years are used for the comparison of the GZSL and TGZSL tasks, respectively. The results are referred from their papers. The proposed SRWGAN is used to generate fake samples for the unseen classes, which are combined with the seen samples to train a softmax classifier. The well-trained classifier can then be used to infer the labels for test data. For each dataset, the number of synthesized samples per unseen class is different: 800 (CUB), 400 (aPY), 400 (AWA), 4000 (AWA2), 600 (SUN), and 800 (FLO). The state-of-the-art results are compared in Table \uppercase\expandafter{\romannumeral4}.\par
As shown, the proposed SRWGAN obtains the highest harmonic mean accuracy for all six benchmark datasets. Specifically, our approach achieves harmonic mean accuracy improvements of 1.3\%, 4.6\%, 5.1\%, 8.6\%, 6.3\%, and 5.7\% over state-of-the-art methods on CUB, aPY, AWA, AWA2, SUN, and FLO, respectively. The harmonic mean accuracies of SRWGAN on AWA, AWA2, and FLO exceed 80\%, which means that the proposed SRWGAN has almost the same performance of a normal image classification task for the more challenging GZSL. In addition, we can observe that the proposed SRWGAN not only improves the accuracy of unseen classes but also improves the accuracy of seen classes, confirming the effectiveness of the prototype matrix-based classification loss, redundancy-free mapping, and tricks in SRWGAN. In the transductive setting, the proposed SRWGAN has the highest harmonic mean accuracy on aPY, AWA, SUN, and FLO and the second highest harmonic mean accuracy on CUB and AWA2. The satisfactory performance validates the effectiveness of the unsupervised auxiliary alignment loss in the proposed hierarchical alignment module.\par
In Figure 4, the results of SRWGAN under different numbers of synthesized samples per unseen class on AWA and FLO for GZSL and TGZSL are shown. Intuitively, the results for the seen classes are steady for both GZSL and TGZSL and are hardly affected by the number of synthesized unseen samples. However, the accuracy of unseen classes is greatly affected by the number of synthesized unseen samples. The generation of a few samples for unseen classes leads to an imbalanced data problem and degrades the performance on unseen classes. For example, the unseen accuracy of FLO is 66.9\%, with 25 samples generated for each unseen class. However, the unseen accuracy is 68.4\%, with 1600 samples generated. Empirically, the proper number of synthesized samples per unseen class should be larger than 400.\par

\begin{table}[thb]
\centering
    \textbf{\caption{Validation for the Proposed Semantic Refinement Techniques}}
    \vspace{-0.5em}
    \setlength{\tabcolsep}{1.8 mm}
    \centering
    \begin{tabular}{c|cccc|cccc}    
        \Xhline{1pt}
        \multirow{2}*{\textbf{Method}} & \multicolumn{4}{c}{\textbf{FLO}}        &  \multicolumn{4}{c}{\textbf{CUB}}  \\
                                                        & $U$ & $S$ & $H$  & $RI $                & $U$ & $S$ & $H$ & $RI $     \\
        \hline
        Model A           & 71.7 & 96.3 & 82.2 & 0.00\%                            & 63.7  & 78.9 & 70.2 & 0.00\%   \\
        Model B           & 70.6 & 96.2 & 81.4 & -0.97\%                           & 62.0  & 77.8 & 69.0  & -1.71\%      \\
        Model C           & 70.9 & 96.3 & 81.7  & -0.61\%                         & 64.4 & 76.2 & 69.8 & -0.57\%     \\
        Model D           & 70.4 & 96.1 & 81.2  & -1.21\%                         & 64.3 & 75.9 & 69.6  & -0.85\%     \\
        Model E           & 68.7 & 95.8 & 80.0 & -2.67\%                          & 64.1 & 74.1 & 68.7 & -2.14\%     \\ 
        Model F           & 68.4 & 95.9 & 79.8 & -2.91\%                          & 63.1 & 75.3 & 68.7 & -2.14\%     \\
        Model G          & 66.4 & 94.8 & 78.1  & -4.98\%                        & 63.1 & 74.4 & 68.3  & -2.71\%     \\
        \Xhline{1pt}       
        \end{tabular}
\vspace{-1em}
\end{table}

\begin{figure}[htb]
\centering
\includegraphics[width=.45\textwidth]{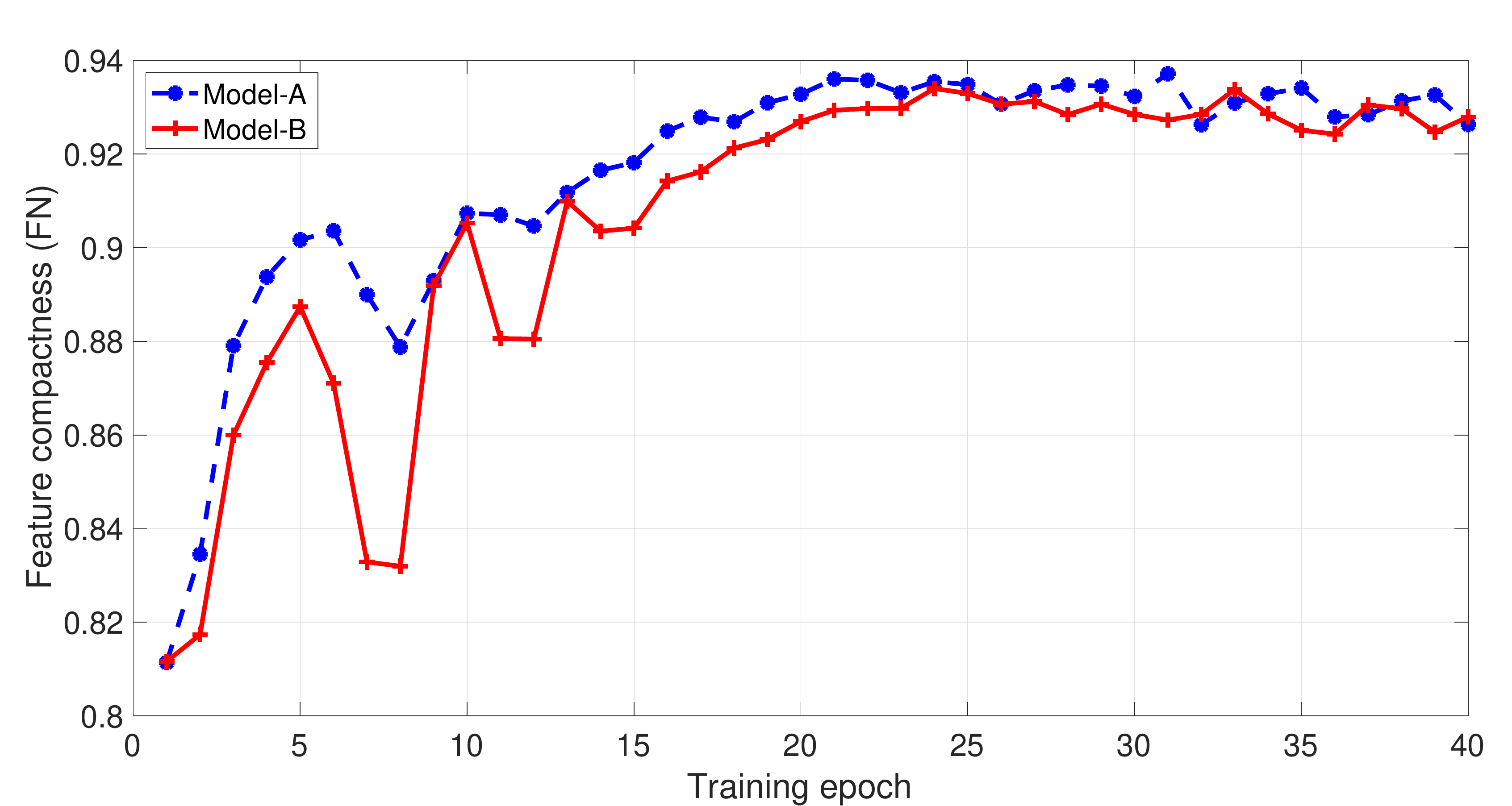} 
\caption{Comparison of seen feature compactness between Model A and Model B on CUB.}
\end{figure}

\begin{figure}[htb]
\centering
\includegraphics[width=.45\textwidth]{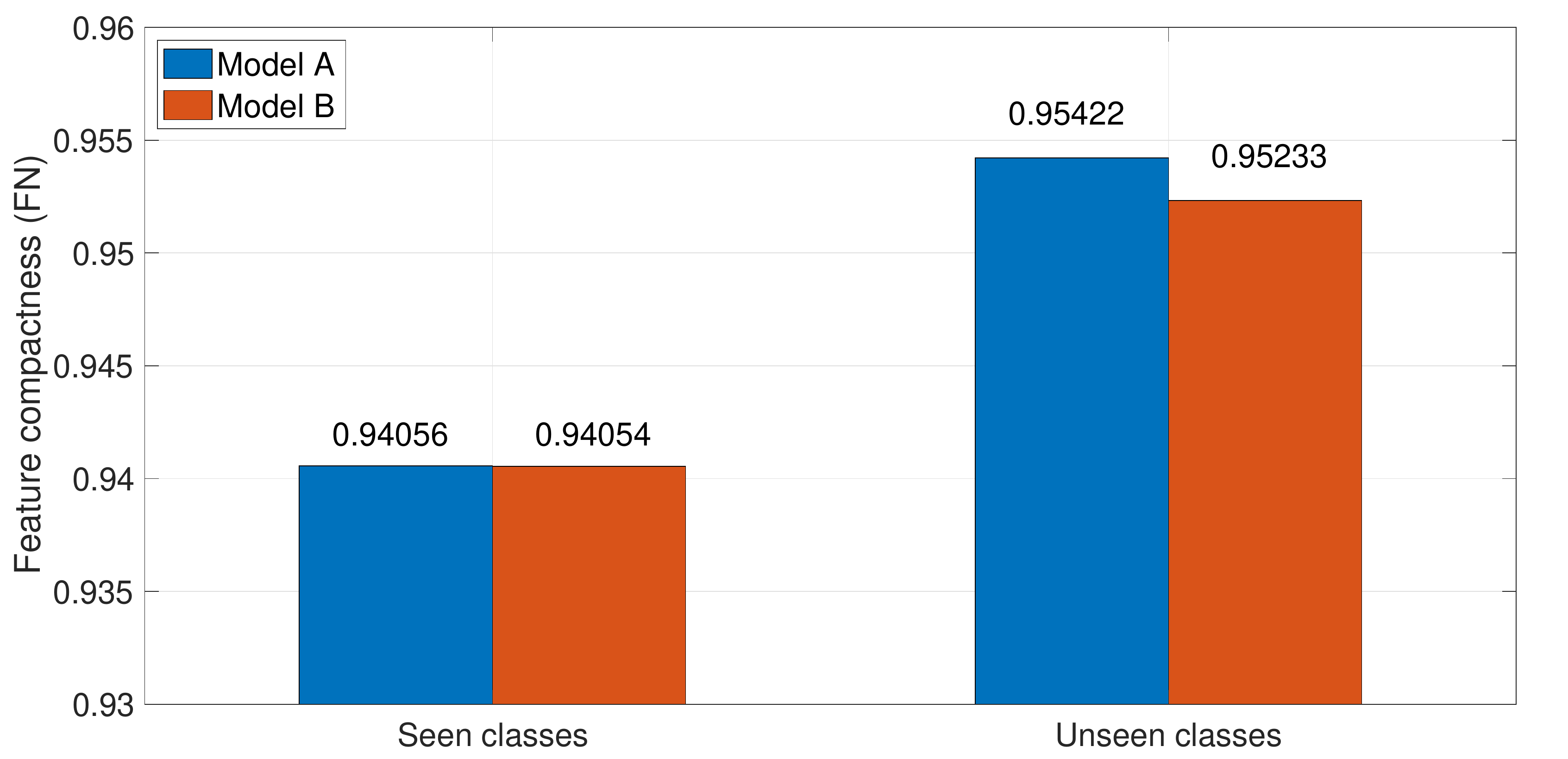} 
\caption{Comparison of the seen and unseen average feature compactness of 1200 epochs (from the 300th epoch to the 1500th epoch) between Model A and Model B on CUB.}
\vspace{-1em}
\end{figure}

\subsection{Evaluation in an FSL Setting}
Here, the proposed SRWGAN is compared with f-CLSWGAN \cite{53}, GMGAN \cite{51}, RFF-GZSL \cite{55}, and LsrGAN \cite{62} on the FSL task, where a few labeled samples of unseen classes are allowed in the training phase. The implementations of f-CLSWGAN and RFF-GZSL were contributed by their authors, and GMGAN and LsrGAN were implemented. For a fair comparison, we first fine-tune the model implementations and parameters in the GZSL setting and then apply them in the FSL setting. For the five compared models, the few labeled samples of unseen classes are the same and used in each batch training with the labeled seen samples together. The results of FSL on the six benchmark datasets are shown in Table \uppercase\expandafter{\romannumeral5}.\par
First, in the few-shot settings, \ie, one-shot, three-shot, and five-shot learning settings, performance improvement is observed on all the datasets and for all the methods. However, the increase in the number of unseen samples has different effects on the six datasets. For example, the results of five-shot learning have an approximately 10\% to 15\% harmonic mean accuracy improvement on the AWA dataset in comparison with the results of one-shot learning, while the accuracy improvement on the SUN dataset is only 5\% to 7\%. This result may be caused by the numerous unseen classes in the SUN dataset. Similarly, the CUB dataset, which has fifty kinds of unseen classes, has only a 5\% to 10\% accuracy improvement, while the APY, AWA2, and FLO datasets, which have fewer unseen classes, have more significant accuracy improvements in the FSL setting.\par
Second, it can be observed that the proposed SRWGAN obtains the highest accuracy in most of the cases. However, the larger the number of provided unseen samples is, the smaller the advantages of the proposed method. As shown in Table \uppercase\expandafter{\romannumeral5}, the SRWGAN obtains the highest harmonic mean accuracy for all six benchmark datasets in the one-shot and three-shot learning settings. However, the highest accuracies are obtained for four benchmark datasets in the five-shot learning setting. The LsrGAN obtains the highest accuracy for the aPY dataset, and the RFF-GZSL obtains the highest accuracy for the AWA dataset in the five-shot learning setting. This is because the designed SRWGAN seeks to eliminate the bias in the disjoint-class generator transfer by semantic refinement techniques. When numerous unseen samples are provided, the bias problem is nonexistent, and the generator for unseen classes is directly trained. Hence, the SRWGAN has fewer advantages in settings with sufficient data. In contrast, the results show the effectiveness of the proposed method in cases of zero samples and few samples.\par
Finally, we show the effects of the number of training epochs on the CUB dataset in the FSL setting in Figure 5. In comparison, the SRWGAN and RFF-GZSL present steadier and higher harmonic mean accuracy than do the other three compared models with redundancy-free features. The SRWGAN shows better performance than RFF-GZSL based on the proposed semantic refinement techniques.
\begin{figure*}[!htb]
\centering 
\subfigure[Features for AWA by SRWGAN]{
\includegraphics[width=5.7cm]{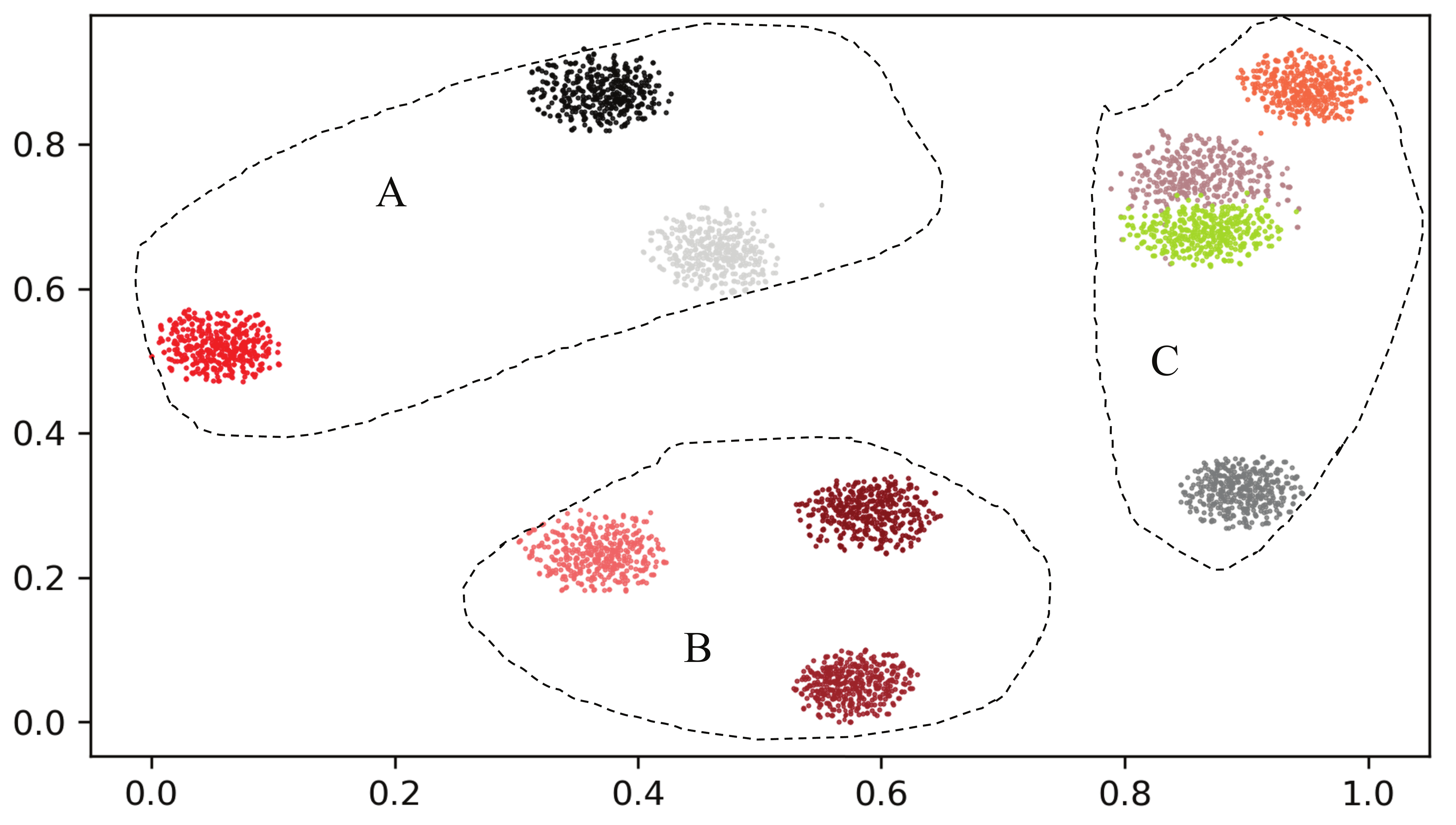}}
\subfigure[Real features of AWA]{
\includegraphics[width=5.7cm]{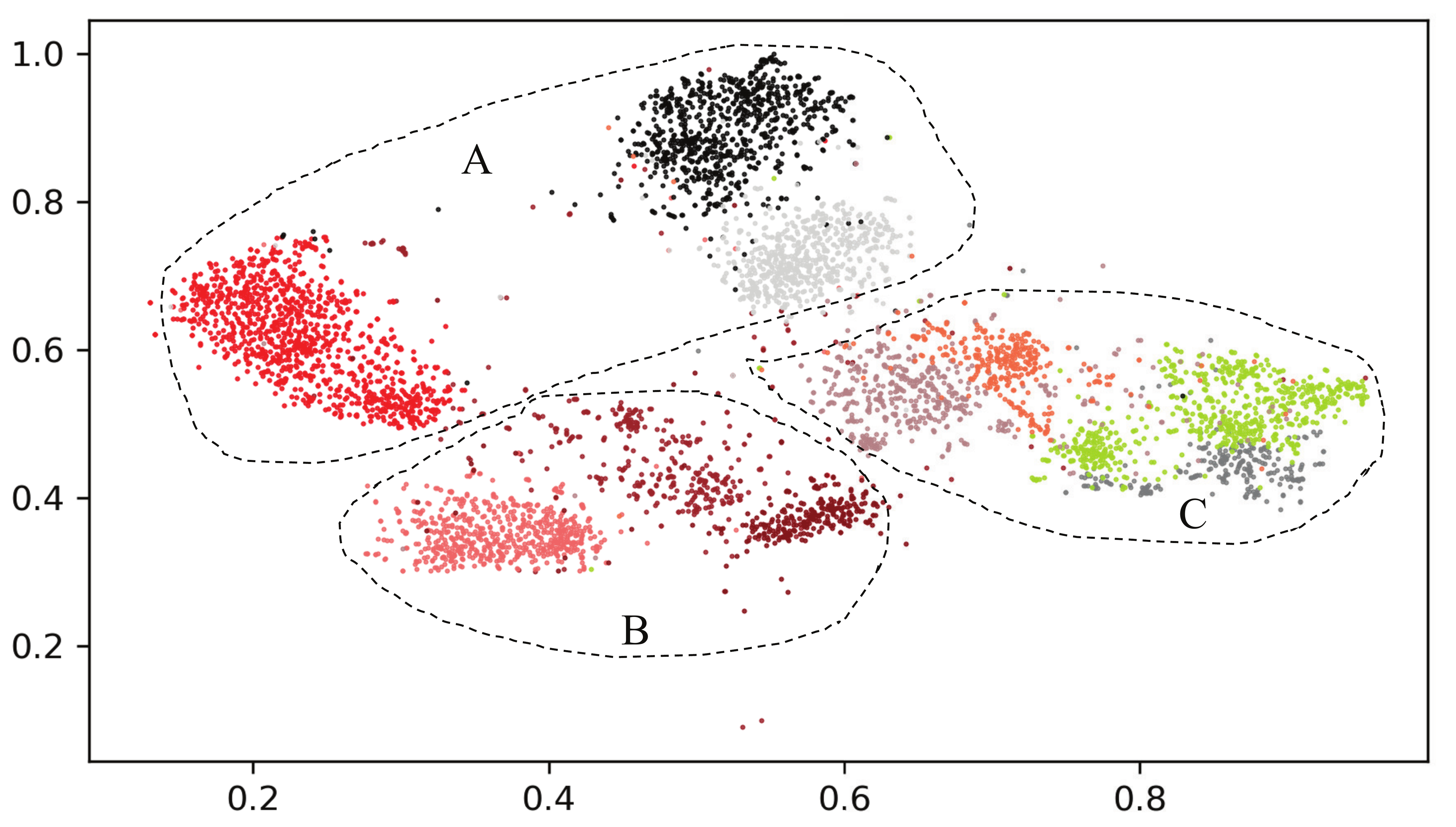}}
\subfigure[Features for AWA by f-CLSWGAN]{
\includegraphics[width=5.7cm]{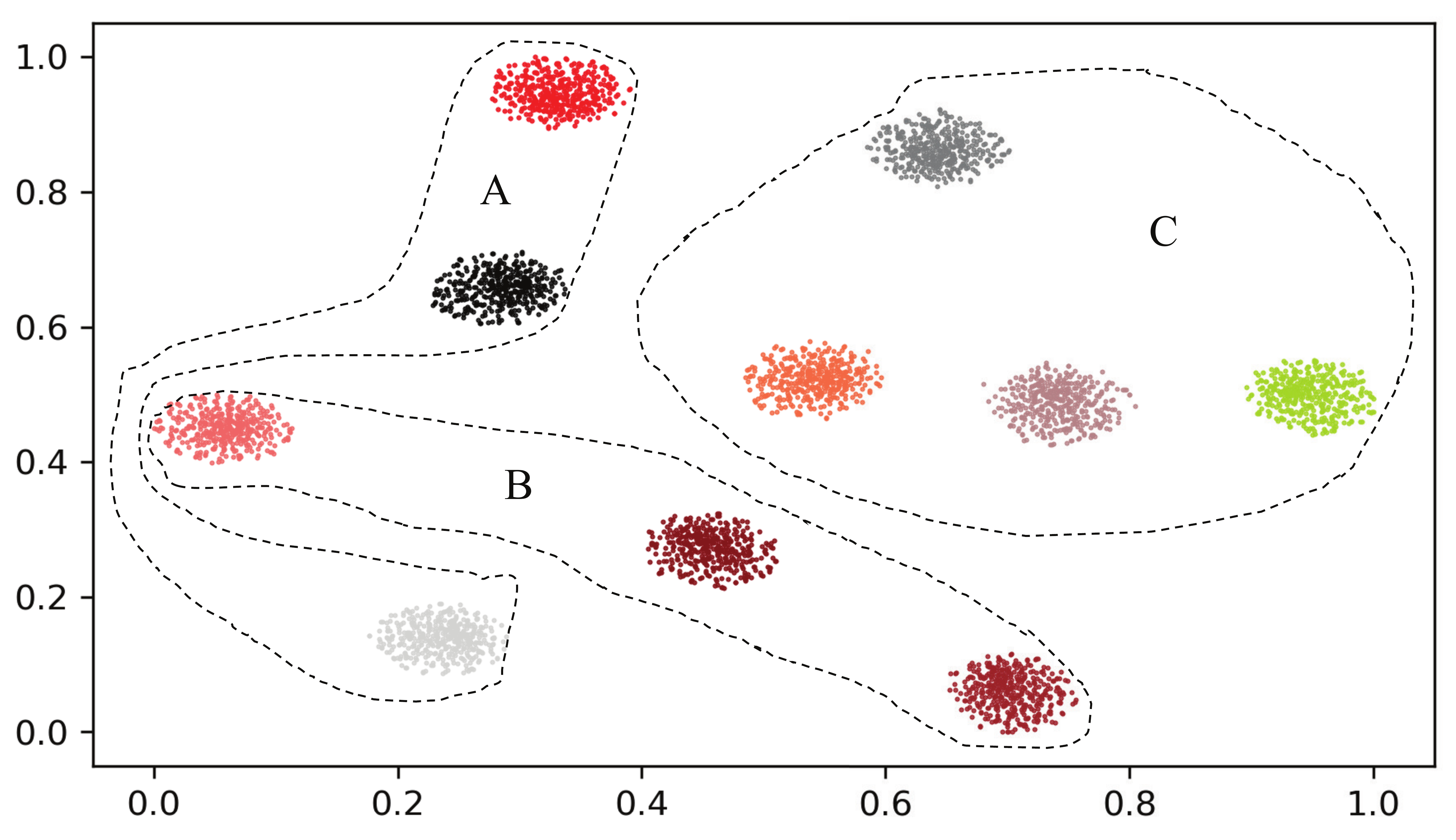}}
\subfigure[Features for FLO by SRWGAN]{
\includegraphics[width=5.7cm]{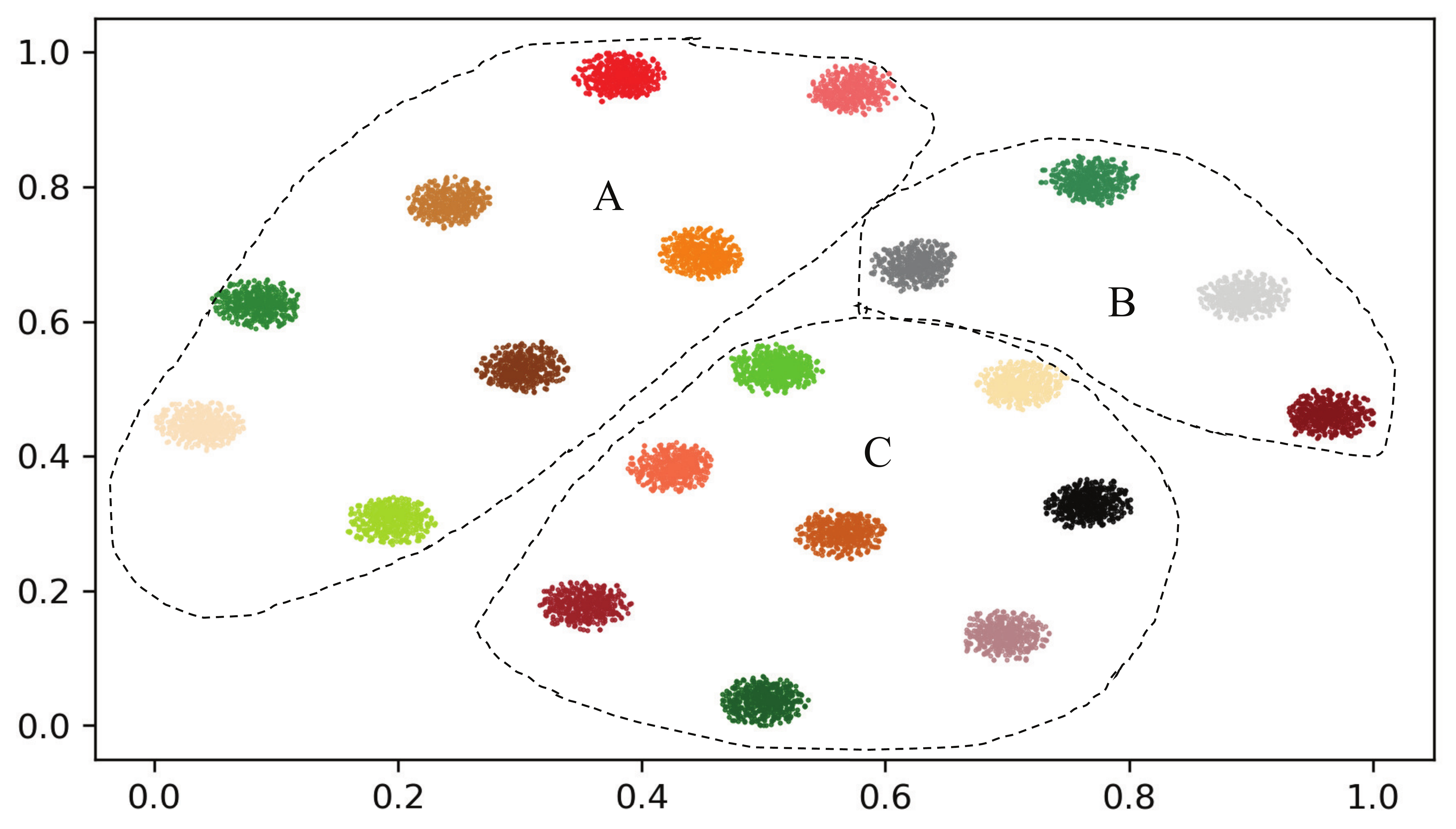}}
\subfigure[Real features of FLO]{
\includegraphics[width=5.7cm]{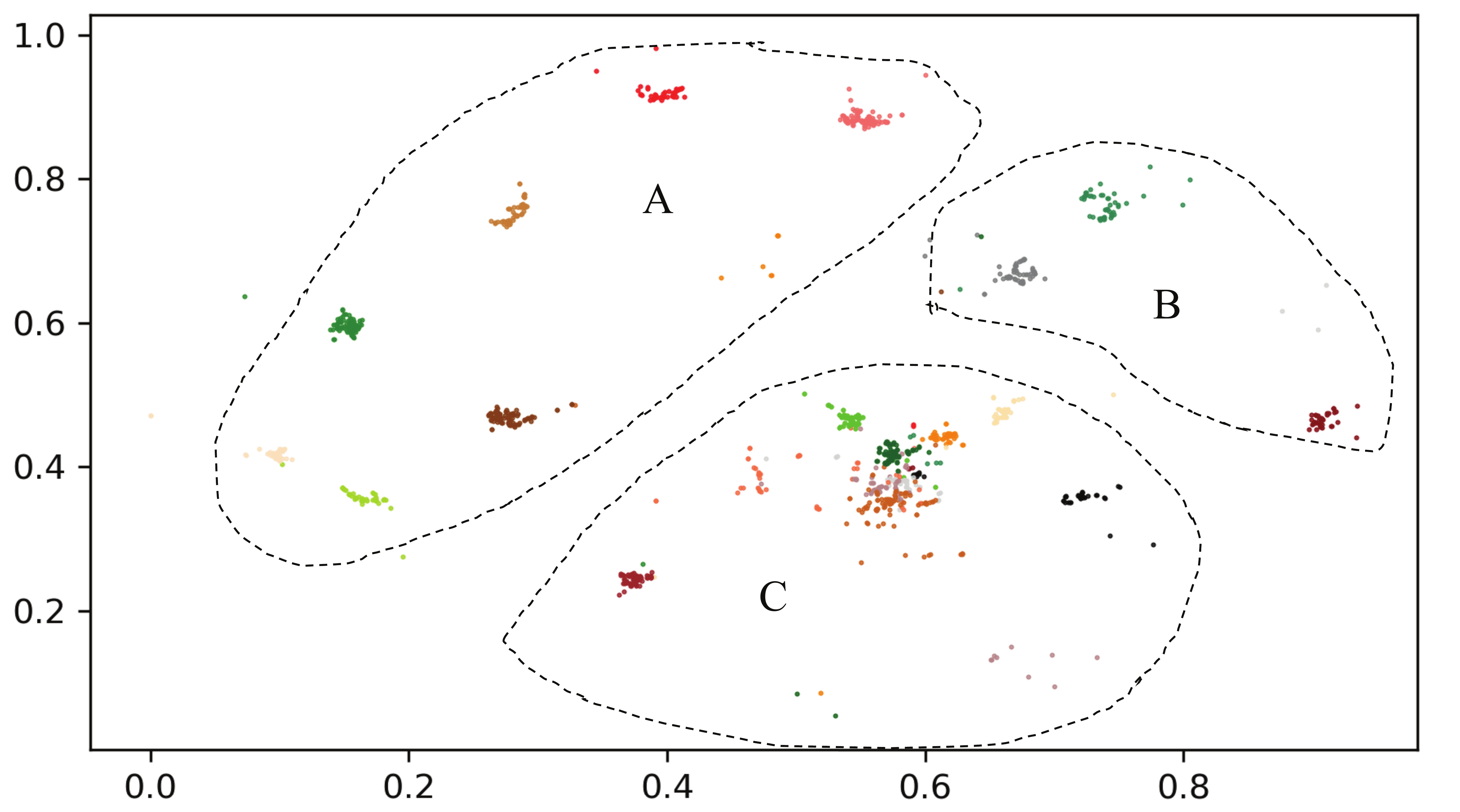}}
\subfigure[Features for FLO by f-CLSWGAN]{
\includegraphics[width=5.7cm]{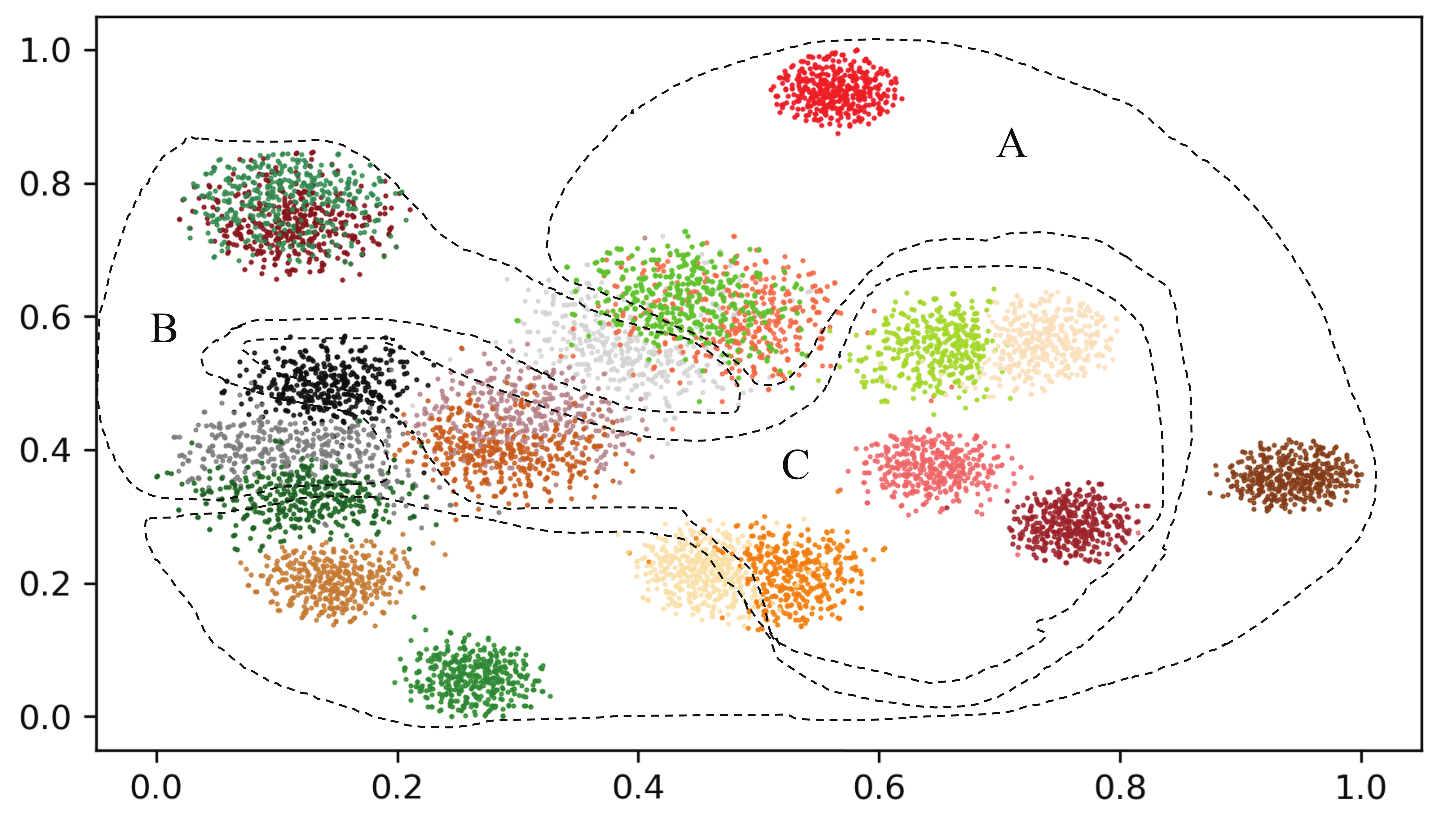}}
\subfigure[Features for AWA2 by SRWGAN]{
\includegraphics[width=5.7cm]{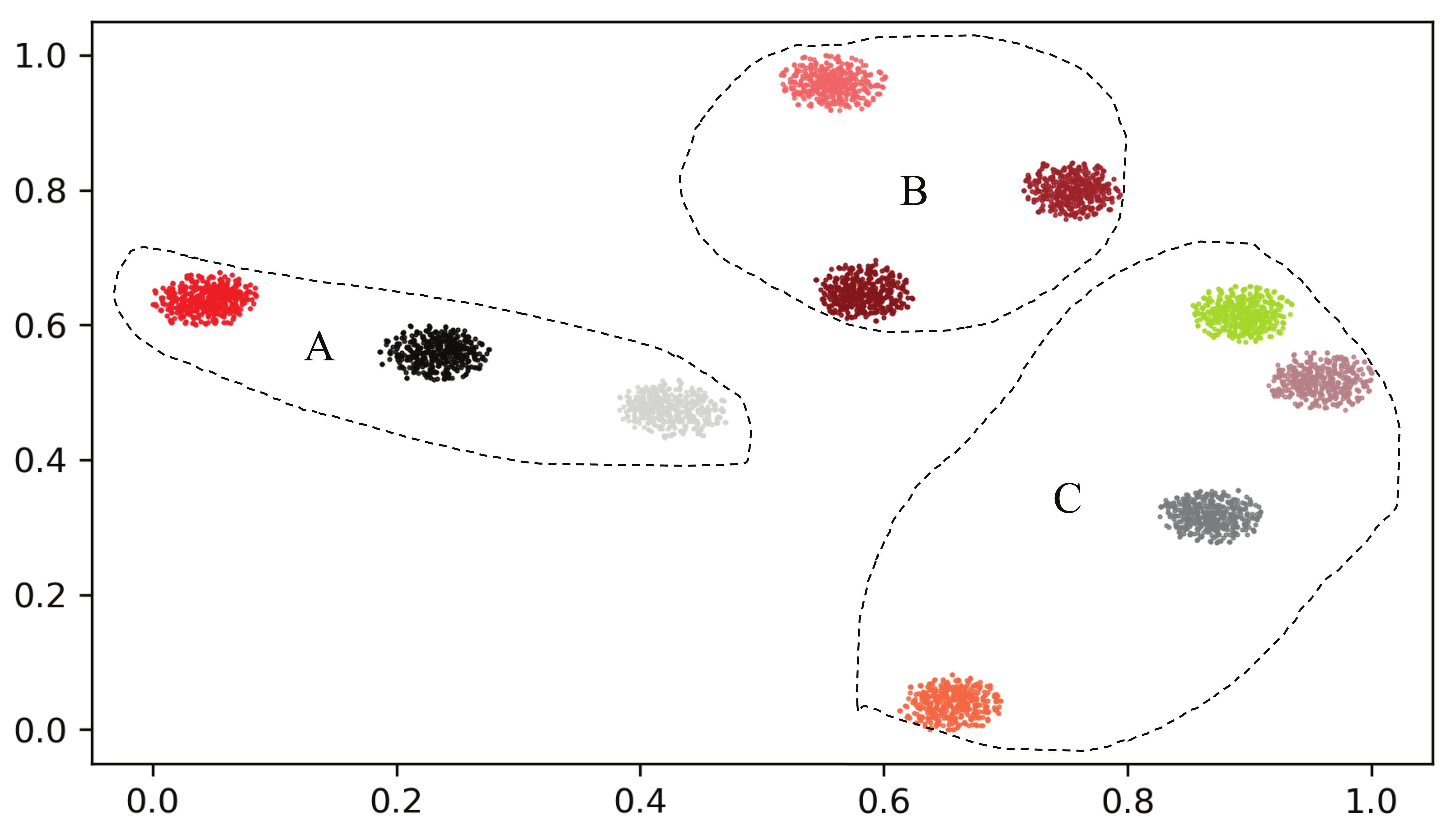}}
\subfigure[Real features of AWA2]{
\includegraphics[width=5.7cm]{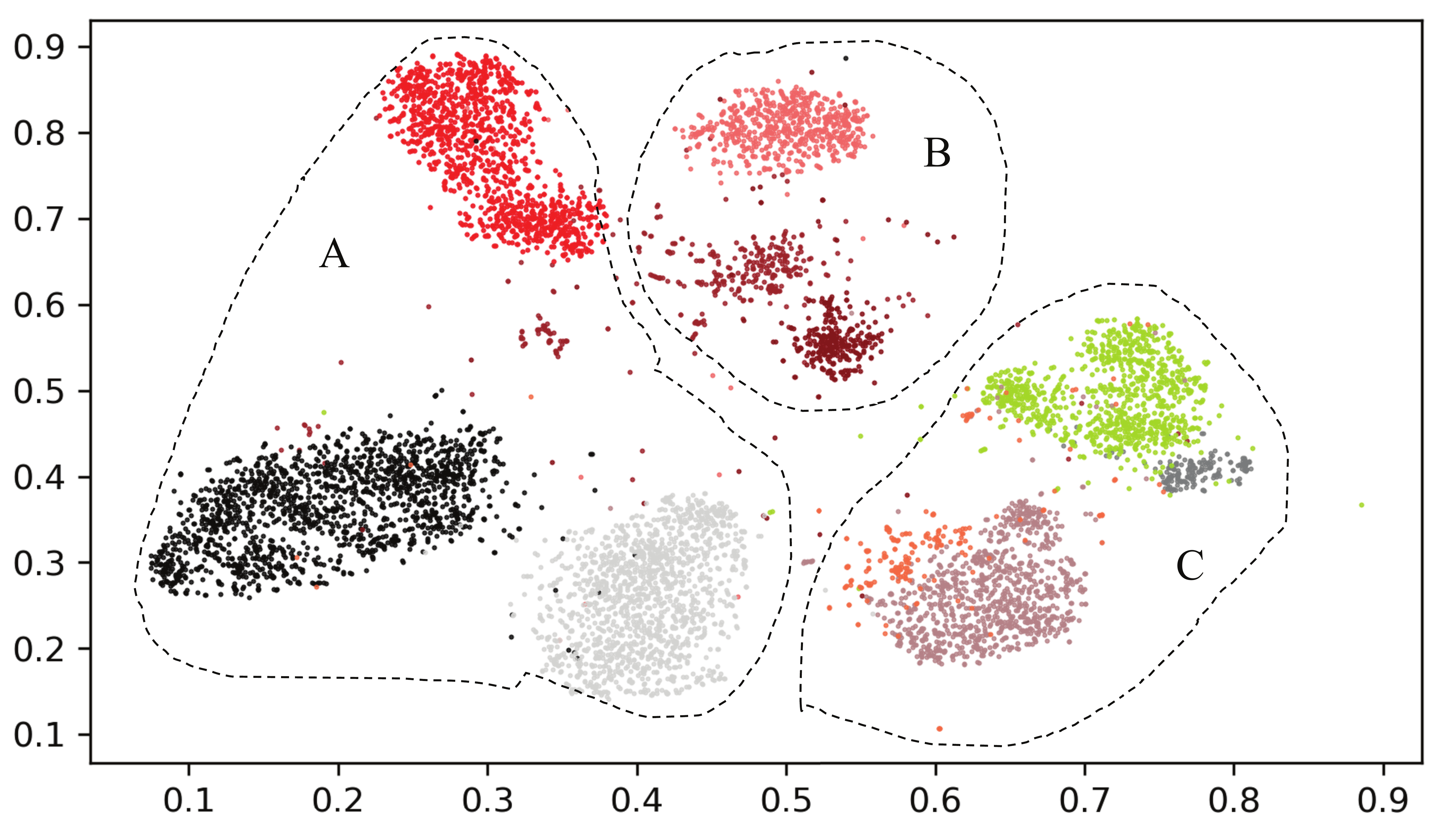}}
\subfigure[Features for AWA2 by f-CLSWGAN]{
\includegraphics[width=5.7cm]{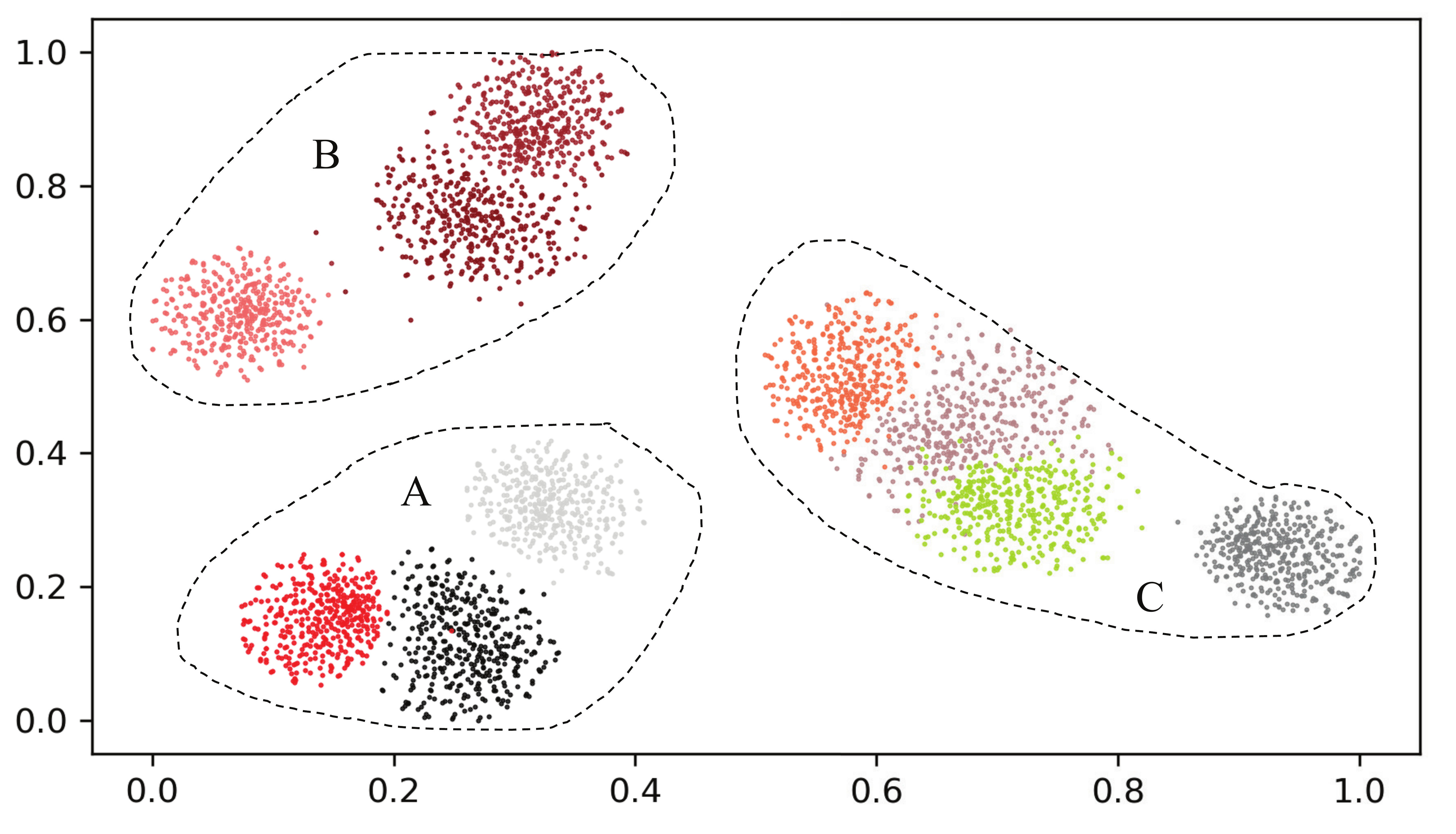}}
\caption{Comparison of feature visualization. The unseen classes of each dataset are divided into three groups for intuitive comparison.}
\vspace{-1.5em}
\end{figure*}

\subsection{Evaluation of the Semantic Refinement Techniques}
Here, seven models are designed to show the effectiveness of the proposed semantic refinement techniques and regularization items. The seven models are listed as below:\\
\vspace{-1em}
\subsubsection{Model A} Model A is the original SRWGAN model defined in Eq. (21).
\subsubsection{Model B} In comparison with Model A, Model B replaces the classification regularization item defined in Eq. (18) with a pretrained softmax classifier \cite{53} for  $\tilde{x}_{s}$. 
\subsubsection{Model C}  In comparison with Model A, Model C is designed with the two-head semantic representation module, which is regularized by the bias-eliminated semantic alignment loss defined in Eq. (6) and the auxiliary alignment loss defined in Eq. (14).
\subsubsection{Model D} In comparison with Model A, Model D is designed with the one-head semantic representation module, which is regularized by the bias-eliminated semantic alignment loss defined in Eq. (6).
\subsubsection{Model E} In comparison with Model D, the one-head semantic representation module of Model E is regularized by $\alpha\mathcal{L}_{SBC} + \beta\mathcal{L}_{UBC}$.
\subsubsection{Model F} In comparison with Model D, the one-head semantic representation module of Model F is regularized by $\alpha\mathcal{L}_{SBC}$.\par
\subsubsection{Model G} In comparison with Model D, Model G is implemented without semantic regularization and semantic representation techniques.\par
\begin{figure*}[!htb]
\centering 
\subfigure[SBC]{
\includegraphics[width=5.7cm]{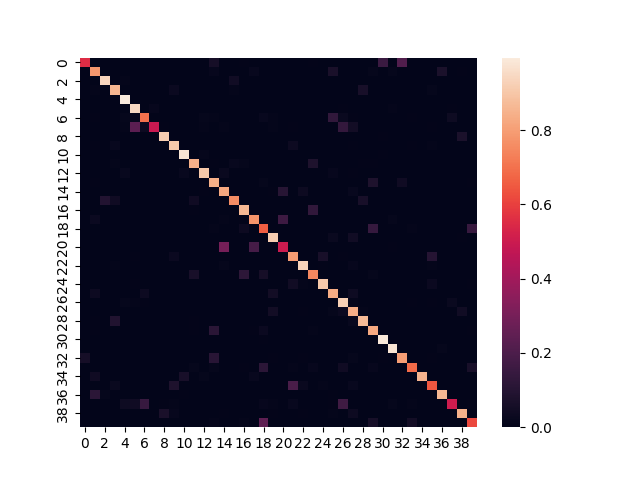}}
\subfigure[UBC]{
\includegraphics[width=5.7cm]{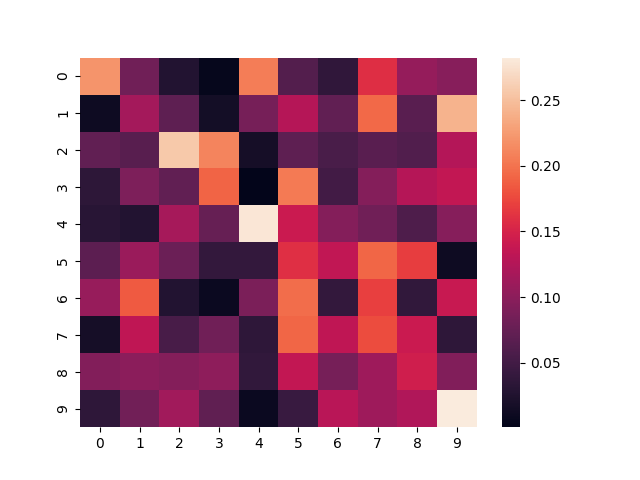}}
\subfigure[CBC]{
\includegraphics[width=5.7cm]{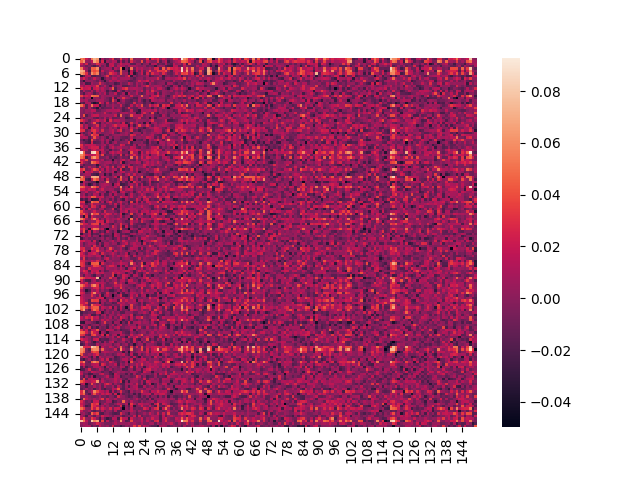}}
\caption{Visualization of the bias-eliminated condition on the AWA2 dataset.}
\vspace{-1em}
\end{figure*}

\begin{figure*}[!htb]
\centering 
\subfigure[ $a^{\ddagger}$ regularized by $\mathcal{L}_{bl}$  ]{
\includegraphics[width=5.7cm]{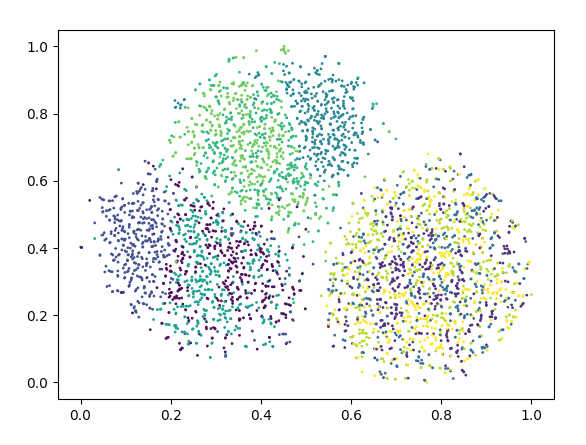}}
\subfigure[$a^{\dagger}$ regularized by $\mathcal{L}_{in-al}$ ]{
\includegraphics[width=5.8cm]{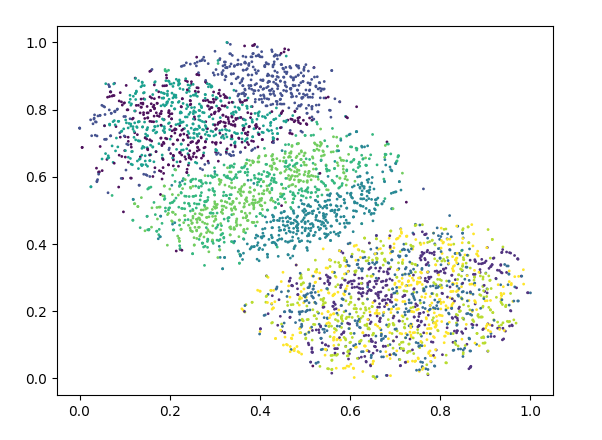}}
\subfigure[${a}'$ regularized by $\mathcal{L}_{rl}$]{
\includegraphics[width=5.6cm]{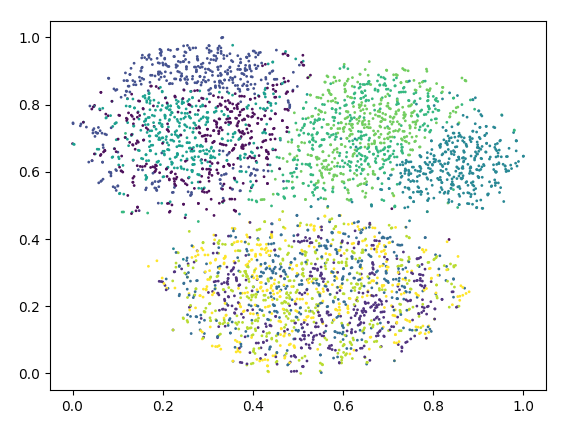}}
\caption{Visualization of the hierarchical alignment loss on the AWA2 dataset.}
\end{figure*}
\begin{figure*}[!htb]
\centering 
\subfigure[$d_{a^{\ddagger}}$]{
\includegraphics[width=5.7cm]{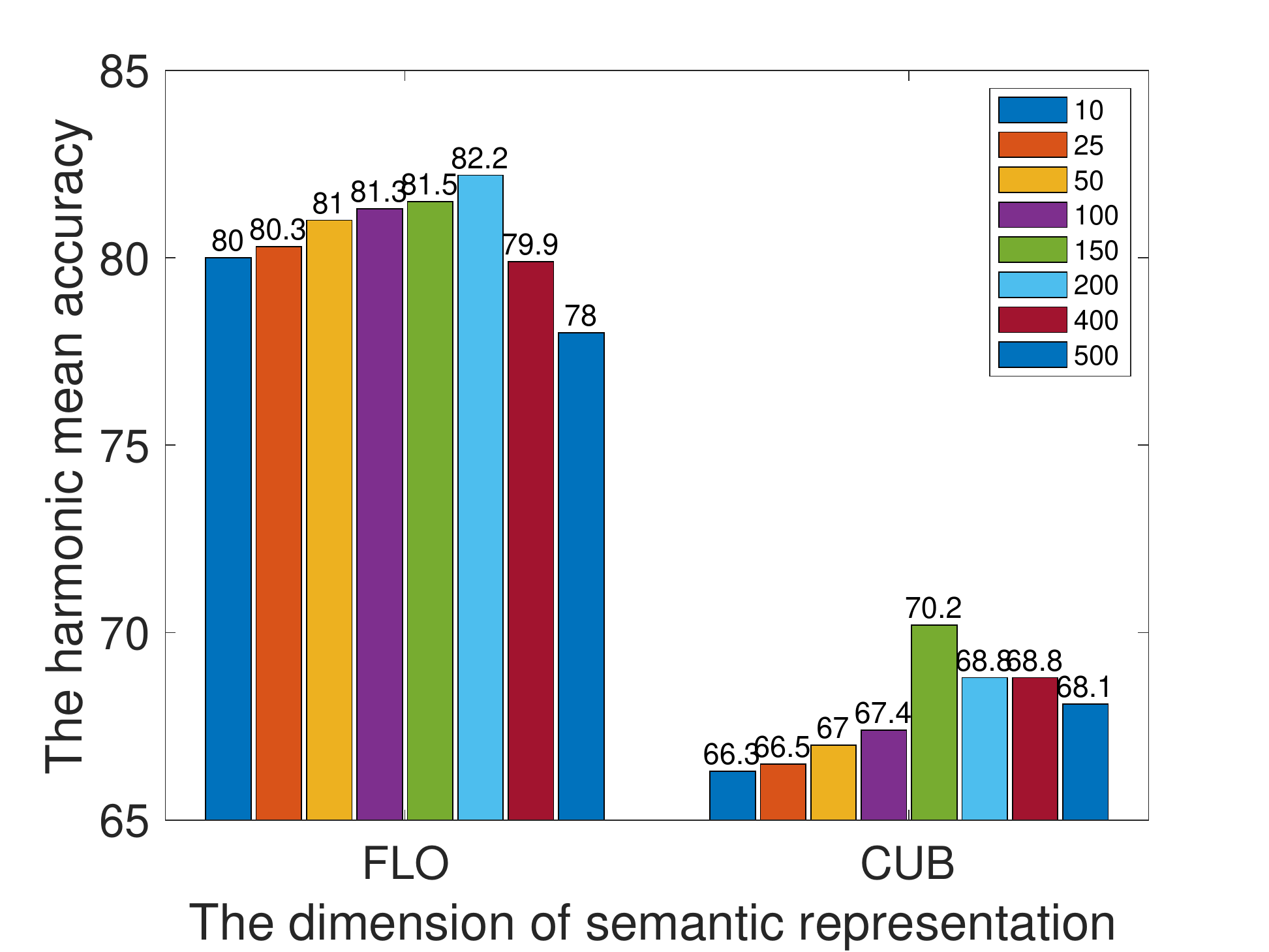}}
\subfigure[ $\alpha$]{
\includegraphics[width=5.7cm]{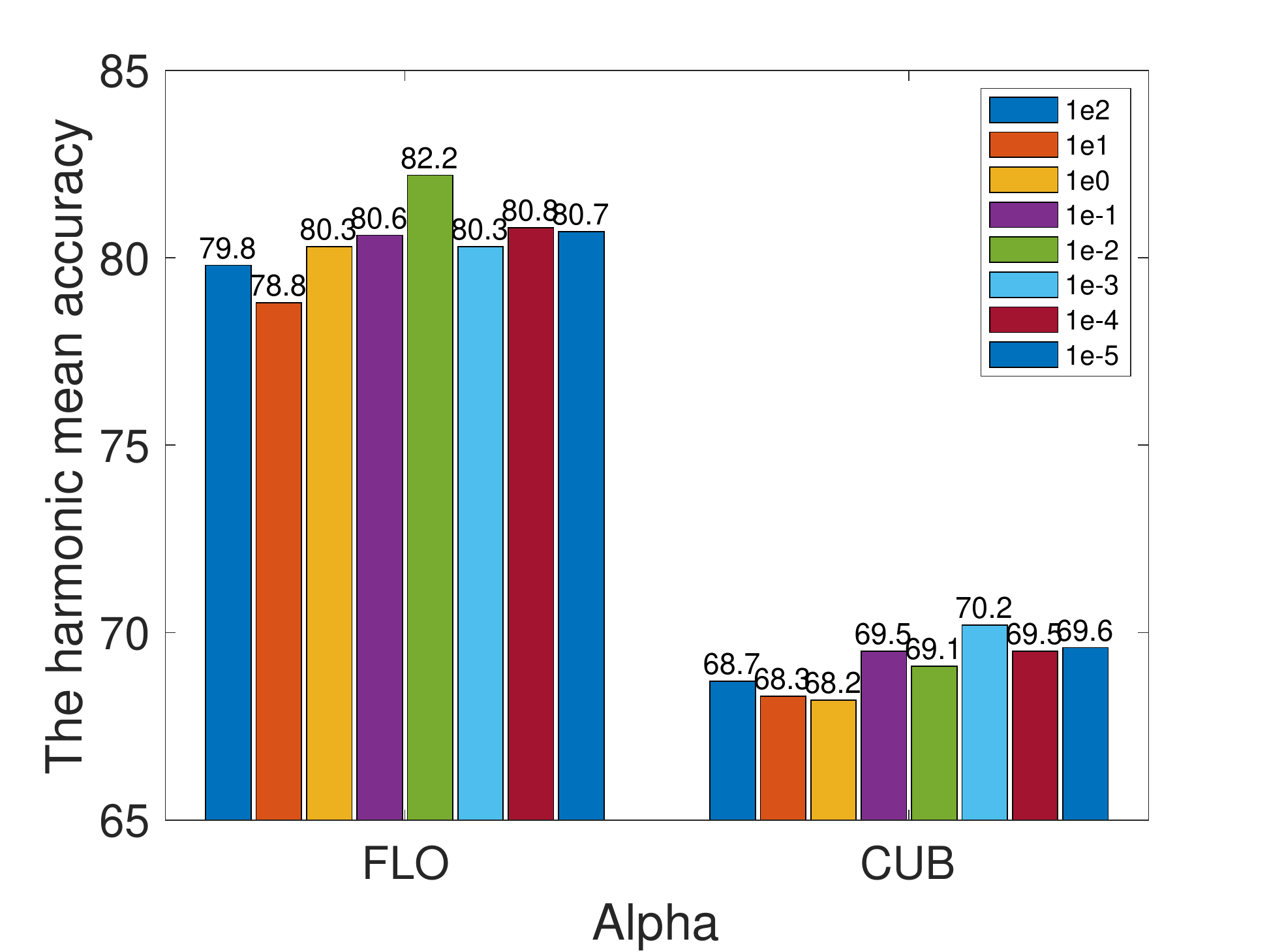}}
\subfigure[$\beta$]{
\includegraphics[width=5.7cm]{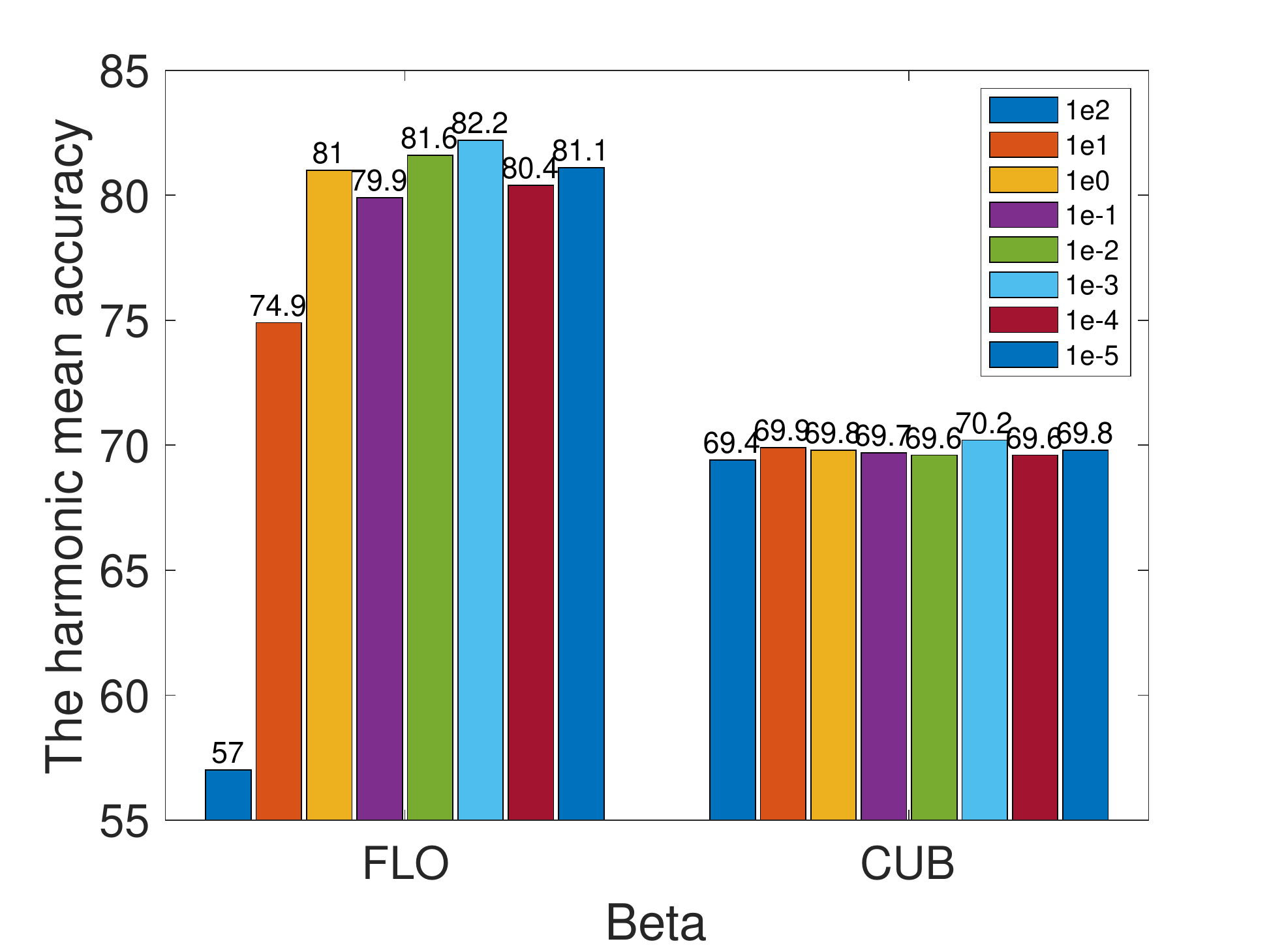}}
\subfigure[$\delta$]{
\includegraphics[width=4.2cm]{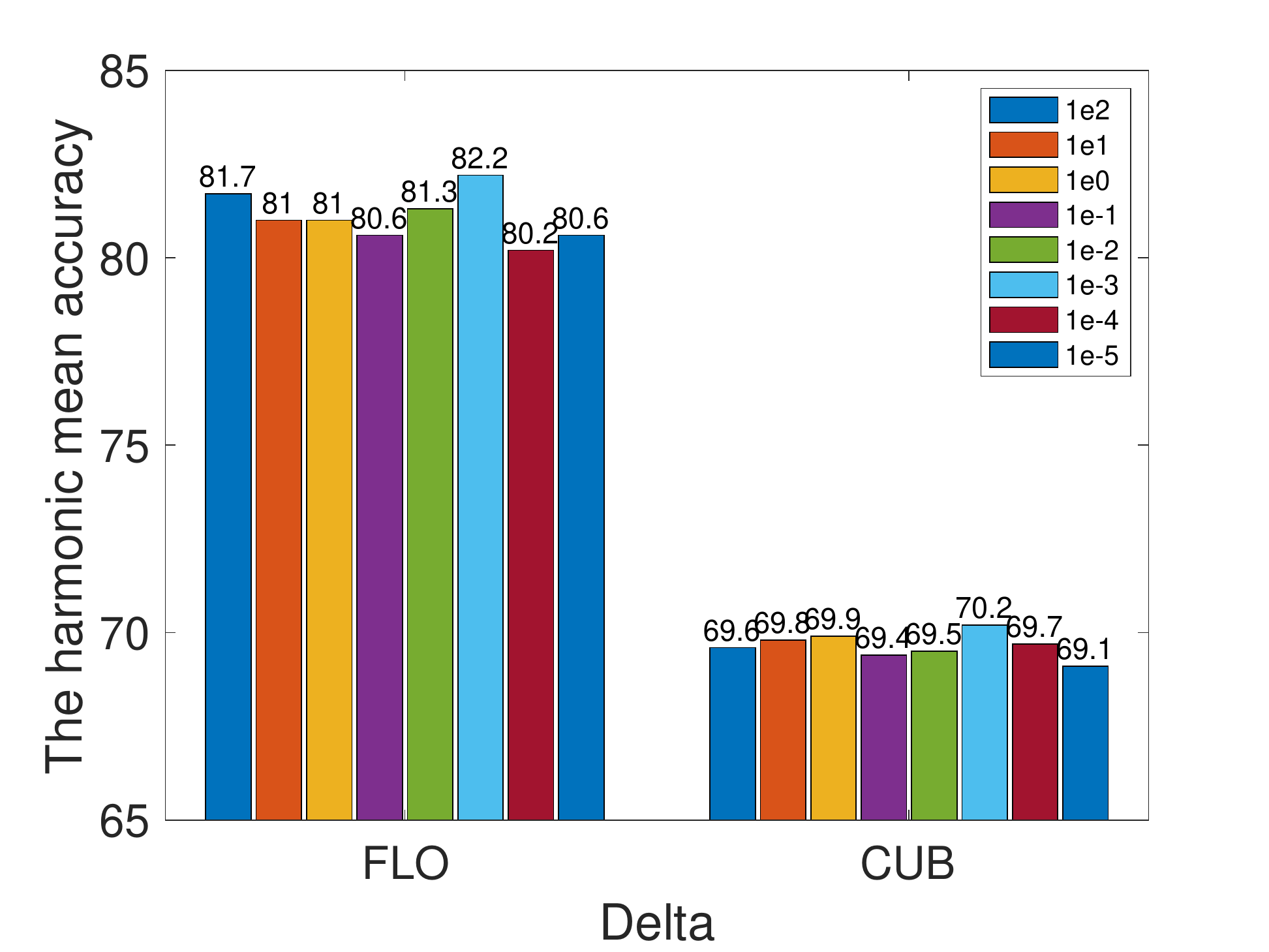}}
\subfigure[$\lambda_{a}$]{
\includegraphics[width=4.2cm]{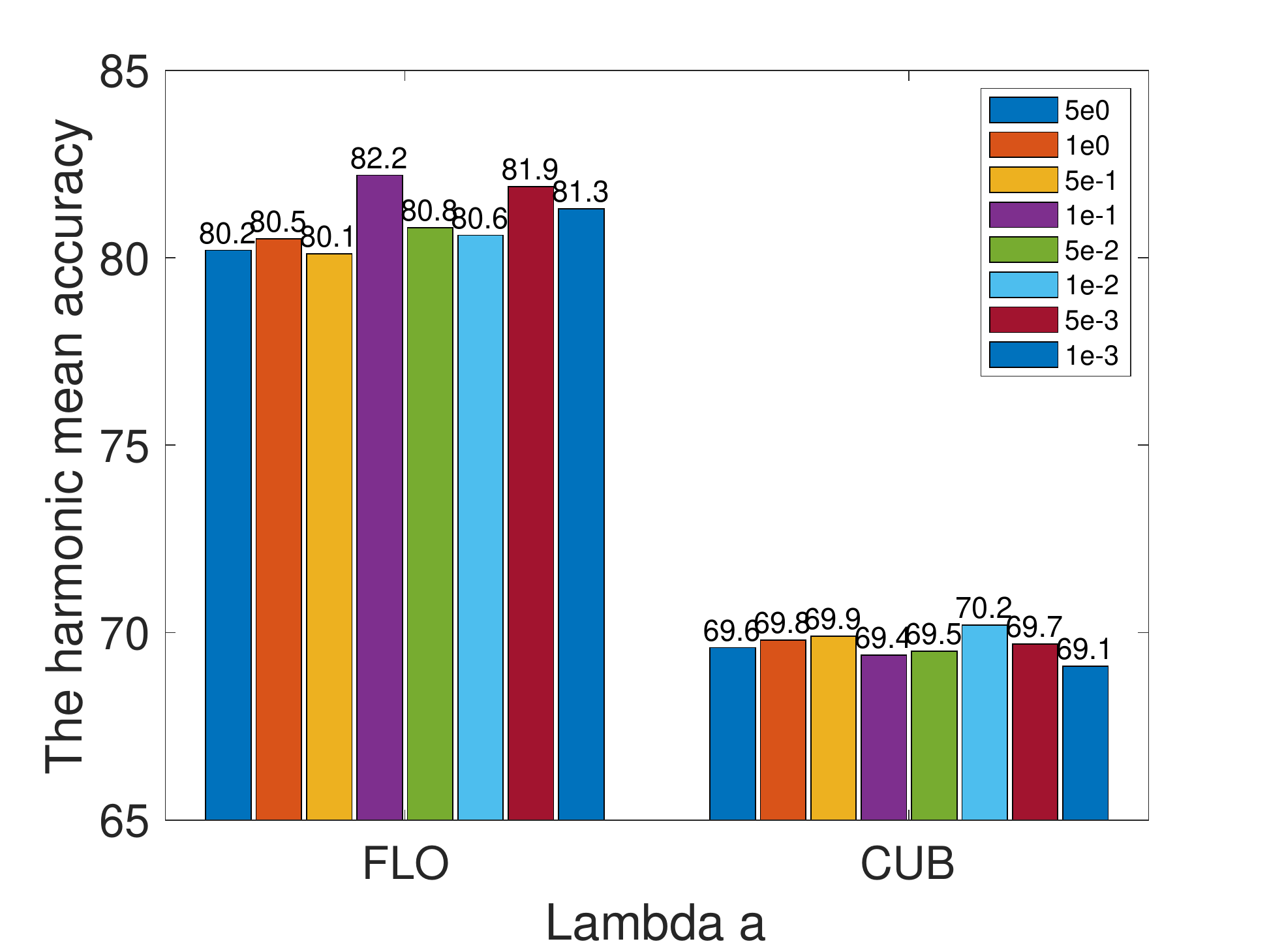}}
\subfigure[$\lambda_{c}$]{
\includegraphics[width=4.2cm]{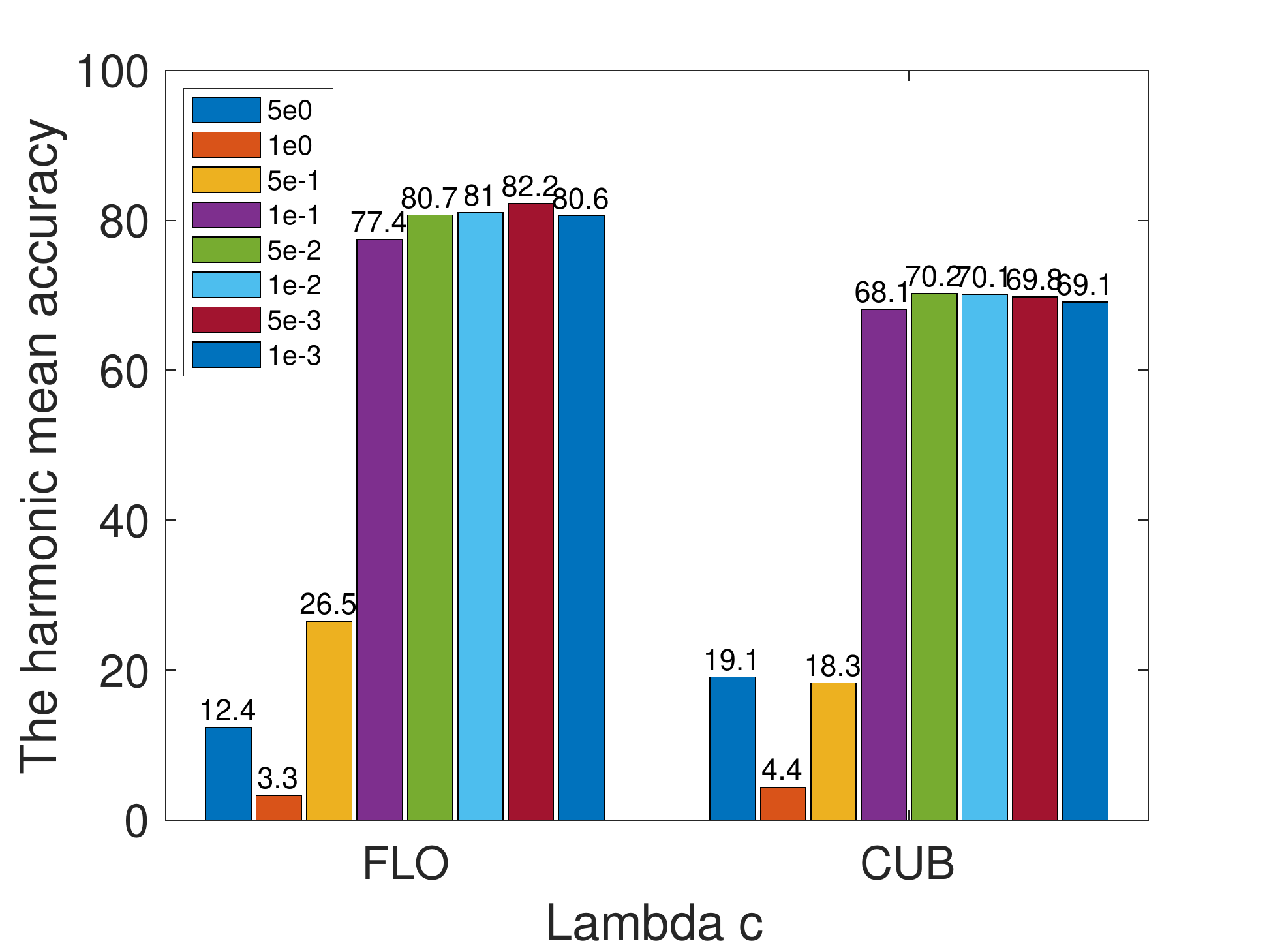}}
\subfigure[$\lambda_{m}$]{
\includegraphics[width=4.2cm]{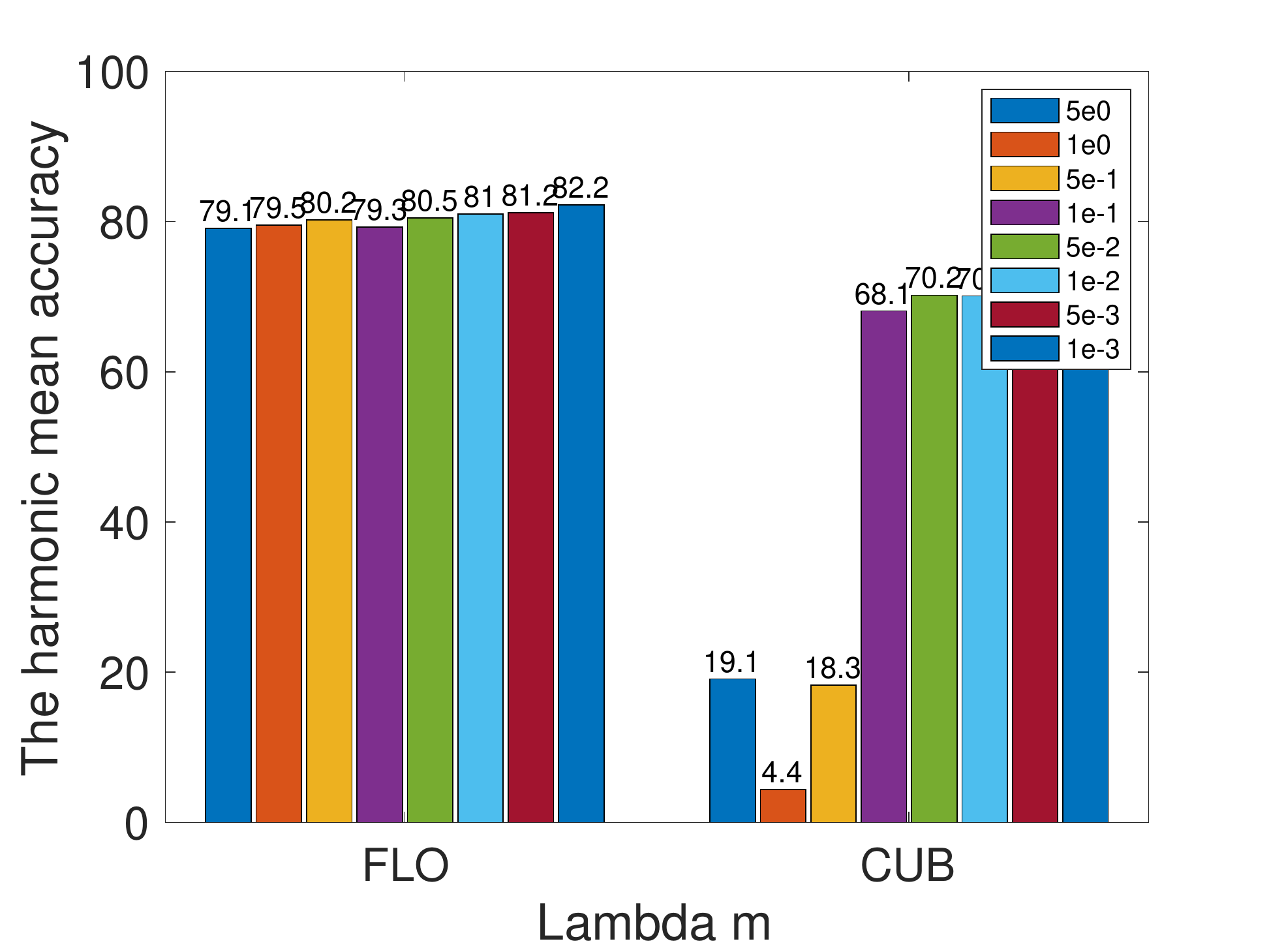}}
\caption{Parameter study based on the FLO and CUB datasets.}
\vspace{-1em}
\end{figure*}
The hyperparameters of the seven models and experimental settings are the same as those in GZSL. Two datasets, \ie, FLO, and CUB, are used. In addition to the basic $U$, $S$, and $H$, we provide the relative improvement (RI) of $H$ between Model A and other models for an intuitive comparison. $RI$ is obtained as follows:
\begin{equation}
\begin{aligned}
RI(H(A), H(others)) = \frac{H(others) - H(A)}{H(A)}*100\%.
\end{aligned}
\end{equation}
The results are summarized in Table \uppercase\expandafter{\romannumeral6}.\par
First, in comparison with Model A, Model B has a harmonic mean accuracy degradation of $1\% \sim 2\%$ on the FLO and CUB datasets. This shows that the classification regularization item defined in Eq. (18) performs better than the conventional pretrained classifier \cite{53}. The essential difference is that Eq. (18) optimizes not only the classification ability of $\tilde{x}_{s}$ but also the compactness between real and fake features based on the similarity score $c(\tilde{x}_{s})$ and hence contributes to the feature generation ability. To validate this, an intuitive strategy is comparing the feature compactness between the real and fake features. The feature compactness is defined as follows:
\begin{equation}
\begin{aligned}
FN = \frac{1}{n}\sum^{n}_{i=1}Q(\mu^{(i)}_{x}, \tilde{\mu}^{(i)}_{x}) ,
\end{aligned}
\end{equation}
where $Q$ is the cosine similarity and $n$ is the number of classes. The comparison of feature compactness of seen classes between Model A and Model B on CUB is shown in Figure 6. In comparison, Model A optimizes the feature compactness for seen classes more quickly than Model B at the beginning stage. Meanwhile, we show the average feature compactness comparison through 1200 epochs (from the 300th epoch to the 1500th epoch) for the seen and unseen classes in Figure 7. Although Model A and Model B present similar average feature compactness on seen classes, Model A obtains higher average feature compactness on unseen classes, which validates the effectiveness of Eq. (18). \par
Second, Model A usually performs better than Model C and Model D, which shows the contribution of random alignment loss and auxiliary alignment loss, respectively. In comparison with Model D, Model E and Model F show performance degradation since the regularization of Model E and Model F on the one-head semantic representation no longer meets the bias-eliminated condition. Meanwhile, we can observe that the loss $\mathcal{L}_{UBC}$ slightly contributes to the model performance by comparing Model E and Model F. In comparison with other models, Model G shows the lowest accuracy, since it uses the combination of noise and naive attributes as the generator's input and is implemented without any semantic regularization techniques.\par
In addition, it is worth mentioning that the implementation of Model G is based on RFF-GZSL with the tricks listed in the last subsection of the methodology, resulting in a strong baseline for the proposed SRWGAN. Generally, the proposed techniques achieve harmonic mean accuracy improvements of 2\% to 5\% on the benchmark datasets. \par

\subsection{Visualization Study}
\subsubsection{Visualization of the generated features} We visualize the generated unseen features by t-SNE \cite{76} to show the strong disjoint-class feature generation ability of SRWGAN. The real features and the features generated by f-CLSWGAN are used for comparison. This visualization is shown in Figure 8, in which the unseen classes of each dataset are divided into three groups for intuitive comparison. For the AWA and FLO datasets, the three groups of features of SRWGAN are very similar to the three groups of real features, which validates the bias-eliminated feature generation of SRWGAN. For the AWA2 dataset, there is a little bias for group A by SRWGAN. However, the features of SRWGAN present intraclass contraction and interclass separation, which are beneficial for classification. In comparison, the feature distributions of different classes by f-CLSWGAN usually overlap. The strong feature generation ability allows SRWGAN to present improved performance for the considered ZSL, GZSL, and FSL tasks.

\subsubsection{Visualization of the bias-eliminated condition} In the methodology, we define the matching function $f$ to enhance the compactness between semantic and visual features. Based on $f$, the SBC and UBC are enforced by the cross-entropy loss in Eq. (7) and entropy loss in Eq. (9), respectively. Here, we visualize the matching score $U_{ys}$ and $U_{yu}$ by averaging the $q(x_{s})$ in Eq. (8) and $q(\tilde{x}_{u})$ in Eq. (10) to demonstrate the feasibility of SBC and UBC. Meanwhile, the bias of CBC, \ie, $U^{T}_{a^{\ddagger}s}U_{a^{\ddagger}s} - U^{T}_{a^{\ddagger}u}U_{a^{\ddagger}u}$, is visualized. Experiments are performed on the AWA2 dataset. The results are shown in Figure 9. The $U_{ys}$ in Figure 9 (a) is almost the standard identity matrix, which indicates that the SBC can be satisfied well with the proposed method. For $U_{yu}$ in Figure 9 (b), it is not optimized to an identity metric, since we train Eq. (9) with the entropy loss in an unsupervised manner and have no real unseen samples in the training stage. However, the diagonal items in Figure 9 (b) are usually larger than others, which indicates that Eq. (9) also contributes to semantic-visual matching. In addition, Figure 9 (c) shows that the bias of CBC is small. The three well-implemented conditions provide the bias-elimination ability for SRWGAN.

\subsubsection{Visualization of the hierarchical alignment loss} In the methodology, the hierarchical alignment loss in Eq. (12), \ie, $\mathcal{L}_{in-sr}$, is defined to regularize the designed multihead semantic representation, \ie, $a^{\ddagger}$, $a^{\dagger}$, and ${a}'$. Here, we visualize the multihead semantic representation to show the effectiveness of the hierarchical alignment loss. For better visualization, we first concatenate $a^{\ddagger}$, $a^{\dagger}$, and ${a}'$ with the original $a$ and then apply the t-SNE algorithm to them. The results are summarized in Figure 10. In comparison with Figure 8, both the semantic and visual features are clustered into three groups, revealing that the semantic representation presents visual-semantic matching. Meanwhile, the three groups of semantic features show different patterns in the two-dimensional space. This property is utilized by concatenating them in the proposed SRWGAN to provide distinguished semantic information for feature generation.\par
\begin{figure}[htb]
\centering
\includegraphics[width=.5\textwidth]{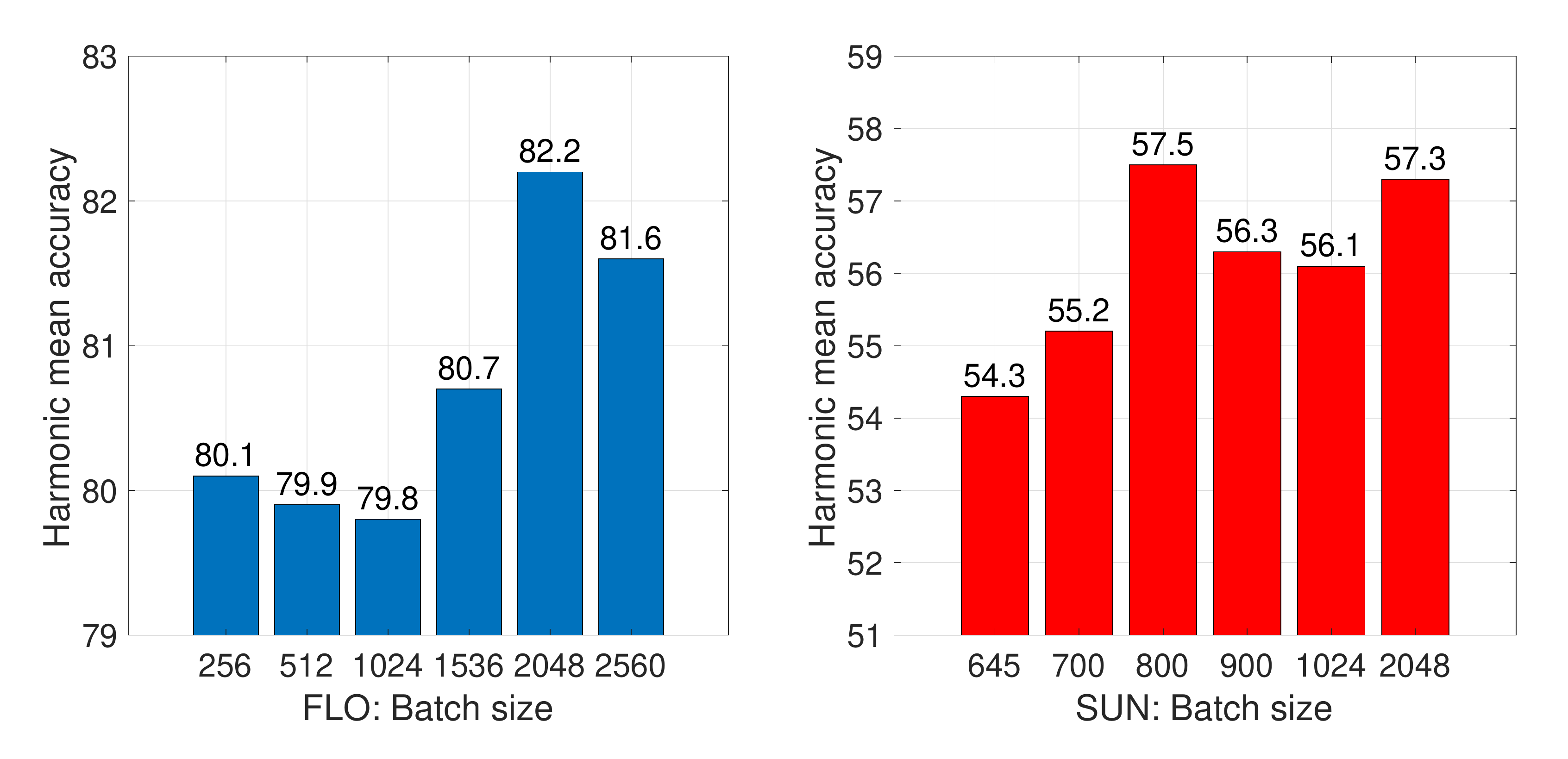} 
\caption{Ablation study for batch size based on FLO and CUB.}
\vspace{-1.5em}
\end{figure}
\subsection{Parameter Study}
In the proposed SRWGAN, the important hyperparameters include $d_{a^{\ddagger}}$, $\alpha$, $\beta$, $\delta$, $\lambda_{a}$, $\lambda_{c}$, and $\lambda_{m}$, which are listed in Table II. Here, we show the grid search results in the GZSL setting on the CUB and FLO datasets for a parameter study. The results are presented in Figure 11. Generally, the semantic dimension $d_{a^{\ddagger}}$ can be set to $150 \sim 200$ to obtain satisfactory results. For the weights of bias-eliminated conditions, \ie, $\alpha$, $\beta$, and $\delta$, the best value is between $0.01$ and $0.001$. For the semantic weight $\lambda_{a}$, the value should not be larger than 0.1 or smaller than 0.005. The classification item $\lambda_{c}$ shows a strong regularization for feature generation. A large $\lambda_{c}$ will lead to the failure of the generation process, and a proper value of $\lambda_{c}$ can be selected from 0.05 to 0.005. The redundancy-free mapping weight $\lambda_{m}$ shows different effects on the FLO and CUB datasets. A small $\lambda_{m}$, \ie, 0.001, contributes to the best model performance of FLO, and the best $\lambda_{m}$ for CUB is 0.05. To summarize, the values of hyperparameters in SRWGAN are usually selected from 1 to 0.001, and the grid search method is helpful to obtain satisfactory performance. Meanwhile, the hyperparameters, $d_{a^{\ddagger}}$, $\beta$, and $\lambda_{c}$, usually present larger effects on the accuracy and can be optimized first. In addition, we provide an ablation study to explore the effects of batch size on FLO and SUN. The results are shown in Figure 12. Usually, a larger batch size contributes to satisfactory performance.\par
\begin{table}[!htb]
\centering
    \textbf{\caption{The State-Of-the-Art Results of ZSL and TZSL}}
    \vspace{-0.1cm}
    \setlength{\tabcolsep}{1.2 mm}
    \centering
    \begin{tabular}{p{15pt}<{\centering}|p{80pt}<{\centering}|p{25pt}<{\centering}|p{18pt}<{\centering}|p{18pt}<{\centering}|p{18pt}<{\centering}|p{18pt}<{\centering}}
        \Xhline{1pt}
        \textbf{Set.} & \textbf{Method} & \textbf{\tabincell{c}{Gener\\-ative}} & \textbf{CUB} & \textbf{FLO} & \textbf{SUN} & \textbf{AWA2}  \\
        \hline
        \multirow{16}*{\textbf{I}} & SJE(2015) \cite{17} & \xmark       & 53.9 & 53.4 & 53.7 & 61.9  \\
                          & ESZSL(2015) \cite{23}        & \xmark             & 53.9 & 51.0 & 54.5 & 58.6  \\
                          & SYNC(2016) \cite{77}         & \cmark             & 55.6 & -- & 56.3 & 49.3  \\
                          & GFZSL(2017) \cite{64}        & \cmark             & 49.3 & -- & 62.9 & 63.8  \\
                          & SE-GZSL(2018) \cite{78}      & \cmark           & 59.6& --   & 63.4 & 69.2 \\
                          & CycGAN(2018) \cite{79}       & \cmark           & 58.6 & 70.3   & 60.0   & -- \\
                          & GXE(2019) \cite{48}         & \cmark               & 54.4 & -- & 62.6 & 71.1  \\
                          & CADA-VAE(2019)  \cite{58}    & \cmark         & 60.4 & 64.0   & 63.0   & 63.2 \\
                          & f-VAEGAN(2019) \cite{21}     & \cmark          & 61.0 & \textcolor[rgb]{0.00,0.00,1.00}{70.4}   & 64.7   & -- \\
                          & LisGAN(2019)  \cite{56}      & \cmark            & 58.8 & 69.6 & 61.7 & -- \\
                          & GMGAN(2019)  \cite{51}       & \cmark          & 64.6 & --   & 64.1 & -- \\
                          & DAZLE(2020) \cite{71}        & \xmark          & 65.9 & --   & --   & --  \\
                          & IZF(2020) \cite{61}          & \cmark             & 67.1 & --   & \textcolor[rgb]{1.00,0.00,0.00}{68.4} & 74.5   \\
                          & ZSML(2020) \cite{57}       & \cmark             & \textcolor[rgb]{0.00,0.00,1.00}{69.7} & --  & 60.2 & \textcolor[rgb]{0.00,0.00,1.00}{77.5} \\
                          & OCD(2020) \cite{73}        & \cmark            & 60.3 & --   & 63.5   & 71.3  \\
                          \cline{2-7}
                          & SRWGAN(Ours)                  & \cmark          & \textcolor[rgb]{1.00,0.00,0.00}{74.8} & \textcolor[rgb]{1.00,0.00,0.00}{78.1} & \textcolor[rgb]{0.00,0.00,1.00}{67.5} & \textcolor[rgb]{1.00,0.00,0.00}{81.4} \\
        \Xhline{1pt}
        \multirow{9}*{\textbf{T}} & GFZSL(2017)  \cite{64}  &  \cmark & 50.0 & --   & 64.0 & --  \\
                          & DSRL(2017) \cite{74}            &  \xmark           & 48.7 & --   & 56.8 & -- \\
                          & PREN(2019) \cite{80}            &  \xmark          & 66.4 & --   & 62.9 & 74.1  \\
                          & GMGAN(2019) \cite{51}           &  \cmark        & 64.6 & --   & 64.3 & -- \\
                          & f-VAEGAN(2019) \cite{21}        &  \cmark       & 71.1 & --   & \textcolor[rgb]{0.00,0.00,1.00}{70.1} & -- \\
                          & GXE(2019) \cite{45}             &  \cmark          & 61.3 & --   & 63.5 & 83.2 \\
                          & LFGAA(2019) \cite{81}           &  \xmark        & \textcolor[rgb]{0.00,0.00,1.00}{78.9} & 75.5   & 66.2 & -- \\
                          & SDGN(2020) \cite{47}            &  \cmark        & 74.9 & \textcolor[rgb]{1.00,0.00,0.00}{84.4}   & 68.4 & \textcolor[rgb]{0.00,0.00,1.00}{89.1}  \\
                          \cline{2-7}
                          & SRWGAN(Ours)     &  \cmark & \textcolor[rgb]{1.00,0.00,0.00}{81.6} & \textcolor[rgb]{0.00,0.00,1.00}{83.4} & \textcolor[rgb]{1.00,0.00,0.00}{72.3} & \textcolor[rgb]{1.00,0.00,0.00}{90.2} \\
        \Xhline{1pt}
        \multicolumn{6}{l}{\tabincell{l}{``\textbf{I}'' and ``\textbf{T}'' denote the inductive and transductive settings.\\ {\color{red}{Red font}} and {\color{blue}{blue font}} are the first two results in each setting.}}      
        \end{tabular}
        \vspace{-1.5em}
\end{table}
\begin{table}[t]
\vspace{-0.5em}
\centering
    \textbf{\caption{The Grid Search Results for the ZSL Setting}}
    \setlength{\tabcolsep}{4.5 mm}
    \vspace{-1em}
    \centering
    \begin{tabular}{ccccccc}
        \Xhline{1pt}
       Dataset                                                                       &  \textbf{CUB}  &   \textbf{FLO} &   \textbf{SUN}   &   \textbf{AWA2}         \\
        \hline
       $d_{a^{\ddagger}}$, $d_{a^{\dagger}}$,  $d_{{a}'}$            &    100       &     100        &    150      &     100                                 \\
       $  \alpha                $                                                   &     1e-3       &      1e-4    &      1e-2    &      1e-3                          \\ 
       $  \beta                 $                                                   &      1e-3       &      1e-3     &      1e1        &      1e-1                       \\ 
       $   \delta               $                                                   &      1e-2        &      1e-3     &      1e-2       &      1e-5                     \\ 
       $  \lambda_{a}      $                                                   &       1e-3        &      1e-1       &       1e-2   &       1e-3                         \\
       $  \lambda_{c}      $                                                   &       5e-2       &       5e-3       &       5e-3   &       1e-2                           \\
       $  \lambda_{m}     $                                                   &      1e-2         &       1e-2       &       1e-1    &       1e-2                        \\
        \Xhline{1pt}
        \end{tabular}
\vspace{-1em}
\end{table}
\subsection{Evaluation in a ZSL setting}
Finally, the model performance in the conventional ZSL setting is evaluated. Sixteen benchmark methods are used for the comparison in the inductive learning setting, and nine benchmark methods are used in the transductive setting. The experiments are performed on four datasets, \ie, CUB, FLO, SUN, and AWA2. The results are summarized in Table \uppercase\expandafter{\romannumeral7}, and the hyperparameters obtained by the grid search method \cite{82, 83, 84, 85} are listed in Table \uppercase\expandafter{\romannumeral8}. The number of synthesized samples per unseen class for each dataset is 1000 (CUB), 600 (FLO), 800 (SUN) and 2000 (AWA2). The proposed SRWGAN generally obtains state-of-the-art results with bias-eliminated feature generation for unseen classes. In the inductive learning setting, accuracy improvements of 5.1\%, 7.7\%, and 3.9\% are achieved for CUB, FLO, and AWA2, respectively. Although the IZF model shows the highest accuracy on the SUN dataset, the proposed SRWGAN obtains the second highest accuracy. In the transductive setting, higher performance can be observed in our approach, which validates the effectiveness of the proposed semantic refinement techniques.\par

\begin{table}[!h]
\vspace{-0.5em}
\centering
    \textbf{\caption{The Comparison of Training Time}}
    \setlength{\tabcolsep}{7.5 mm}
    \vspace{-1em}
    \centering
    \begin{tabular}{c|cccccc}
        \Xhline{1pt}
       Training Time(Hours)                 &   \textbf{FLO} &    \textbf{AWA2}         \\
        \hline
       f-CLSWGAN(2018)  \cite{53}  &     \textcolor[rgb]{1.00,0.00,0.00}{0.34}        &     \textcolor[rgb]{0.00,0.00,1.00}{0.44}              \\
       LisGAN(2019)         \cite{56}    &     0.81      &      0.50             \\ 
       f-VAEGAN(2019)     \cite{21}   &     0.50       &     0.73              \\ 
       RFF-GZSL(2020)   \cite{55}     &    0.53      &      \textcolor[rgb]{1.00,0.00,0.00}{0.40}            \\
       \Xhline{1pt}
       SRWGAN (Ours)                     &     \textcolor[rgb]{0.00,0.00,1.00}{0.44}      &     0.61            \\ 
        \Xhline{1pt}   
        \end{tabular}
\vspace{-0.5em}
\end{table}
In addition, the training time of SRWGAN is compared with that of other generative models on FLO and AWA2. The training time here is defined as the time cost of a model to obtain its best performance. We use PyTorch 1.2.0 for model implementation and a 2080 Ti GPU for model training. The comparison is shown in Table \uppercase\expandafter{\romannumeral9}. Although the proposed SRWGAN seems time-consuming and tedious with the proposed regularization, it is actually quite efficient for obtaining satisfactory results. Similarly, the complex models, \ie, RFF-GZSL and f-VAEGAN, sometimes take less time than the basic models, \ie, f-CLSWGAN and LisGAN, since additional techniques are usually helpful in improving model performance and generating reliable features with fewer epochs.
\section{Conclusions}
In this paper, we propose to refine the semantic descriptions for any-shot learning tasks. A new model, namely, the SRWGAN, is developed by designing the multihead semantic representation technique and the hierarchical semantic alignment technique. Toward the bias-eliminated conditions, the proposed SRWGAN generates strong virtual features for the disjoint-unseen classes, the effectiveness of which is validated on six benchmark datasets in both inductive and transductive settings. In comparison with some state-of-the-art methods, accuracy improvements of 2\% to 4\% are observed for the ZSL, GZSL, and FSL tasks. In addition to the addressed bias problem, there may be some overlapping information in the learning for the seen and unseen classes. The problem of how to define and utilize the intersection between the seen and unseen classes for a generator or other mappings is worthy of discussion and research for any-shot learning.



\ifCLASSOPTIONcaptionsoff
  \newpage
\fi



%

\vspace{-12mm}
\begin{IEEEbiography}[{\includegraphics[width=0.9in,height=1.1in,clip,keepaspectratio]{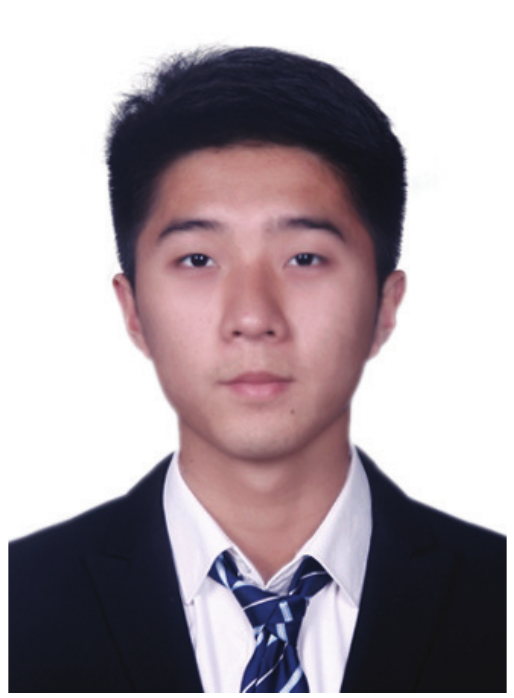}}]{Liangjun Feng}
received his B.Eng. from North China Electric Power University, Beijing, China, in 2017. He is currently pursuing his Ph.D. at the College of Control Science and Engineering, Zhejiang University, Hangzhou, China. He has authored or coauthored more than 10 papers in peer-reviewed international journals. He is the reviewer of several journals, including IEEE TII and IEEE TNNLS. His current research interests include zero-shot and few-shot learning.
\end{IEEEbiography}
\vspace{-15 mm}
\begin{IEEEbiography}[{\includegraphics[width=0.9in,height=1.1in,clip,keepaspectratio]{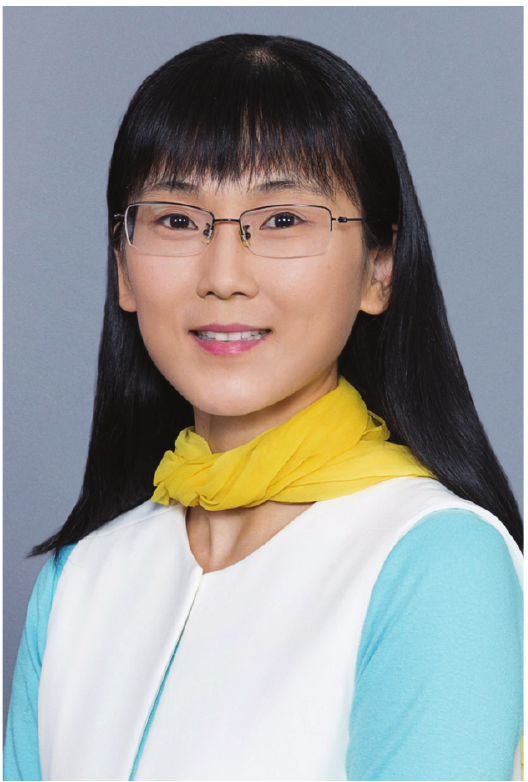}}]{Chunhui Zhao}
(SM'15) received her Ph.D. degree from Northeastern University, China, in 2009. From 2009 to 2012, she was a Postdoctoral Fellow with the Hong Kong University of Science and Technology and the University of California, Santa Barbara, Los Angeles, CA, USA. Since January 2012, she has been a Professor with the College of Control Science and Engineering, Zhejiang University, China. Her research interests include statistical machine learning and data mining.\par
\end{IEEEbiography}
\vspace{-15 mm}
\begin{IEEEbiography}[{\includegraphics[width=0.9in,height=1.1in,clip,keepaspectratio]{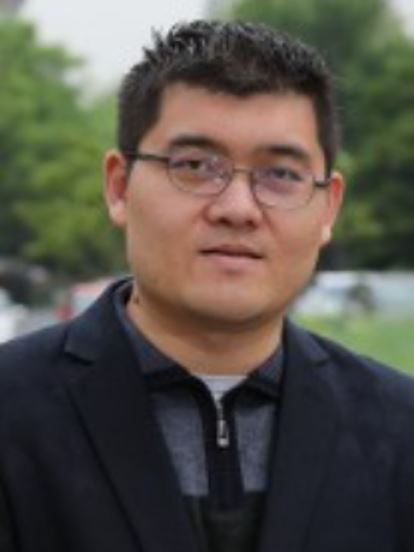}}]{Xi Li}
received his Ph.D. degree in 2009 from the National Laboratory of Pattern Recognition, Chinese Academy of Sciences, Beijing, China. From 2009 to 2010, he was a Post-Doctoral Researcher with CNRS Telecom ParisTech, France. He was a senior researcher at the University of Adelaide, Australia. He is currently a Full Professor at Zhejiang University, China. His research interests include visual tracking, compact learning, motion analysis, face recognition, and data mining.\par
\end{IEEEbiography}

\end{document}